%% file: main.tex
\pdfoutput=1

\documentclass[lettersize,journal]{IEEEtran}
\usepackage{amsmath,amsfonts}
\usepackage{algorithmic}
\usepackage{algorithm}
\usepackage{array}
\usepackage[caption=false,font=normalsize,labelfont=sf,textfont=sf]{subfig}
\usepackage{textcomp}
\usepackage{stfloats}
\usepackage{url}
\usepackage{verbatim}
\usepackage{graphicx}
\usepackage{cite}

\usepackage{latexsym}
\usepackage{bm}
\usepackage{multirow}
\usepackage{colortbl}
\usepackage{bm}
\usepackage{booktabs}
\usepackage{balance}
\usepackage{enumitem}
\usepackage{bbding}
\usepackage{balance}
\usepackage{multicol}
\usepackage[percent]{overpic}
\usepackage{nicematrix}
\usepackage{comment}
\usepackage{times}
\usepackage{soul}
\usepackage[utf8]{inputenc}
\usepackage[small]{caption}
\usepackage{amsmath}
\usepackage{booktabs}
\usepackage{color}
\usepackage{amsfonts}
\usepackage{amssymb}
\usepackage{bbm}
\usepackage{arydshln}
\usepackage{listings}
\usepackage{tabularx}
\usepackage{tikz}
\usepackage{rotating} %
\usepackage[fixed]{fontawesome5}

\usepackage{makecell}

\newtheorem{defn}{\hspace{-1mm} Definition}

\newtheorem{discussion}{\noindent \textbf{Discussion}}

\definecolor{c1}{HTML}{344C11}
\definecolor{c2}{HTML}{7e0f12}

\definecolor{t1}{HTML}{56004f}
\definecolor{t2}{HTML}{be002f}
\definecolor{t3}{HTML}{003371}
\definecolor{t4}{HTML}{ff7500}
\definecolor{t5}{HTML}{426666}
\definecolor{t6}{HTML}{9b4400}
\definecolor{t7}{HTML}{057748}

\usepackage[edges]{forest}
\definecolor{hiddendraw}{RGB}{205, 44, 36}
\definecolor{hidden-blue}{RGB}{194,232,247}
\definecolor{hidden-orange}{RGB}{243,202,120}
\definecolor{hidden-yellow}{RGB}{242,244,193}
\definecolor{uclablue}{rgb}{0.15, 0.45, 0.68}

\usepackage{hyperref}
\hypersetup{
    breaklinks,
    colorlinks=true,
    citecolor=uclablue,
    urlcolor=uclablue,
}

\tikzstyle{mybox}=[
    rectangle,
    draw=hiddendraw,
    rounded corners,
    text opacity=1,
    minimum height=1.5em,
    minimum width=5em,
    inner sep=2pt,
    align=center,
    fill opacity=.5,
    ]

\newcommand{\cmark}{{\textcolor{c1}\Checkmark}}
\newcommand{\xmark}{{\textcolor{c2}\XSolidBrush}}

\newcommand{\rmnum}[1]{\romannumeral #1}
\newcommand{\Rmnum}[1]{\expandafter\@slowromancap\romannumeral #1@}

\makeatletter
\renewcommand*\l@section[2]{%
  \@dottedtocline{1}{0em}{2.3em}{\bfseries #1}{#2}
}
\renewcommand*\l@subsection[2]{%
  \@dottedtocline{2}{1.5em}{2.8em}{\small #1}{#2}
}
\renewcommand*\l@subsubsection[2]{%
  \@dottedtocline{3}{3.0em}{3.2em}{\small #1}{#2}
}

\makeatother

\begin{document}

\title{Knowledge Graphs Meet Multi-Modal Learning: \\A Comprehensive Survey}

\input{00-author.tex}

\markboth{Journal of \LaTeX\ Class Files,~Vol.~14, No.~8, August~2021}%
{Shell \MakeLowercase{\textit{et al.}}: A Sample Article Using IEEEtran.cls for IEEE Journals}


\maketitle

\begin{abstract}
Knowledge Graphs (KGs) play a pivotal role in advancing various AI applications, with the semantic web community's exploration into multi-modal dimensions unlocking new avenues for innovation. 
In this survey, we carefully review over 300 articles, focusing on KG-aware research in two principal aspects: KG-driven Multi-Modal (KG4MM) learning, where KGs support multi-modal tasks, and Multi-Modal Knowledge Graph (MM4KG), which extends KG studies into the MMKG realm.
We begin by defining KGs and MMKGs, then explore their construction progress. 
Our review includes two primary task categories: KG-aware multi-modal learning tasks, such as Image Classification and Visual Question Answering, and intrinsic MMKG tasks like Multi-modal Knowledge Graph Completion and Entity Alignment, highlighting specific research trajectories. 
For most of these tasks, we provide definitions, evaluation benchmarks, and additionally outline essential insights for conducting relevant research. Finally, we discuss current challenges and identify emerging trends,  such as progress in Large Language Modeling and Multi-modal Pre-training strategies.
This survey aims to serve as a comprehensive reference for researchers already involved in or considering delving into KG and multi-modal learning research, offering insights into the evolving landscape of MMKG research and supporting future work. 
\end{abstract}


{
  \hypersetup{linktoc=page}
  \tableofcontents
}


\input{1-intro}

\input{2-pre}

\input{3-kg-cst}

\input{4-kg-mm-task}

\input{5-mm-kg-task}

\input{7-kg-mm-fut}

\input{8-conclusion}

\bibliographystyle{IEEEtran}
\bibliography{reference}

\end{document}

%% file: 00-author.tex
%

\author{Zhuo Chen, Yichi Zhang, Yin Fang, Yuxia Geng, Lingbing Guo, Xiang Chen, Qian Li, Wen Zhang$^{*}$, \\Jiaoyan Chen, Yushan Zhu, Jiaqi Li, Xiaoze Liu, Jeff Z. Pan, Ningyu Zhang, Huajun Chen$^{*}$\\
\faGithub \textbf{\tt \url{https://github.com/zjukg/KG-MM-Survey}}
\thanks{
Zhuo Chen (zhuo.chen@zju.edu.cn), Yichi Zhang, Yin Fang, Lingbing Guo, Xiang Chen, Wen Zhang (zhang.wen@zju.edu.cn), Yushan Zhu, Ningyu Zhang and Huajun Chen (huajunsir@zju.edu.cn) are from \textbf{Zhejiang University}, China. 
Yuxia Geng is from \textbf{Hangzhou Dianzi University}, China.
Jiaoyan Chen is from \textbf{The University of Manchester and University of Oxford}, UK. 
Jiaqi Li is from \textbf{Southeast University}, China.
Xiaoze Liu is from \textbf{Purdue University}, USA.
Jeff Z. Pan is from \textbf{The University of Edinburgh}, UK.
}
\thanks{$*$ denotes corresponding authors.}
}

%% file: 1-intro.tex
\begin{figure}[!htbp]
  \centering
   \vspace{-1pt}
\includegraphics[width=0.95\linewidth]{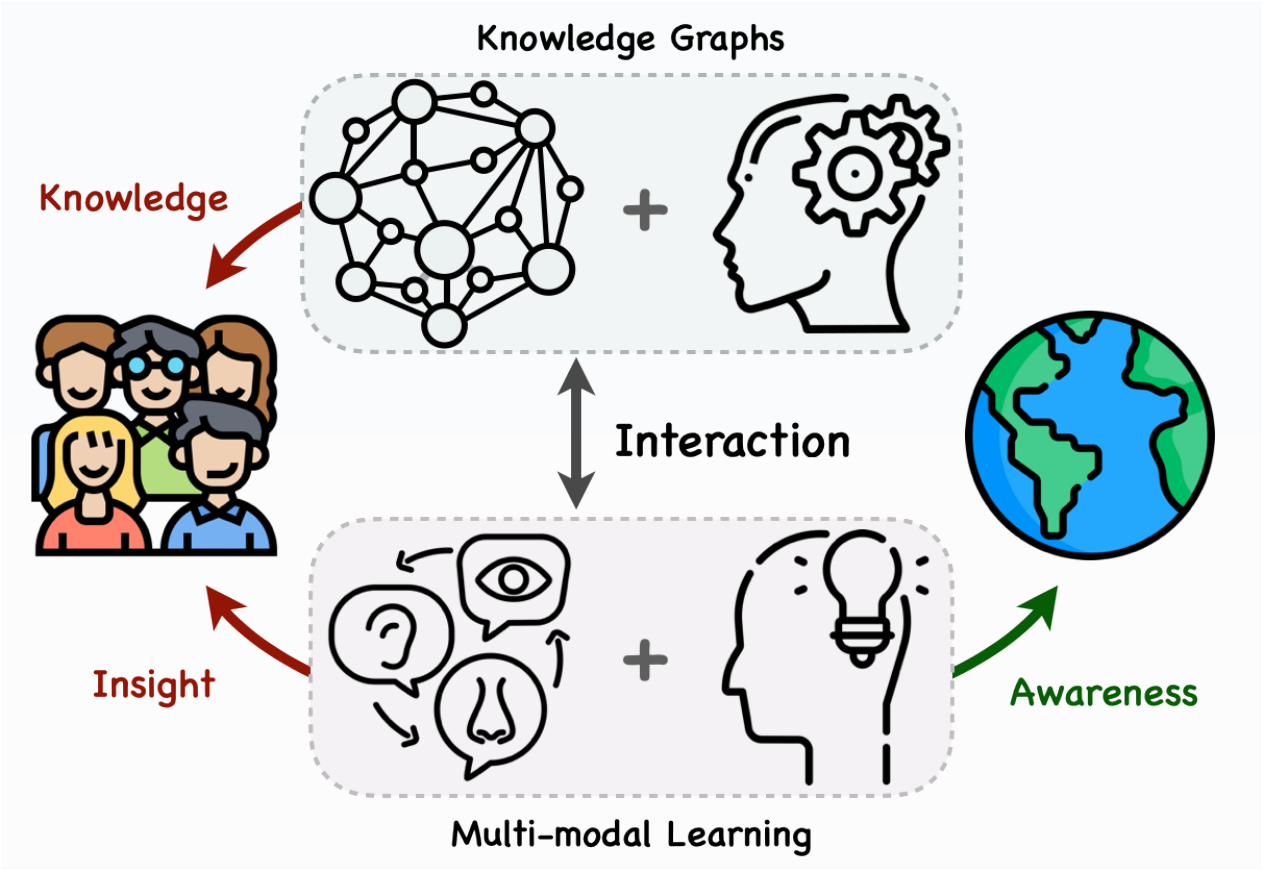}
  \vspace{-1pt}
  \caption{Knowledge Graphs Meet Multi-modal Learning.
  }
  \label{fig:Intro}
 \vspace{-10pt}
\end{figure}

\section{Introduction}
Considering knowledge reasoning and multi-modal perception in isolation from each other may not be the most appropriate strategy~\cite{lecun2022path}. This parallels human cognition, where the brain's accumulation of memories over time forms a crucial base for societal adaptation and survival, enabling meaningful actions and interactions.
These memories can be divided into two primary categories.

The first category resembles conditioned reflexes. Through repeated practice, humans develop a sort of intuitive memory that enhances intuitive and analogical reasoning skills, often referred to as shallow knowledge. When such shallow knowledge is combined with sensory inputs like visual, auditory, and tactile data, it enables us to efficiently perform basic tasks. This ability is at the heart of what traditional multi-modal tasks strive to achieve.
Multi-modal tasks, which involve data from multiple modalities for problem-solving, more closely mimic real-life situations than traditional uni-modal Natural Language Processing (NLP) or Computer Vision (CV) tasks.
For example, Visual Question Answering builds on NLP Q\&A task by incorporating visual data to predict answers from both an image and a textual question. Similarly, Image Captioning extends NLG principles by creating descriptive sentences for images, providing a fuller understanding of the content.
Consequently, with the rapid advancement of the Internet and the removal of bandwidth limitations, multi-modal information sources have become crucial and readily accessible, enabling more precise access to information.

The second type, known as \textit{Torso-to-tail Knowledge}, is encountered less frequently in everyday life and often does not lead to conditioned reflex formation. This category requires active memorization or contemplation, highlighting the significance of Knowledge Graphs (KGs) in capturing and structuring long-tail knowledge. While current large-scale pre-training efforts assimilate knowledge, they also face challenges such as hallucination phenomena and blurring of unusual knowledge~\cite{DBLP:journals/corr/abs-2308-14217,DBLP:journals/corr/abs-2308-10168,DBLP:journals/corr/abs-2308-06374,DBLP:journals/corr/abs-2306-08302}. In contrast, our study primarily focuses on the utilization of symbolic, structured knowledge within KGs. Given the vital role of KGs in organizing long-tail knowledge and their proven effectiveness as a foundational knowledge representation element across many successful AI and information systems~\cite{DBLP:journals/csur/HoganBCdMGKGNNN21}, 
it becomes evident that integrating KGs with multi-modal learning offers a promising avenue for further addressing those existing challenges.

\input{1.1-mot}
\input{1.2-diff}

\input{tab/param-tb}
\input{1.3-org}

%% file: 1.1-mot.tex
\subsection{Motivation and Contribution}
As illustrated in Fig~\ref{fig:Intro}, individuals in real life need to simultaneously process multi-modal information from the environment, while also continuously absorbing and utilizing outside knowledge.
These elements should not function in isolation; rather, knowledge and multi-modality are inherently complementary. Despite this intrinsic connection, historically, these two domains have developed independently. Previous work focus either on KG-enhanced multi-modal learning or on multi-modal KG research itself. Until now, no study or review has yet provided a comprehensive, balanced analysis of these fields,  leading to a further divide in their development.

Within this paper, we first trace the evolution from conventional KGs to MMKGs, noting the evolving focus of the semantic web community. We then carefully categorize KG-driven multi-modal tasks, where KGs serve as pivotal repositories of knowledge, providing both a basis for inference and essential knowledge for various downstream multi-modal tasks.  Following this, we explore the impact of multi-modal techniques on KGs,  discussing both their current state and future prospects. Detailed analysis covers methodological developments within each task and benchmarks key areas, enabling effective comparison across tasks.
Focusing primarily on research from the past three years (2020-2023), this survey also includes a discussion on the recent advancements in Large Language Models (LLMs), exploring their interaction with the topics covered.
It is suitable for all AI researchers, especially beneficial for those delving into knowledge-driven multi-modal reasoning and cross-modal knowledge representation, as well as serving as a valuable resource for practitioners in semantic web techniques seeking new insights.

\textbf{Literature Collection Methodology:}
For our paper, we source literature primarily from Google Scholar and arXiv. Google Scholar provides broad access to leading computer science conferences and journals, while arXiv serves as a key platform for preprints across various disciplines, including a significant repository recognized by the computer science community. We employ a systematic search strategy on these platforms, using relevant keyword combinations to assemble our references.
We rigorously curate this collection, manually filtering out irrelevant papers and incorporating initially overlooked studies mentioned in their main texts. By exploiting Google Scholar's citation tracking, we thoroughly augment our list through iterative depth and breadth traversal.

%% file: 1.2-diff.tex
\subsection{Related Literature Reviews}
\input{tab/tab-related}
Several studies have reviewed literature pertinent to KGs and multi-modal learning. Distinct from these, our survey highlights specific differences, as shown in Table \ref{tab:differ}.

\begin{enumerate}[label={[\arabic*]}]
\item [1)] Zhu et al.~\cite{DBLP:journals/corr/abs-2202-05786} explore various characteristics of mainstream MMKGs and their constructions, primarily from a CV perspective. This include aspects like labeling images with KG symbols and symbol-image grounding.  Conversely, Peng et al.~\cite{DBLP:journals/bdr/PengHHY23} offer a detailed analysis of MMKG from a semantic web perspective, providing a definition and an analysis of its construction and ontology architectures. However, both studies present limited insights into tasks within and beyond MMKG, such as Multi-modal Entity Alignment (MMEA) and Multi-modal Knowledge Graph Completion (MKGC), potentially overlooking MMKG's inherent limitations. To fully grasp the challenges facing MMKG, extensive benchmarks and analyses across various academic and industrial tasks are necessary.

\item [2)] Monka et al.~\cite{DBLP:journals/semweb/MonkaHR22} provide an overview of Knowledge Graph Embedding (KGE) methods and their integration with high-dimensional visual embeddings, emphasizing the significance of KGs in visual information transfer. Lymperaiou et al.~\cite{DBLP:journals/corr/abs-2211-12328} discuss the enhancement of multi-modal learning with knowledge, aspiring to merge the realms of visual language representation and KGs. However, these studies primarily focus on KG's unilateral support to multi-modal tasks, overlooking the bidirectional feedback and co-evolution between KG and multi-modal methods. We advocate for a collaborative development of KG and multi-modal systems to overcome application  barriers and advance the journey towards Advanced General Intelligence (AGI).

\item [3)] The analyses by Zhu et al.~\cite{DBLP:journals/corr/abs-2202-05786} and Peng et al.~\cite{DBLP:journals/bdr/PengHHY23} are anchored in the developments up to 2021, leaving a gap in integrating the latest insights from the MMKG community. Similarly, the studies by Monka et al.~\cite{DBLP:journals/semweb/MonkaHR22} and Lymperaiou et al.~\cite{DBLP:journals/corr/abs-2211-12328} extend only up to 2021 and 2022, respectively. 
In response to the rapid advancements in AGI from 2022 to 2023, our survey explores the intricate relationship between MM4KG and KG4MM. It places a strong emphasis on emerging areas like LLMs, AI-for-Science, and industrial applications, aiming to fill critical knowledge gaps. Our goal is to provide a clear roadmap for future research and highlight the challenges and opportunities in these fast-evolving fields.
\end{enumerate}

%% file: tab/tab-related.tex
	


\begin{table*}[!htbp]
\scriptsize
\centering
\renewcommand{\arraystretch}{1.3}
\caption{\small Comparison of our survey with other related review papers on multi-modal learning and knowledge graphs. Abbreviations used: D.S. Tasks (Downstream Tasks), Const. (Construction), MLMPT (Multi-modal Language Model Pre-training), Industrial App. (Industrial Applications), 4 (for), Sci. (Science).}\label{tab:differ}
\resizebox{1.0\linewidth}{!}{
\begin{NiceTabular}{lccccccccccccc}
\CodeBefore
\rowcolor{gray!20}{1}
\rowcolor{gray!20}{2}
\rowcolor{gray!5}{7}
\Body
\toprule[0.8pt]
\multirow{2}{*}{\raisebox{-1ex}{\makebox[1.3cm][c]{\hspace{4em}\textbf{Survey Papers}}}} & \multicolumn{5}{c}{\textbf{KG4MM}} & \multicolumn{5}{c}{\textbf{MMKG}} & \multicolumn{3}{c}{\textbf{Challenges and Opportunities}} 
\\
\cmidrule(lr){2-6}
\cmidrule(lr){7-11}
\cmidrule(lr){12-14}
& {\scriptsize \textbf{KG Const.}} & {\scriptsize \textbf{D.S. Tasks}} & {\scriptsize \textbf{MLMPT}} & {\scriptsize \textbf{Benchmark}} & {\scriptsize \textbf{Industrial App.}} & {\scriptsize \textbf{MMKG Const.}} & {\scriptsize \textbf{D.S. Tasks}} & {\scriptsize \textbf{Benchmark}} & {\scriptsize \textbf{Industrial App.}} &  {\scriptsize \textbf{AI4Sci.}} &  {\scriptsize \textbf{KG4MM}} &  {\scriptsize \textbf{MMKG}} &  {\scriptsize \textbf{LLM}}     \\
\midrule[0.8pt]
Zhu et al.~\cite{DBLP:journals/corr/abs-2202-05786}              
& \xmark & \cmark & \xmark & \xmark & \xmark & \cmark & \cmark & \xmark & \xmark & \xmark & \xmark & \cmark & \xmark \\
Monka et al.~\cite{DBLP:journals/semweb/MonkaHR22}
& \xmark & \cmark & \xmark & \xmark & \xmark & \xmark & \xmark & \xmark & \xmark & \xmark & \cmark & \xmark & \xmark \\
Lymperaiou et al.~\cite{DBLP:journals/corr/abs-2211-12328}
& \cmark & \cmark  & \cmark & \xmark& \xmark & \xmark& \xmark & \xmark & \xmark & \xmark & \cmark & \xmark & \xmark \\
Peng et al.~\cite{DBLP:journals/bdr/PengHHY23}
& \xmark & \xmark & \xmark & \xmark & \xmark & \cmark & \cmark & \xmark & \xmark & \xmark & \xmark & \cmark & \xmark \\
\midrule
Ours              
& \cmark & \cmark & \cmark & \cmark & \cmark & \cmark & \cmark & \cmark & \cmark & \cmark & \cmark & \cmark & \cmark \\
\bottomrule[0.8pt]
\end{NiceTabular}
\vspace{-10pt}
}
\end{table*}

%% file: tab/param-tb.tex
\begin{table}[ht]
	\centering  
        \caption{Frequently Used Symbols.}
        \setlength{\tabcolsep}{2.5pt}
	\resizebox{0.97\linewidth}{!}{
	\begin{tabular}{lp{60mm}}
            \toprule[1.5pt]
            \midrule
            \textbf{Notations} & \textbf{Descriptions} \\
            \midrule
            $\mathcal{G}$ & Knowledge graph defined as  $\mathcal{G}=\{\mathcal{E}, \mathcal{R}, \mathcal{A}, \mathcal{T}, \mathcal{V}\}$. \\
            \midrule
            $ \mathcal{E}$ &  Entity set, including typical ($\mathcal{E}_{KG}$) and multi-modal entities ($\mathcal{E}_{MM}$).  \\
            \midrule
            $ \mathcal{R}$ & Set of relation predicates ($r$). \\
            \midrule
            $\mathcal{A}$ & Set of attribute predicates ($a$). \\
            \midrule
            $ \mathcal{T}$ & Statements set, comprising relational ($\mathcal{T_R}$) and attribute triples ($\mathcal{T_A}$). \\
            \midrule
            $\mathcal{V}$ & Attribute values set, including literals like string, date, integer, decimal ($\mathcal{V}_{KG}$) and multi-modal values ($\mathcal{V}_{MM}$). \\
            \midrule
            $\mathcal{I}$ & Set of visual images ($i$) in MMKGs. \\
            \midrule
            $(h, r, t)$ & Relational triple from $\mathcal{T_R}$ with head entity $h$ ($h \in \mathcal{E}$), tail entity $t$ ($t \in \mathcal{E}$), and relation predicate $r$. \\
            \midrule
            $(e, a, v)$ & Attribute triple from $\mathcal{T_A}$ with entity $e$, attribute predicate $a$ and value $v$. \\
            \midrule
            $< w_1, \dots, w_n>$ & Text corpus.  \\
            \midrule
            $\mathcal{X}$ & Input domain of multi-modal data across $K$ modalities, $\mathcal{X}=\mathcal{X}^{(1)}\times\cdots\times\mathcal{X}^{(K)}$ and $x^{(k)}\in\mathcal{X}^{(k)}$. \\
            \midrule
            $\mathcal{Y}$ & Target domain with $y\in\mathcal{Y}$. \\
            \midrule
            $\mathcal{D}$ & Data distribution for a downstream task. \\
            \midrule
            $\mathcal{Z}$ & Latent space with $z\in\mathcal{Z}$. \\
            \midrule
            $g_{\cdot}(\cdot)$ & Mapping function from the input domain (using all of $K$ modalities) to the latent space ($\mathcal{X}\mapsto\mathcal{Z}$).  \\
            \midrule
            $q_{\cdot}(\cdot)$ & Task mapping function from the latent space to the target domain ($\mathcal{Z}\mapsto\mathcal{Y}$). \\
        \midrule
        \bottomrule[1.5pt]
	\end{tabular} 
	\label{tab:param}
	\vspace{-10pt}
 }
\end{table}

%% file: 1.3-org.tex
\subsection{Article’s Organization}
\mbox{\S\,\ref{sec:pre}} introduces the preliminary, defining key concepts in KG and multi-modal Learning, and offering an overview of KG4MM and MM4KG settings. 
In \mbox{\S\,\ref{sec:kg-cst}}, we discuss the scope and construction of KGs, transitioning into the evolution of MMKGs. 
\mbox{\S\,\ref{sec:kg4mmtask}} delves into various KG4MM tasks, detailing the resources for each task and benchmarks advanced methods developed in the last three years, categorizing them into four paradigms: Understanding \& Reasoning; Classification; Content Generation; Retrieval; and Multi-modal Pre-training.
\mbox{\S\,\ref{sec:mm4kgtask}} reviews tasks within the MM4KG domain, categorizing key tasks into four areas: MMKG Acquisition, Fusion, Inference, and MMKG-driven Tasks.
While categorizing by specific tasks risks method overlap between KG4MM and MM4KG, it clarifies the research landscape and fosters interdisciplinary work and task integration. We carefully balance detail to address content overlaps across tasks, focusing on representative tasks.

We also analyze the current trends and industrial applications of KG4MM and MM4KG, providing insights into their impact across various sectors. 
Looking ahead, \mbox{\S\,\ref{sec:fut}} contemplates future integrations of multi-modal methods with (MM)KGs, proposing potential enhancements for previously discussed tasks.  It also explores the challenges and opportunities in sustaining KG4MM and MM4KG growth,  especially given the rapid development of LLMs and AI-for-Science applications.
Finally, \mbox{\S\,\ref{sec:conclusion}} concludes this article.

%% file: 2-pre.tex
\section{Preliminary}\label{sec:pre}
\begin{figure*}[!htbp]
  \centering
   \vspace{-1pt}
\includegraphics[width=0.99\linewidth]{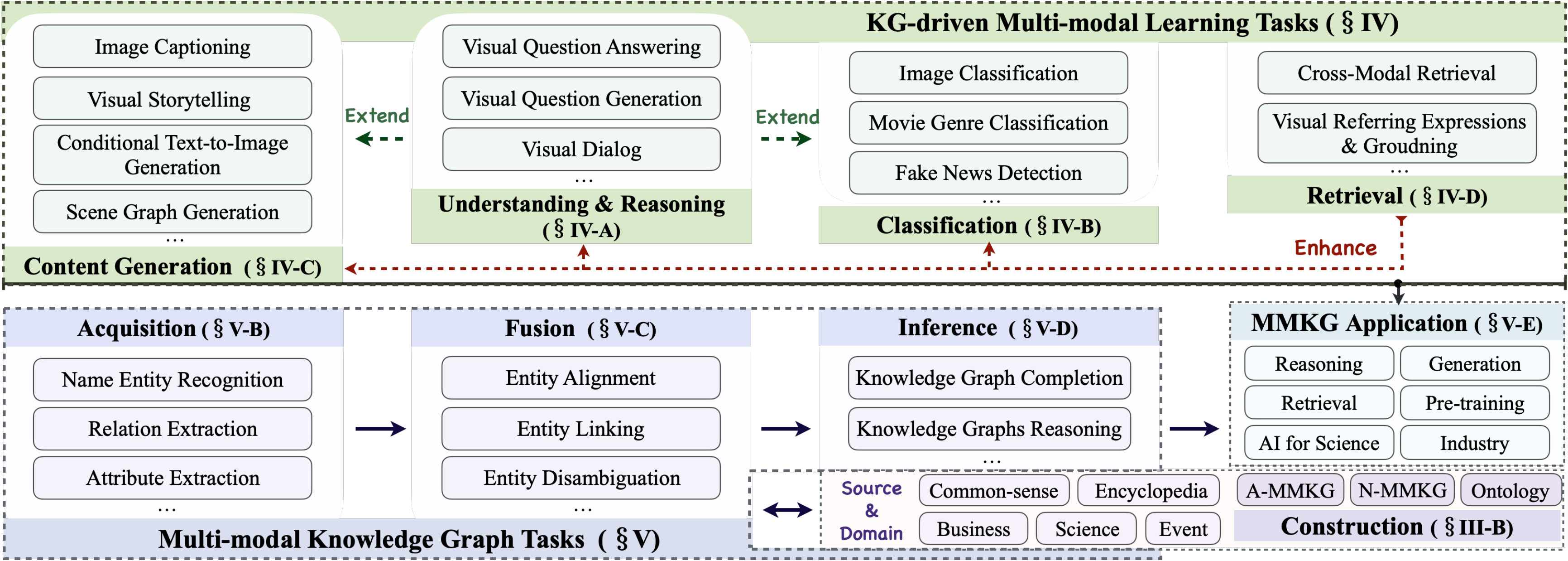}
  \caption{Comprehensive Overview of Integrating Knowledge Graphs with Multi-Modal Learning.
  }
  \label{fig:tasks}
  \vspace{-10pt}
\end{figure*}

\input{2.1-kg-pre}
\input{2.2-mm-pre}
\input{2.3-kg-mm}

\input{2.4-mm-kg}

%% file: 2.1-kg-pre.tex
\subsection{Knowledge Graph}\label{sec:kg-pre}
Since their inception around 2007, Knowledge Graphs (KGs) have become pivotal in various academic domains, marked by foundational projects such as Yago \cite{DBLP:conf/www/SuchanekKW07}, DBPedia \cite{DBLP:conf/semweb/AuerBKLCI07}, and Freebase \cite{DBLP:conf/sigmod/BollackerEPST08}. The integration of Google's Knowledge Panels into web search in 2012 highlighted a significant milestone in the adoption of KGs. These KGs now play an essential role in enhancing search engines like Google and Bing and are integral to the functionality of voice assistants like Amazon Alexa and Apple Siri, reflecting their widespread business importance and increasing prevalence.

\textbf{Structural Composition:}
KGs represent real-world entities and relationships using a graph structure, where nodes symbolize  real-world entities or atomic values (attributes), and edges denote relations. Knowledge in KGs is often captured in triples, such as \textit{(Hangzhou, locatedAt, China)}. They utilize an ontology-based schema, discussed in \mbox{\S\,\ref{sec:onto}}, to define basic entity classes and their relationships,  usually in a taxonomic structure. This semi-structured nature merges structured data's clear semantics (from ontologies) with the flexibility of unstructured data, allowing easy expansion through new classes and relations.

\textbf{Accessibility and Advantages:}
KGs support a wide array of downstream applications,
accessible primarily via \textit{Lookup} and \textit{Querying} methods.

\textit{Lookup} in KGs, also known as KG retrieval, identifies relevant entities or properties based on input strings, leveraging lexical indices (surface) from entity and relation labels. An example of this is the DBpedia online lookup service\footnote{\url{https://lookup.dbpedia.org/}} \cite{DBLP:conf/semweb/AuerBKLCI07}.

Alternatively, \textit{Querying} returns results from input queries crafted in the RDF query language SPARQL\footnote{\url{https://www.w3.org/TR/rdf-sparql-query/}}. These queries typically involve sub-graph patterns with variables, yielding matched entities, properties, literals, or complete sub-graphs.

Note that KGs, especially those with OWL ontologies, support symbolic reasoning, including consistency checks to identify logical conflicts and entailment reasoning to infer hidden knowledge via Description Logics.
KGs also facilitate inter-domain connections. An example is the linkage between the \textit{Movie} and \textit{Music} domains through common entities like individuals who are both \textit{actors} and \textit{singers}. This interconnectivity not only enhances machine comprehension but also improves human understanding, benefiting applications like search, question answering, and recommendations.
Furthermore, recent developments in LLMs highlight the crucial role of KGs, particularly in managing long-tailed knowledge, as evident in several studies \cite{DBLP:journals/corr/abs-2308-14217,DBLP:journals/corr/abs-2308-10168,DBLP:journals/corr/abs-2308-06374,DBLP:journals/corr/abs-2306-08302}.

\textit{1)\textbf{~Formulation:}}
Aiming to align with established literature, we first introduce a widely-accepted definition of KG and its foundational operations, followed by an exploration of KGs enriched with ontologies from the semantic web perspective. We conclude with an expansive view on the diverse interpretations and uses of KGs beyond the semantic web scope.

\begin{defn}{\textbf{Knowledge Graph.}}\label{def:kg}
 A Knowledge Graph (KG) is denoted as $\mathcal{G}=\{\mathcal{E}, \mathcal{R}, \mathcal{T}\}$, comprising an entity set $\mathcal{E}$, a relation set $\mathcal{R}$, and a statement set $\mathcal{T}$. A statement is either a relational fact triple $(h, r, t)$ or an attribute triple $(e, a, v)$. Specifically, KGs consist of a set of relational facts forming a multi-relational graph, wherein nodes represent entities ($h$ and $t$ in $\mathcal{E}$ symbolize head and tail entities, respectively) and edges are denoted by relations ($r \in \mathcal{R}$). Regarding attribute triples, the attribute $a$ ($a \in \mathcal{A}$) indicates that an entity $e$ has a certain attribute with a corresponding value $v$ ($v \in \mathcal{V}$). These values can include various literals, such as strings or dates, and cover metadata like labels and textual definitions, represented through either built-in or custom annotation properties.
\end{defn}

\textit{2)\textbf{~Ontology:}}\label{sec:onto}
Within the semantic web, ontologies serve as KG schemas, utilizing languages like RDFS\footnote{RDF Schema, \url{https://www.w3.org/TR/rdf-schema/}} and OWL\footnote{Web Ontology Language, \url{https://www.w3.org/TR/owl2-overview/}} to ensure richer semantics and superior quality \cite{horrocks2008ontologies}.
Key features of ontologies include:
\begin{itemize}
     \item Hierarchical classes, often termed as concepts\footnote{To distinguish between \textit{class} in machine learning tasks and \textit{class} in KGs, we refer to the latter as \textit{concept}.}.
    \item Properties that specify the terms used in relationships.
    \item Hierarchies involving both concepts and relations.
    \item Constraints, including the domain and range of relations, as well as class disjointness.
    \item Logical expressions that encompass relation composition.
\end{itemize}

Languages like RDF, RDFS, and OWL introduce built-in vocabularies to capture these knowledge elements, 
with predicates like \textit{rdfs:subClassOf} denoting concept subsumption, and \textit{rdf:type} indicating instance-concept associations. Additionally, RDFS provides annotation properties like \textit{rdfs:label} and \textit{rdfs:comment} for providing meta-information about resources.

\textit{3)\textbf{~KG Scope Extension:}}
Widely accepted KGs include WordNet~\cite{miller1995wordnet}, a lexical database defining word interrelations, and ConceptNet~\cite{speer2017conceptnet}, which archives commonsense knowledge interlinked by different terms.
In this paper, we extend the conventional view of KGs beyond standard-format entities and relations. 
Besides, the ontology alone, often utilized to define domain knowledge including conceptualization and vocabularies like terms and taxonomies, is also considered a form of KG. Further elaborating on this expanded perspective, as outlined by Chen et al. \cite{DBLP:journals/pieee/ChenGCPHZHC23}, our scope includes simpler graph structures, such as basic taxonomies with hierarchical classes and graphs with weighted edges denoting quantitative relationships like similarity and distance between entities. Additionally, we categorize any structured data organized in a graph format with nodes that have explicit semantic interpretations as part of this broader KG definition.  A prominent example is the Semantic Network, which connects various concepts with labeled edges to represent different relationships.

%% file: 2.2-mm-pre.tex
\subsection{Multi-modal Learning}\label{sec:mm-pre}
Our world is perceived through diverse modalities, including sight, sound, movement, touch, and smell \cite{DBLP:journals/alife/SmithG05}.  It is intuitive that models, which integrate data from various modalities, generally surpass uni-modal models by accumulating more information.
A ``modality'' typically refers to a specific type of data or information channel, characterized by sensory input or representation format. Each modality, such as visual, auditory, textual inputs, encapsulates unique features from specific sensory sources or data acquisition methods. Multi-modal learning typically aims to develop a unified representation or mapping from multiple modalities to an output space, leveraging the complementarity and redundancy across modalities to improve prediction. The challenge lies in effectively aligning, fusing, and integrating information from various modalities to cohesively exploit their collective power.

\textit{1)\textbf{~Difference to Multi-view Learning:}}
Unlike multi-view analysis, which suggests that each view (e.g., different perspectives of a flower) can independently yield accurate predictions \cite{DBLP:journals/corr/abs-1304-5634,DBLP:conf/iclr/Federici0FKA20}, multi-modal learning contends with scenarios where the absence of one modality could impede task completion \cite{DBLP:conf/nips/HuangDXCZH21} (e.g., an image-lacking Visual Question Answering scenario). 
Additionally, multi-view learning typically involves varying perspectives of the same data type, originating from a single source, such as different features of image data. In contrast, multi-modal learning deals with disparate data types, like text and images, derived from multiple sources. In this paper, our exploration of multi-modal tasks and the application of multi-modal learning on KGs are grounded in this broader understanding of multi-modal learning.

\begin{defn}{\textbf{Multi-modal Learning.}}\label{def:mm}
Assume given data \(\hat{x}=\left(x^{(1)},\ldots, x^{(K)}\right)\) consists of \(K\) modalities, with \(x^{(k)}\in \mathcal{X}^{(k)}\) representing the domain set of the \(k\)-th modality and \(\mathcal{X}=\mathcal{X}^{(1)}\times\cdots\times\mathcal{X}^{(K)}\). 
Let \(\mathcal{Y}\) denote the target domain and \(\mathcal{Z}\) represent a latent space. Denote \(g:\mathcal{X}\mapsto\mathcal{Z}\) as the true mapping from the input space (utilizing all \(K\) modalities) to the latent space, and  \(q:\mathcal{Z}\mapsto\mathcal{Y}\) as the true task mapping.
For example, in aggregation-based multi-modal fusion, \(g\) serves as an aggregation function built upon \(K\) separate sub-networks, and \(q\) is a multi-layer neural network \cite{DBLP:conf/nips/WangHSXRH20}. In a learning task, a data pair \((\hat{x},y)\in\mathcal{X}\times \mathcal{Y}\) is generated from an unknown distribution \(\mathcal{D}\), such that
\begin{equation}
    \mathbbm{P}_{\mathcal{D}}(\hat{x},y) = \mathbbm{P}_{y \mid \hat{x}}\left(y \mid q \circ g(\hat{x})\right)\mathbbm{P}_{\hat{x}}(\hat{x})\,,\label{dis}
\end{equation}
  where \(q \circ g(\hat{x}) = q(g(\hat{x}))\) represents the composite function of \(q\) and \(g\). 
\end{defn}

\textit{2)\textbf{~Scope of Multi-modal Settings:}}
This paper primarily focuses on visiolinguistic (VL) tasks that involve text and image data, aiming to provide in-depth analysis and maintain continuity in related research. While certain special cases may briefly touch upon modalities like video or those specific to the biochemistry domain, these are comparatively less prevalent in our study.
With a focus on two modalities (i.e, language and vision), the input domain simplifies to $\mathcal{X}=\mathcal{X}^{\mathbbm{l}}\times\mathcal{X}^{\mathbbm{v}}$ and $\hat{x}=(x^\mathbbm{l},x^\mathbbm{v})$, where $x^\mathbbm{l}\in\mathcal{X}^\mathbbm{l}$ and $x^\mathbbm{v}\in\mathcal{X}^\mathbbm{v}$  denote input data from the language and visual domains, respectively. Tasks involving other special modalities will be separately addressed at relevant sections.

%% file: 2.3-kg-mm.tex
\subsection{KG-driven Multi-modal Setting}\label{sec:kg-mm}
Generally, KGs serve as repositories for various knowledge types such as domain-specific and commonsense knowledge, which have been widely applied in multi-modal scenarios.

\begin{figure*}[!htbp]
  \centering
   \vspace{-1pt}
\includegraphics[width=1.0\linewidth]{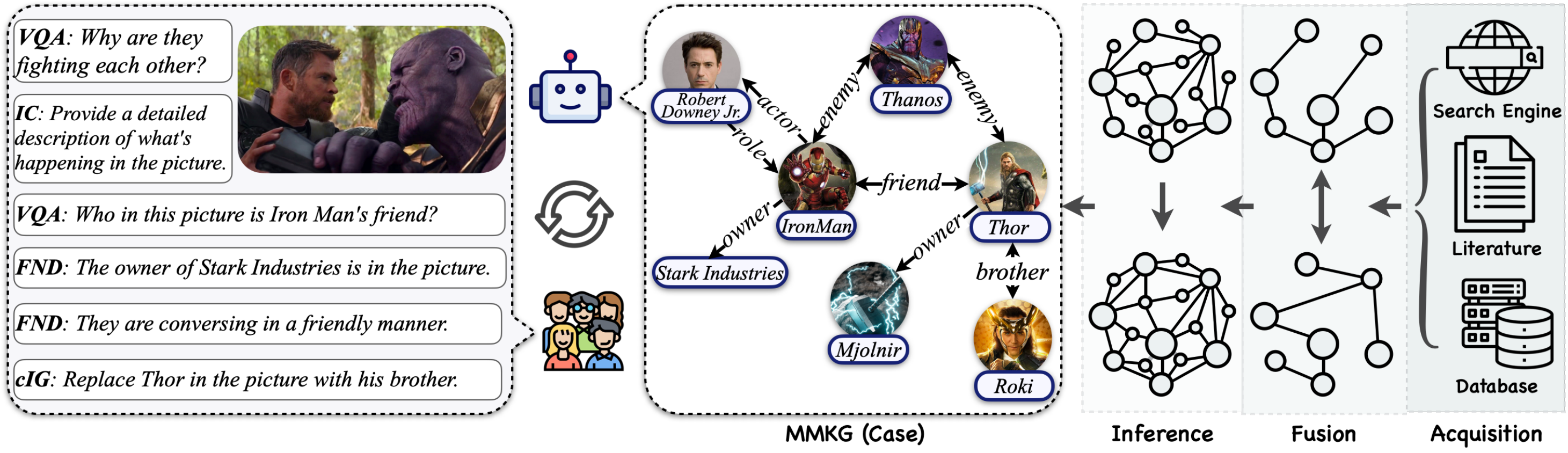}
  \caption{Roadmap for Multi-Modal Knowledge Graph (MMKG) construction and application in downstream multi-modal tasks.
  }
  \label{fig:roadmap}
  \vspace{-10pt}
\end{figure*}

\textit{1)\textbf{~Sub-KG Extraction:}} 
Practical applications often require tapping into localized knowledge to address specific tasks effectively. An straightforward solution involves integrating sub-KG extraction (isolating minimal knowledge units or triplets from a large KG like WordNet \cite{miller1995wordnet}) with downstream tasks to reduce noise from irrelevant information. This typically requires algorithms for retrieval, routing, or semantic parsing.

\textit{2)\textbf{~Task-Oriented KG Construction:}}
In certain KG-driven multi-modal studies, researchers construct task-specific KGs from scratch rather than using existing KGs or sub-KG extraction. This approach, tailored to the unique requirements of each task, usually involves creating a KG directly from datasets or by combining multiple KGs, and generally aligns with one of two dominant paradigms:
\textbf{\textit{(\rmnum{1})}} \textbf{Static Domain KGs Construction}: 
This technique involves creating stable, domain-specific KGs with pre-defined entities and relations, encapsulating crucial background knowledge. 
Its necessity becomes apparent in two primary scenarios related to the limitations of general KGs:
\ul{When a General KG Lacks Adequate Knowledge for a Specific Task}:
In this case, crafting a tailored background KG becomes imperative.
An illustrative example is Zero-shot Image Classification tasks, which necessitate building KGs that capture visual attributes or taxonomy associations \cite{hu2021graph,zhao2020knowledge,DBLP:conf/www/GengC0PYYJC21,pahuja2023bringing}. These KGs are designed to cover all relevant classification knowledge, serving as static domain knowledge bases. Typically, textual data, such as class labels, are utilized to delineate class relationships, aiding in the formation of KG edges.
\ul{When a General KG Only Partially Covers a Task}:
This scenario requires the precise selection and reorganization of the existing KG to fill the gaps adequately.
For instance, in knowledge-aware Visual Question Answering \cite{DBLP:journals/pami/WangWSDH18,DBLP:conf/cvpr/MarinoRFM19}, questions may require both commonsense and encyclopedic knowledge.  Some works \cite{DBLP:conf/ijcai/ZhuYWS0W20,DBLP:conf/cvpr/MarinoCP0R21,DBLP:conf/semweb/0007CGPYC21,DBLP:conf/jist/0007HCGFP0Z22} selectively extract RDF facts from various KGs, such as ConceptNet \cite{speer2017conceptnet}, WebChild \cite{DBLP:conf/wsdm/TandonMSW14}, and DBpedia \cite{DBLP:conf/semweb/AuerBKLCI07,DBLP:conf/acl/TandonMW17},  creating a unified background KG to augment the models.
\textbf{\textit{(\rmnum{2})}} \textbf{Dynamic Temporary KGs Construction}: This technique centers on building dynamic, temporary KGs during task execution, leveraging KG reasoning algorithms for task support. 
For instance, 
establishing co-occurrence relations between classes (e.g., food ingredients) involves analyzing their frequency in training datasets, along with common class attributes and hierarchies.  Further, Li et al.~\cite{DBLP:journals/tmm/LiJ19} employ scene graphs to combine visual and semantic information from images, deriving semantic relationship features from entity triples for Image Captioning.

%% file: 2.4-mm-kg.tex
\subsection{Multi-modal Knowledge Graph Setting}\label{sec:mm-kg}
The evolution of AI gradually exposes the limitations of traditional uni-modal (text-based) KGs in handling the variety of multi-modal applications on the Internet. 
This has spurred both academic and industrial research to develop Multi-modal Knowledge Graphs (MMKGs).

\textit{1)\textbf{~~MMKG Scope:}} 
A KG qualifies as multi-modal (MMKG) when it contains knowledge symbols expressed in multiple modalities, which can include, but are not limited to, text, images, sound, or video. 
This survey distinguishes between two MMKG representation methods, A-MMKG and N-MMKG, as inspired by Zhu et al. \cite{DBLP:journals/corr/abs-2202-05786}, where A-MMKGs treat images as entity attributes, and N-MMKGs allow images to stand as independent entities with direct relationships.

\begin{defn}{\textbf{Multi-modal Knowledge Graph.}}\label{def:mmkg}
Consistent with Definition \ref{def:kg}, a KG is defined as $\mathcal{G}=\{\mathcal{E}, \mathcal{R}, \mathcal{A}, \mathcal{T}, \mathcal{V}\}$ where $\mathcal{T} = \{\mathcal{T_A}, \mathcal{T_R}\}$ with $\mathcal{T_R}=\mathcal{E}\times\mathcal{R}\times\mathcal{E}$ and $\mathcal{T_A}=\mathcal{E}\times\mathcal{A}\times\mathcal{V}$. 
\textbf{\textit{(\rmnum{1})}}~\textbf{A-MMKG} utilizes multi-modal data (e.g., images) as specific attribute values for entities or concepts, with $\mathcal{T_A}=\mathcal{E}\times\mathcal{A}\times(\mathcal{V}_{KG}\cup\mathcal{V}_{MM})$, where $\mathcal{V}_{KG}$ and $\mathcal{V}_{MM}$ are values of KG and multi-modal data, respectively.
\textbf{\textit{(\rmnum{2})}}~\textbf{N-MMKG} treats multi-modal data as KG entities, with $\mathcal{T_R}=(\mathcal{E}_{KG}\cup\mathcal{E}_{MM})\times\mathcal{R}\times(\mathcal{E}_{KG}\cup\mathcal{E}_{MM})$, separating typical KG entities ($\mathcal{E}_{KG}$) from multi-modal entities ($\mathcal{E}_{MM}$).
\end{defn}

For instance, in N-MMKG, a relation triple $(h, r, t)$ in $\mathcal{T_R}$ may include $h$ or $t$ as an image, with $r$ defining the relationship. Conversely, in A-MMKG, an attribute triple $(e, a, v)$ in $\mathcal{T_A}$ might associates an image as $v$ with the attribute $a$, typically designated as \textit{hasImage}.
Note that N-MMKG and A-MMKG are not strictly exclusive: N-MMKG might be considered a particular case of A-MMKG, especially when an entity in A-MMKG takes the form of an image, thereby transforming it into N-MMKG.
Given the convenience in data access and similarity to traditional KGs, A-MMKG forms the basis for most current applications and methodologies in MMKG research, as elaborated in \mbox{\S\,\ref{sec:mmkga}} and \mbox{\S\,\ref{sec:mmkgc}}. 

%% file: 3-kg-cst.tex
\section{Knowledge Graph Construction}\label{sec:kg-cst}

\input{3.1-norm-kg}
\input{3.2-mm-kg}

%% file: 3.1-norm-kg.tex
\subsection{Typical KG Construction}\label{sec:norm-kg-cst}
In this paper we categorize typical KGs into two fundamental types~\cite{DBLP:journals/corr/abs-2308-14217}: entity-based KGs and text-rich KGs.

\textit{1)\textbf{~Entity-based KGs:}}
When constructing entity-based KGs, both ontology and data adhere to strict standards, wherein KG nodes typically represent entities in a one-to-one correspondence with real-world objects.
These KGs are prominent in both academic projects like Yago~\cite{DBLP:conf/www/SuchanekKW07} and Freebase~\cite{DBLP:conf/sigmod/BollackerEPST08}, and industry initiatives like OpenBG~\cite{DBLP:journals/corr/abs-2308-14217} and TeleKG~\cite{DBLP:conf/icde/00070HCGYBZYSWY23}. 
%
These KGs are typically built upon manually defined ontologies, offering clear semantics with minimal ambiguity and overlap in entity types and relationships. This results in a relatively low number of entities and relationships per domain, making them manageable for manual definition. For example, Freebase identifies only 52 entity types and 155 relationships within its \textit{Movie} domain. 
The construction of these KGs often involves processing entities and relationships from structured sources like relational databases. Wikipedia~\cite{denoyer2006wikipedia}, with its entity descriptions and hyperlinks between entity pages, serves as a common starting point for knowledge acquisition. Early KGs like Yago, DBPedia~\cite{DBLP:conf/semweb/AuerBKLCI07}, and Freebase benefit from the high accuracy of Wikipedia data by transforming Infoboxes into entities and relationships. Additional sources, such as IMDb, MusicBrainz, and Goodreads, enhance coverage, especially for entities of varying popularity.

Integrating knowledge from various structured sources requires tackling three heterogeneity types \cite{DBLP:journals/corr/abs-2308-14217}: \textbf{\textit{(\rmnum{1})} Schema Heterogeneity}, where different data sources may represent the same entity type and relationship differently; \textbf{\textit{(\rmnum{2})} Entity Heterogeneity}, where varied source names might depict the same real-world entity; \textbf{\textit{(\rmnum{3})} Value Heterogeneity}, where different sources may offer dissimilar or outdated attribute values for identical entities.
Addressing these issues has spurred numerous research tasks, including Entity Linking in incomplete KG and data fusion (e.g., KG Completion and Entity Alignment) across diverse KGs.
Besides, techniques for extending KG content include extracting knowledge from semi-structured data, such as websites. Here, each page typically represents a topic entity, and information is displayed in key-value pairs, consistently positioned across different pages. These techniques aim to capture long-tail knowledge, often using manually constructed extraction patterns and supervised extraction algorithms. 

\textit{2)\textbf{~Text-rich KGs:}}
Unlike entity-based KGs, text-rich KGs, with their dominant text attributes, face challenges in extracting clean, unambiguous entities, making them more akin to bipartite graphs than to conventional connected graphs.
Typically, they tolerate greater ambiguities, representing nodes as free texts rather than well-defined entities, making them particularly suited to domains like Products and Encyclopedia where semantic distinctions between values and classes are often unclear \cite{DBLP:journals/tacl/WangGZZLLT21}. The construction of text-rich KGs, especially in domains without a specialized structured knowledge base like Wikipedia, generally depends on extraction models. These models extract structural information from relevant, unstructured source data, employing Named Entity Recognition methods to identify patterns indicative of specific attributes.

%% file: 3.2-mm-kg.tex
 \subsection{MMKG Construction}\label{sec:mm-kg-cst}

\textit{1)\textbf{~Paradigms for MMKG Construction:}}
We outline two principal paradigms for MMKG construction following Zhu et al.~\cite{zhu2022multi}: \textbf{\textit{(\rmnum{1})} Annotating Images with Symbols from a KG} and \textbf{\textit{(\rmnum{2})} Grounding KG Symbols to Images}.

The first paradigm prioritizes the extraction of visual entities/concepts, relations, and events, which are crucial for the dynamic creation of KGs like scene and event graphs \cite{DBLP:conf/acl/MaWLCCLSDWSS22}. 
This approach, however, encounters challenges in representing infrequent (i.e., long-tail) multi-modal knowledge, primarily due to the recurrent depiction of common real-world entities across diverse contexts. The use of supervised methods further compounds these challenges, as they are inherently constrained by the finite scope of pre-existing labels.  Moreover, this system demands substantial pre-processing, including the formulation of specific rules, the creation of predetermined entity lists, and the application of pre-trained detectors and classifiers, all of which pose significant scalability challenges~\cite{DBLP:conf/acl/LiZLPWCWJCVNF20,DBLP:conf/iccv/ChenSG13}.
Further exploration of these issues and their implications for MMKG construction is detailed in \mbox{\S\,\ref{sec:mmkge}}.

Besides, the typical construction paradigm for most of the current MMKG is grounding KG symbols to images, which involves: entity grounding (i.e., associating entities with corresponding images from online sources \cite{DBLP:conf/akbc/Onoro-RubioNGGL19}), concept grounding (i.e., selecting diverse, representative images for visual concepts and abstracting common visual features), and relation grounding (i.e., choosing images that semantically mirror the relation of the input triples). This paradigm currently poses the principal challenge in large-scale MMKG construction.

\input{tab/mmkg-sta}
\textit{2)\textbf{~Evolution Process:}}
This section explores the development of MMKGs, with Table~\ref{tab:mmkg-sta} providing statistics for various MMKGs.
Notably, the earliest MMKG in a general sense could be traced back to \textbf{ImageNet} (2009) \cite{DBLP:conf/cvpr/DengDSLL009}, a large-scale image ontology based on the WordNet \cite{miller1995wordnet} structure. 
Despite its rich semantic hierarchy and millions of annotated images, ImageNet, as an A-MMKG, is primarily utilized for object classification, with its knowledge components often underutilized.
\textbf{NEIL} (2013)~\cite{DBLP:conf/iccv/ChenSG13} represents an early effort to construct visual knowledge from the Internet through a cycle of relationship extraction, data labeling, and classifiers/detectors learning. 
However, NEIL's scalability is limited, demonstrated by its intensive computational requirement to classify 400K visual instances of 2273 objects, whereas typical KGs require grounding billions of instances.
Further developments \cite{DBLP:conf/cvpr/ChenSG14,DBLP:conf/cvpr/JohnsonKSLSBL15,DBLP:conf/naacl/YatskarOF16,DBLP:conf/ijcai/GongW17,DBLP:conf/eccv/LuKBL16} focus on improving visual detection and object segmentation from complex images, with Chen et al. \cite{DBLP:conf/cvpr/ChenSG14} leveraging learned top-down segmentation priors from visual subcategories to aid in the construction.

\textbf{Visual Genome} (2016) \cite{DBLP:journals/ijcv/KrishnaZGJHKCKL17} provides dense annotations of objects, attributes, and relationships.
However, it primarily aids scene understanding tasks like image description and question answering.
\textbf{ImageGraph} (2017) \cite{DBLP:conf/akbc/Onoro-RubioNGGL19}, rooted in Freebase \cite{DBLP:conf/sigmod/BollackerEPST08}, and \textbf{IMGpedia} (2017) \cite{DBLP:conf/semweb/FerradaBH17}, linking Wikimedia Commons\footnote{\url{http://commons.wikimedia.org}} visual data with DBpedia metadata, represent further expansions into MMKGs. ImageGraph, assembled through a web crawler parsing image search results and applying heuristic data cleaning rules (e.g., deduplication and ranking), focuses on reasoning over visual concepts, enabling relation prediction and multi-relational image retrieval. As an N-MMKG, IMGpedia, with its emphasis on visual descriptors and similarity relations, supports visual-semantic queries but is limited by its scope of commonsense and encyclopedic knowledge.

In 2019, Liu et al. \cite{DBLP:conf/esws/LiuLGNOR19} first formally introduced the term \textbf{``MMKG''}, launching three A-MMKG datasets for Link Prediction and Entity Matching research, constructed using a web crawler as the image collector based on Freebase15K (FB15K) \cite{DBLP:conf/nips/BordesUGWY13}, averaging 55.8 images per entity. 
Meanwhile, DBpedia15k (DBP15K) and Yago15k (YG15K) were developed by aligning entities from DBpedia and Yago with FB15K, enriching these KGs with numeric literals, image information, and \textit{sameAs} predicates for cross-KG Entity Linking.

\textbf{GAIA} (2020) \cite{DBLP:conf/acl/LiZLPWCWJCVNF20} is an MMKG extraction system that supports complex graph queries and multimedia information retrieval. It integrates Text Knowledge Extraction and Visual Knowledge Extraction processes on identical document sets, generating modality-specific KGs which are then merged into a coherent MMKG. Concurrently, \textbf{VisualSem} \cite{DBLP:journals/corr/abs-2008-09150} emerges as an A-MMKG, sourcing entities and images from BabelNet \cite{DBLP:journals/ai/NavigliP12} with meticulous filtering to ensure data quality and diversity. Entities in VisualSem are linked to Wikipedia, WordNet synsets \cite{miller1995wordnet}, and, when available, high-resolution images from ImageNet \cite{DBLP:conf/cvpr/DengDSLL009}.
As a N-MMKG, \textbf{Richpedia} \cite{DBLP:journals/bdr/WangWQZ20} 
collects images and descriptions from Wikipedia \cite{DBLP:journals/cacm/VrandecicK14}, using hyperlinks and text for manual relationship identification among image entities, supplemented by a web crawler for broader image entity collection.

Recent focus in the MMKG community has shifted from construction to application, emphasizing areas such as MMKG Representation Learning (\mbox{\S\,\ref{sec:mmkgr}}), Acquisition (\mbox{\S\,\ref{sec:mmkge}}), Fusion (\mbox{\S\,\ref{sec:mmkga}}), Inference (\mbox{\S\,\ref{sec:mmkgc}}), and MMKG-driven Applications (\mbox{\S\,\ref{sec:mmapp}}). 
While MMKG acquisition extends construction efforts, it mainly addresses multi-modal extraction challenges~\cite{DBLP:conf/acl/MaWLCCLSDWSS22}, highlighting the scarcity of large-scale MMKG resources and the demand for task-specific datasets to address MMKG's limitations and support novel downstream tasks.

Specifically,
Baumgartner et al. \cite{DBLP:conf/mm/BaumgartnerRB20} employ multi-modal detectors and a semantic web-informed scheme for semantic relation extraction between movie characters and locations to support Deep Video Understanding.
Peng et al. \cite{DBLP:conf/apweb/PengXTWH22}  explore image quality control in MMKG construction through an Image Refining Framework that uses clustering for de-duplication and noise reduction, taps into Wikidata for entity descriptions, and relies on a pre-trained model to gauge image-text similarity. It discards images that do not meet a specific relevance threshold.
In MMKG construction, accurately aligning concepts with their corresponding images is crucial~\cite{DBLP:conf/dasfaa/JiangLLLXWLX22,DBLP:journals/bdr/PengHHY23}.
The challenge arises from distinguishing between visualizable concepts (VCs), like ``\textit{dog}'', which have clear visual representations, and non-visualizable concepts (NVCs), such as ``\textit{mind}'' or ``\textit{texture}'', which lack direct visual counterparts. This distinction complicates their inclusion in MMKGs.
Jiang et al. \cite{DBLP:conf/dasfaa/JiangLLLXWLX22}  introduce a visual concept classifier that identifies VCs and NVCs, utilizing ImageNet instances to exemplify the former. However, this binary classification serves merely as a preliminary phase. The deeper challenge in MMKG construction lies in selecting representative images for entities, potentially involving clustering methods like K-means or spectral clustering \cite{zhu2022multi}. 
Building upon this, Zhang et al. \cite{DBLP:conf/cikm/ZhangWWLX23} introduce \textbf{AspectMMKG},
enriching MMKGs by associating entities with aspect-specific images, derived from Wikipedia, and refining image selection with a trained model.
Besides, Wu et al. \cite{wu2023mmpedia} present \textbf{MMpedia}, a scalable, high-quality MMKG constructed via a novel pipeline that leverages DBpedia \cite{DBLP:conf/semweb/AuerBKLCI07} to filter non-visualizable entities and refine entity-related images using textual and type information.

Gong et al.~\cite{DBLP:journals/corr/abs-2302-06891} introduce \textbf{UKnow} (2023), a unified knowledge protocol that classifies N-MMKG triples into five unit types: in-image, in-text, cross-image, cross-text, and image-text. They establish an efficient pipeline for adapting existing datasets to UKnow's format, facilitating the automated generation of new datasets from existing image-text pairs. 
Further, Zha et al. \cite{zha2023m2conceptbase} propose a framework to construct a multi-modal conceptual MMKG, named \textbf{M\textsuperscript{2}ConceptBase}. Initially, they extract candidate concepts from textual descriptions in image-text pairs and refine them using rule-based filters.  These concepts are then aligned with corresponding images and detailed descriptions through context-aware multi-modal symbol grounding. For concepts not fully grounded, GPT-3.5-Turbo generates supplementary descriptions. 
Note that the nodes in M\textsuperscript{2}ConceptBase and AspectMMKG are not linked or mapped to existing public KGs. Instead, their focus is on decomposing entity concepts and associating them with fine-grained images. As a result, most nodes within these MMKGs remain isolated, rendering the graphs more akin to multi-modal extensions of text-rich KGs, as discussed in \mbox{\S\,\ref{sec:norm-kg-cst}}.
Song et al. \cite{yaoxian2023scenedriven} unveil a scene-driven MMKG construction method that starts with natural language scene descriptions and employs a prompt-based scene-oriented schema generation. This approach, combined with traditional knowledge engineering and LLMs, streamlines the creation and refinement of the \textbf{ManipMob-MMKG}, a specialized MMKG tailored for indoor robotic tasks such as manipulation and mobility.

Exploring MMKGs' utility in downstream tasks, Xu et al. \cite{DBLP:conf/mm/Xu0WZC22} introduce two MMKG Link Prediction datasets: \textbf{MKG-W} and \textbf{MKG-Y}. Derived from OpenEA benchmarks \cite{DBLP:journals/pvldb/SunZHWCAL20}, these datasets integrate structured data from Wikipedia and YAGO with expert-validated entity images sourced from the web. Wang et al.~\cite{wang2023tiva} further investigate the role of different modalities in Link Prediction through \textbf{TIVA-KG}, an MMKG covering text, image, video, and audio modalities. Built upon the foundation of ConceptNet \cite{speer2017conceptnet}, TIVA-KG also supports \textbf{triplet grounding} (i.e., associating a common sense triplet with tangible representations like images), enabling the mapping of symbolic knowledge within its multi-modal framework.
Similarly, Lee et al.~\cite{DBLP:conf/emnlp/LeeCLJW23-VISITA} propose \textbf{VTKGs}, where both the entities and triplets are attached with images, with accompanying textual descriptions for each entity and relation.
Focusing on Multi-modal Entity Alignment tasks, Li et al. \cite{DBLP:journals/corr/abs-2302-08774} introduce \textbf{Multi-OpenEA}, extending the OpenEA benchmarks with 16 MMKGs and Google-sourced images. To study the effect of missing visual modality on MMKG representation and alignment, Chen et al.~\cite{chen2023rethinking} randomly removed images from the DBP15K \cite{DBLP:conf/aaai/0001CRC21} and Multi-OpenEA datasets \cite{DBLP:journals/corr/abs-2302-08774}, releasing the \textbf{MMEA-UMVM} datasets.
Additionally, Zhang et al. \cite{DBLP:conf/iclr/000100LDC23} define a new task on multi-modal analogical reasoning over KGs, which requires the ability to reason using multiple modalities and background knowledge. They also develop a dataset, MARS, and a corresponding MMKG, \textbf{MarKG}, for benchmarking purposes.

\begin{figure*}[!htbp]
  \centering
   \vspace{-1pt}
\includegraphics[width=1.0\linewidth]{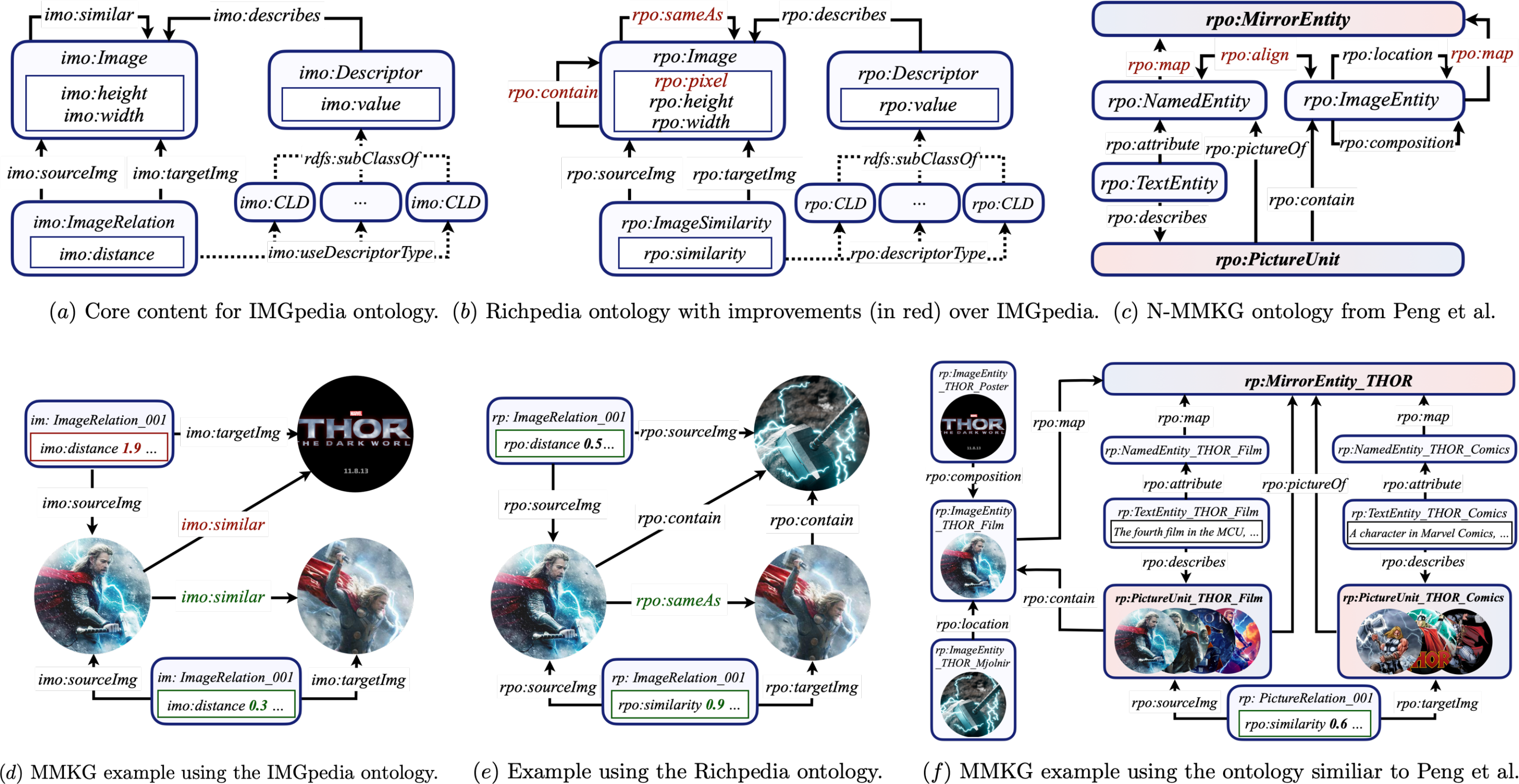}
  \caption{Currently representative N-MMKG ontologies and corresponding MMKG examples using those ontologies.
  }
  \label{fig:mmkgonto}
  \vspace{-10pt}
\end{figure*}

\textit{3)\textbf{~N-MMKG Ontology:}} 
Considering that the A-MMKG ontology largely mirrors standard KGs, with the primary distinction being the inclusion of visual attributes, we mainly discuss several representative N-MMKG ontologies in this part. 
This emphasis is due to the complex design considerations involved in integrating image entities into N-MMKGs.

URI prefixes play a pivotal role in ontologies, serving as unique  identifiers for classes and properties. Standard prefixes (e.g., \textit{rdf}, \textit{rdfs}, \textit{owl}) ensure cross-domain consistency, while custom ones (e.g., \textit{imo} for IMGpedia and \textit{rpo} for Richpedia) bring in domain-specific nuances. This setting not only optimizes data representation and interoperability in KGs but also aligns with broader RDF standards, laying the foundation for MMKGs which facilitates  efficient querying and intricate manipulation of diverse multimedia data. 

{IMGpedia Ontology}~\cite{DBLP:conf/semweb/FerradaBH17} (Fig.~\ref{fig:mmkgonto}(a)) extends terms from the DBpedia Ontology\footnote{\url{schema.org}} and the Open Graph Protocol to represent multi-modal data in RDF.
Specifically, the \textit{imo:Image} denotes an abstract resource representing an image, which captures its dimensions (\textit{imo:height}, \textit{imo:width}), URL (\textit{imo:fileURL}), and an \textit{owl:sameAs} link to its corresponding resource in DBpedia Commons. \textit{imo:Descriptor} defines visual descriptors linked via \textit{imo:describes}, with types including \textit{imo:HOG} (Histogram of Oriented Gradient), \textit{imo:CLD} (Color Layout Descriptor), and \textit{imo:GHD} (Gradation Histogram Descriptor). \textit{imo:ImageRelation} encapsulates similarity links between images, detailing the descriptor type used and the Manhattan distance between image descriptors, with an additional \textit{imo:similar} relation for k-nearest neighbor ($k$-nn) images.

Richpedia ontology~\cite{DBLP:journals/bdr/WangWQZ20} (Fig.~\ref{fig:mmkgonto}(b)) aligns closely with the IMGpedia Ontology. Here, \textit{rpo:KGEntity} denotes textual KG entities, while \textit{rpo:Image} stands for a Richpedia image entity characterized by a URL and dimensions (e.g., \textit{rpo:Height} and \textit{rpo:Width}, both expressed in the \textit{xsd:float} datatype for numerical values).
Subclasses of \textit{rpo:Descriptor}, like \textit{rpo:GHD}, capture visual traits of images. Semantic relations like \textit{rpo:sameAs} and \textit{rpo:imageOf} link these entities, with \textit{rpo:ImageSimilarity} quantifying image likeness between \textit{rpo:sourceImage} and \textit{rpo:targetImage} through pixel-level comparisons.

Following Richpedia~\cite{DBLP:journals/bdr/WangWQZ20}, Peng et al. \cite{DBLP:journals/bdr/PengHHY23} explore a new MMKG ontology (Fig.~\ref{fig:mmkgonto}(c)) to tackle the issue of entities with multiple visual representations (i.e., aspects), a phenomenon emphasized by AspectMMKG \cite{DBLP:conf/cikm/ZhangWWLX23} and M\textsuperscript{2}ConceptBase~\cite{zha2023m2conceptbase}. The key of this paradigm is to introduce the \textit{Mirror Entity} and \textit{Picture Unit} as foundational concepts. \textit{rpo:MirrorEntity} denotes a particular concept, with \textit{rpo:NamedEntity} pointing to a related KG entity. Its visual counterpart, the \textit{rpo:ImageEntity}, is sourced from the \textit{rpo:PictureUnit}, which might aggregate multiple such image entities under the same aspect. Besides, various \textit{rpo:PictureUnit} maintain a degree of similarity through \textit{rpo:similarity}.An \textit{rpo:align} linkage is established when \textit{rpo:NamedEntity} and \textit{rpo:ImageEntity} both reference a common \textit{rpo:MirrorEntity}. Further, the \textit{rpo:pictureOf} relation binds \textit{rpo:PictureUnit} to \textit{rpo:NamedEntity}, with the \textit{rpo:TextEntity} serving as a bridge, encapsulating shared descriptions.
In essence, this ontology enriches the prior MMKG by offering a hierarchical structure, effectively clustering and associating images from diverse aspects.

Fig.~\ref{fig:mmkgonto} visualizes the evolutionary trajectory of MMKG ontologies, highlighting the unique challenges N-MMKGs face: \textbf{\textit{(\rmnum{1})}}  An individual entity may have multiple visual representations (i.e., varied aspects). \textbf{\textit{(\rmnum{2})}}  Efficiently extracting information from visual modalities across entities is crucial. \textbf{\textit{(\rmnum{3})}} The development of distinct multi-modal representation methods can extend beyond entity-level to include relation and triple-level representations, as explored in works like \cite{wang2023tiva,DBLP:conf/emnlp/LeeCLJW23-VISITA}.  Discussions on future directions continue in \mbox{\S\,\ref{sec:fut-mmkg}}.

%% file: tab/mmkg-sta.tex
\begin{table*}[t]
\caption{Overview of various MMKGs, detailing their publish (Pub.) time, types, scale, data sources, and supported (Sup.) tasks, where symbol $\ast$ indicates the inclusion of triple-level multi-modal information within the MMKG. Not that only part of the Sup. tasks are listed that have been experimentally validated in original studies, although MMKGs have a wider potential task range. 
The key distinctions among nodes, entities, and concepts are based on their representation: entities typically correspond directly to real-world object names, nodes include both these entities and textual elements~\cite{DBLP:journals/corr/abs-2008-09150} like Wikipedia articles (detailed in \mbox{\S\,\ref{sec:norm-kg-cst}}), and  concept is a further decomposition of entity where each entity has multiple concepts, corresponding to different aspects such as ``\textit{culture}'', ``\textit{geography}'', and ``\textit{history}''~\cite{DBLP:conf/cikm/ZhangWWLX23,zha2023m2conceptbase}.
Besides, this table primarily lists MMKGs in general visual multi-modal scenarios, excluding other event-based or domain-specific MMKGs like ManipMob-MMKG~\cite{yaoxian2023scenedriven}, which focuses on indoor scenes. Abbreviations used: \textbf{Data source}: CN (ConceptNet); DBP (DBpedia); Freebase (FB); VG (VisualGenome); WP (Wikipedia); WN (WordNet); WD (Wikidata); Wikimedia (WM); Web Search Engine (WSE); YG (YAGO). \textbf{Tasks}: Image Classification (IMGC); Cross-Modal Retrieval (CMR); Object Detection (OD); Scene Graph Generation (SGG); Visual Question Anwering (VQA);  Concept Understanding (CU); Multi-modal Knowledge Graph Completion (MKGC), Knowledge Graph Reasoning (MKGR), Entity Alignment (MMEA), Entity Linking (MMEL) and Information Extraction (MMIE). } %
\centering
\resizebox{0.97\textwidth}{!}{
\begin{NiceTabular}{cccccc}
\CodeBefore
  \rowcolor{gray!40}{1}
  \rowcolors{2}{gray!15}{white}
\Body
\toprule[0.8pt]
\multicolumn{1}{c}{\textbf{Pub. Time}} & \multicolumn{1}{c}{\textbf{MMKGs}} & \multicolumn{1}{c}{\textbf{Types}} & \multicolumn{1}{c}{\textbf{Scale (\#nodes / \#images)}}  & \multicolumn{1}{c}{\textbf{Data Sources}} & \multicolumn{1}{c}{\textbf{Sup. Tasks}} \\ 
\midrule[0.8pt]
2013-12 & NEIL~\cite{DBLP:conf/iccv/ChenSG13} & N-MMKG & 1152 (classes) / 300K  & WN / Image WSE & OD, etc. \\
2014-09 & ImageNet~\cite{russakovsky2015imagenet} & A-MMKG & 21K (classes) / 3.2M  & WN / Image WSE & IMGC, OD, etc. \\
2016-02 & VisualGenome~\cite{DBLP:journals/ijcv/KrishnaZGJHKCKL17} & A-MMKG & 35 (classes) / 108K  & WN / MS COCO /  YFCC~\cite{DBLP:journals/cacm/ThomeeSFENPBL16} & SGG, VQA, etc. \\
2016-09 & WN9-IMG~\cite{DBLP:conf/ijcai/XieLLS17-IKRL} & A-MMKG & 6.5K (entities) / 14K  & WN / ImageNet & MKGC \\ 
2017-01 & ImageGraph~\cite{DBLP:journals/tip/LiuWZT17} & A-MMKG & 15K (entities) / 837K  & FB / Image WSE & CMR \\
2017-10 & IMGpedia~\cite{DBLP:conf/semweb/FerradaBH17} & N-MMKG & 2.6M (entities) / 15M  & DBP / WM Commons  & CMR \\
2019-03 & MMKG~\cite{DBLP:conf/esws/LiuLGNOR19} & A-MMKG & 45K (entities) / 37K  & FB / DBP / YG / Image WSE & MMEA, MKGC \\
2020-07 & GAIA~\cite{DBLP:conf/acl/LiZLPWCWJCVNF20} & N-MMKG & 457K (entities) / NA  & FB / GeoNames / News Websites & MMIE \\
2020-08 & VisualSem~\cite{DBLP:journals/corr/abs-2008-09150} & N-MMKG & 90K (nodes) / 938K   & WP / WN / ImageNet & CMR \\
2020-09 & DBP-DWY-Vis~\cite{DBLP:conf/aaai/0001CRC21} & A-MMKG & 178K (entities) / 117K  & WP / DBP15k \cite{DBLP:conf/semweb/SunHL17} / DWY15K \cite{DBLP:conf/icml/GuoSH19} & MMEA \\
2020-12 & Richpedia~\cite{DBLP:journals/bdr/WangWQZ20} & N-MMKG & 2.8M (entities) / 2.9M  & WD / WM / Image WSE & MMKG Querying \\ 
2021-06 & RESIN~\cite{DBLP:conf/naacl/WenLLPLLZLWZYDW21} & N-MMKG & 51K (events) / NA  & WD / News Websites & MMIE \\
2022-10 & MKG-W\&Y~\cite{DBLP:conf/mm/Xu0WZC22} & A-MMKG & 30K (entities) / 29K  & OpenEA~\cite{DBLP:journals/pvldb/SunZHWCAL20} / Image WSE & MKGC \\ 
2022-10 & MarKG~\cite{DBLP:conf/iclr/000100LDC23} & A-MMKG & 11K (entities) / 76K  & WD / Image WSE & MKGR \\
2023-02 & Multi-OpenEA~\cite{DBLP:journals/corr/abs-2302-08774} & A-MMKG & 920K (entities) / 2.7M  & OpenEA / Image WSE  & MMEA \\ 
2023-03 & UKnow~\cite{DBLP:journals/corr/abs-2302-06891} & N-MMKG & 1.4M (entities) / 1.1M  & WP / Image WSE  & MKGC, CMR \\ 
2023-07 & UMVM~\cite{chen2023rethinking} & A-MMKG & 238K (entities) / 205K  & DBP-DWY-Vis / Multi-OpenEA & MMEA \\ 
2023-08 & AspectMMKG~\cite{DBLP:conf/cikm/ZhangWWLX23} & A-MMKG & 2.3K (entities) / 645K  & WP / Image WSE & MMEL \\ 
2023-10 & TIVA-KG~\cite{wang2023tiva} & A-MMKG$\ast$ & 440K (entities) / 1.7M  & CN / Image WSE & MKGC \\ 
2023-11 & MMpedia~\cite{wu2023mmpedia} & A-MMKG & 2.7M (entities) / 19.5M  & DBP / Image WSE & MKGC \\ 
\rowcolor{white}2023-12 & VTKGs~\cite{DBLP:conf/emnlp/LeeCLJW23-VISITA} & A-MMKG$\ast$ & 43K (entities) / 460K & \makecell{\rowcolor{white}CN / WN / UnRel~\cite{DBLP:conf/iccv/PeyreLSS17} / VRD~\cite{DBLP:conf/eccv/LuKBL16} \\ \rowcolor{white}HICO-DET~\cite{DBLP:conf/wacv/ChaoLLZD18} / VisKE~\cite{DBLP:conf/cvpr/SadeghiDF15}}
 & MKGC \\
2023-12 & {M\textsuperscript{2}ConceptBase}~\cite{zha2023m2conceptbase} & A-MMKG & 152K (concepts) / 951K & \makecell{\rowcolor{gray!25}Wukong~\cite{DBLP:conf/nips/GuMLHMLYHZJXX22} / Baidu Encyclopedia} & VQA, CU \\
\bottomrule[0.8pt]
\end{NiceTabular}
}
\label{tab:mmkg-sta}
\vspace{-7pt}
\end{table*}

%% file: 4-kg-mm-task.tex
\section{KG-driven Multi-modal Learning Tasks}\label{sec:kg4mmtask}
This section explores the role of KGs in enhancing multi-modal learning tasks. 
As pivotal symbolic knowledge carriers, KGs underpin diverse tasks that require rich background knowledge, including but not limited to Generation, Reasoning, Understanding, Classification, Retrieval, and Pre-training.  By presenting a systematic taxonomy within a unified framework, we clarify the methods' core aspects to enhance domain comprehension and guide future research.

\input{4.2-reason}

\input{4.3-cls}
\input{4.1-gen}
\input{4.4-ret}
\input{4.5-plm}

%% file: 4.2-reason.tex
\subsection{Understanding \& Reasoning  Tasks}\label{sec:kgr}
Multi-modal reasoning tasks, like knowledge-based Visual Question Answering (VQA)~\cite{DBLP:conf/cvpr/MarinoRFM19,DBLP:journals/pami/WangWSDH18}, Visual Commonsense Reasoning (VCR), Visual Question Generation (VQG), Visual Dialog (VD), and Multi-modal Sarcasm Explanation (MuSE)~\cite{DBLP:conf/acl/JingSOJN23}, demand knowledge that goes beyond regular daily experiences~\cite{DBLP:journals/internet/KhanBC22}.  These tasks often delve into  less common, long-tail knowledge domains that typically require intentional learning or reflection, with KGs providing a crucial structured repository for this extensive, specialized knowledge.

\begin{defn}{\textbf{KG-aware Understanding \& Reasoning.}}\label{def:kgr}
Aligning with previous Definition \ref{def:kg} and \ref{def:mm}, a KG is denoted as $\mathcal{G}={\mathcal{E}, \mathcal{R}, \mathcal{T}}$ with $\mathcal{T} = \{\mathcal{T_A}, \mathcal{T_R}\}$. 
Given a image-question pair ($x^\mathbbm{v}$, $x^\mathbbm{l}$), the goal is to derive an answer $y$, utilizing the background KG ($\mathcal{G}$) as a foundational support. 
\end{defn}


\subsubsection{Visual Question Answering}\label{sec:kgvqa}
\begin{figure}[!htbp]
  \centering
   \vspace{-1pt}
\includegraphics[width=0.85\linewidth]{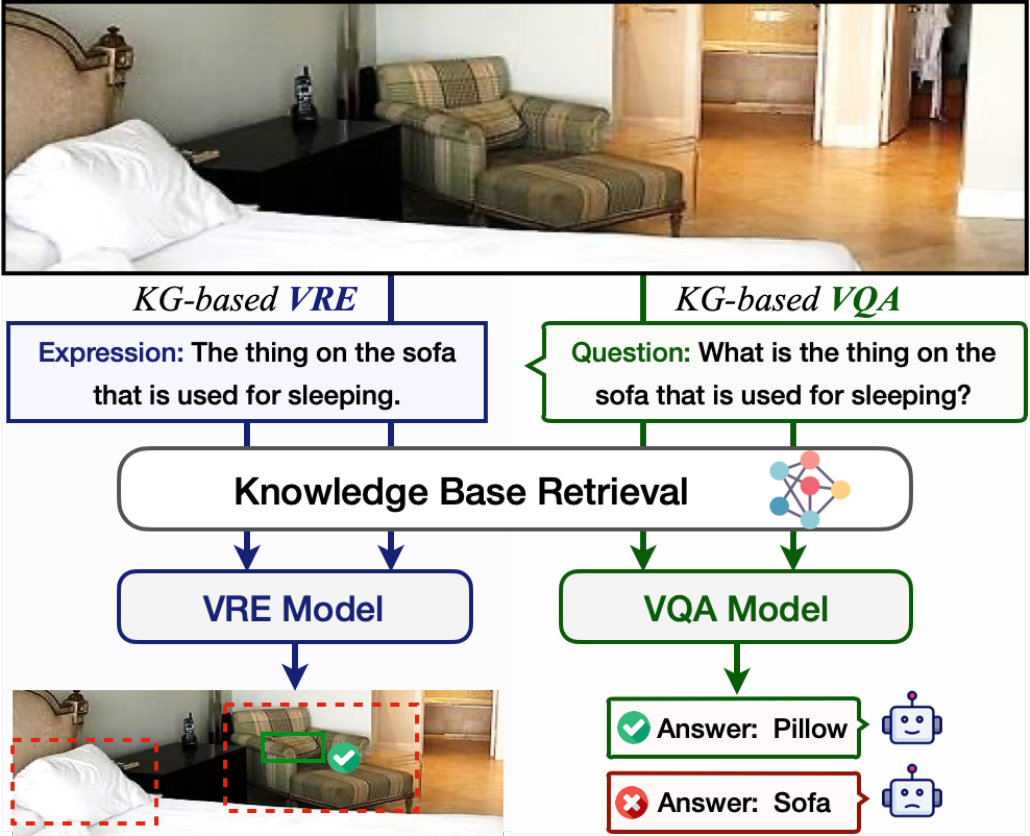}
\caption{Illustration of KG-based Visual Question Answering (VQA) (\mbox{\S\,\ref{sec:kgr}}) and Visual Referring Expressions (VRE) (\mbox{\S\,\ref{sec:kgret}}). To some extent, KG-based VRE can be viewed as an extension of KG-based VQA, incorporating an additional step of grounding answers.}
  \label{fig:vqavre}
  \vspace{-6pt}
\end{figure}

VQA~\cite{DBLP:conf/iccv/AntolALMBZP15,DBLP:journals/corr/abs-2311-00308} is a cornerstone task in multi-modal learning, serving as a benchmark for evaluating the capabilities of most multi-modal models~\cite{DBLP:conf/nips/LuBPL19,DBLP:conf/nips/AlayracDLMBHLMM22,DBLP:conf/icml/0008LSH23} due to its straightforward task definition and relevance to everyday scenarios. 
KG-based VQA (Fig.~\ref{fig:vqavre}), emerging around 2015~\cite{DBLP:conf/cvpr/WuWSDH16}, diverges from traditional approaches by integrating an external Knowledge Base (KB) for more complex question analysis and deeper reasoning assistance~\cite{DBLP:journals/pami/WangWSDH18}.

\begin{figure*}[!htbp]
  \centering
   \vspace{-1pt}
\includegraphics[width=0.95\linewidth]{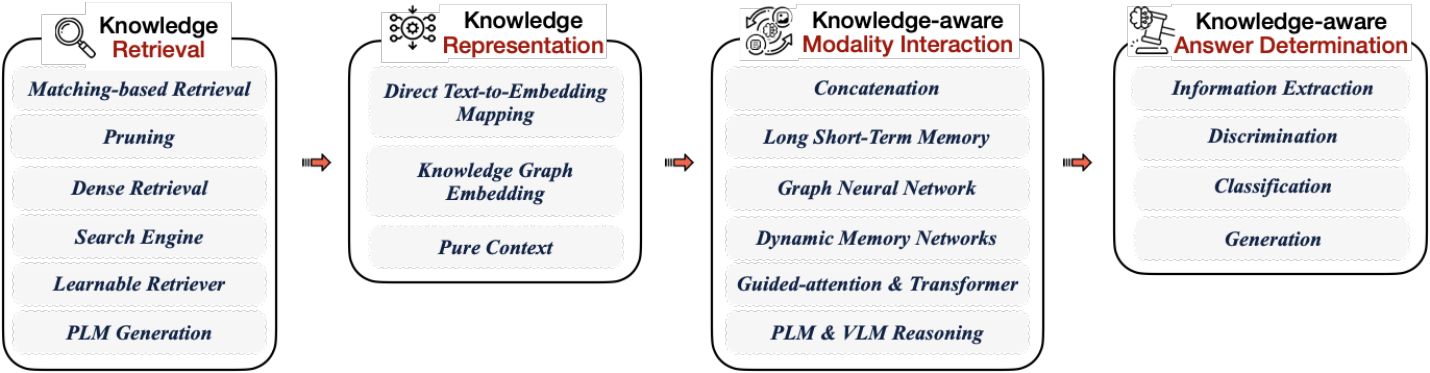}
  \caption{Current KG-aware Understanding \& Reasoning research pipeline, which typically involves four key stages for incorporating knowledge. Note that studies often employ one or more of these stages.
  }
  \label{fig:kg4mmr}
  \vspace{-8pt}
\end{figure*}

\textbf{METHODS:} As illustrated in Fig.~\ref{fig:kg4mmr}, current KG-aware VQA research typically involves four key stages for incorporating knowledge: \textit{Knowledge Retrieval}, \textit{Knowledge Representation}, \textit{Knowledge-aware Modality Interaction}, and \textit{Knowledge-aware Answer Determination}. These stages, integral to the workflow of KG-aware understanding and reasoning tasks, may be adopted individually or in combination across different studies to form a comprehensive approach.

The KG-based VQA process can be expressed as follows:
\begin{equation}
p(A|Q,I,\mathcal{G},\mathbf{\Theta}) =  \underbrace{p(\mathcal{G}_{ret}|Q, I, \mathcal{G}; \mathbf{\Phi})}_{\text{Retriever (if have)}} \cdot \underbrace{p(A|Q, I, \mathcal{G}_{ret}; \mathbf{\Theta})}_{\text{Reader}},
\end{equation}
where $Q$, $I$, $A$ represent the textual question ($x^\mathbbm{l}$), image ($x^\mathbbm{v}$), and answer ($y$), respectively.
$\mathcal{G}$ and $\mathcal{G}_{ret}$ denote the overall background KG and the retrieved relevant sub-KG, while $\mathbf{\Phi}$ refers to the  model parameters  used in the knowledge retrieval step. 
Given that implicit knowledge can be encoded within $\mathbf{\Theta}$, typically pre-trained on large-scale datasets for specialized self-supervised tasks, the use of a retriever becomes optional, yet it remains advantageous for task completion.

\includegraphics[scale=0.018]{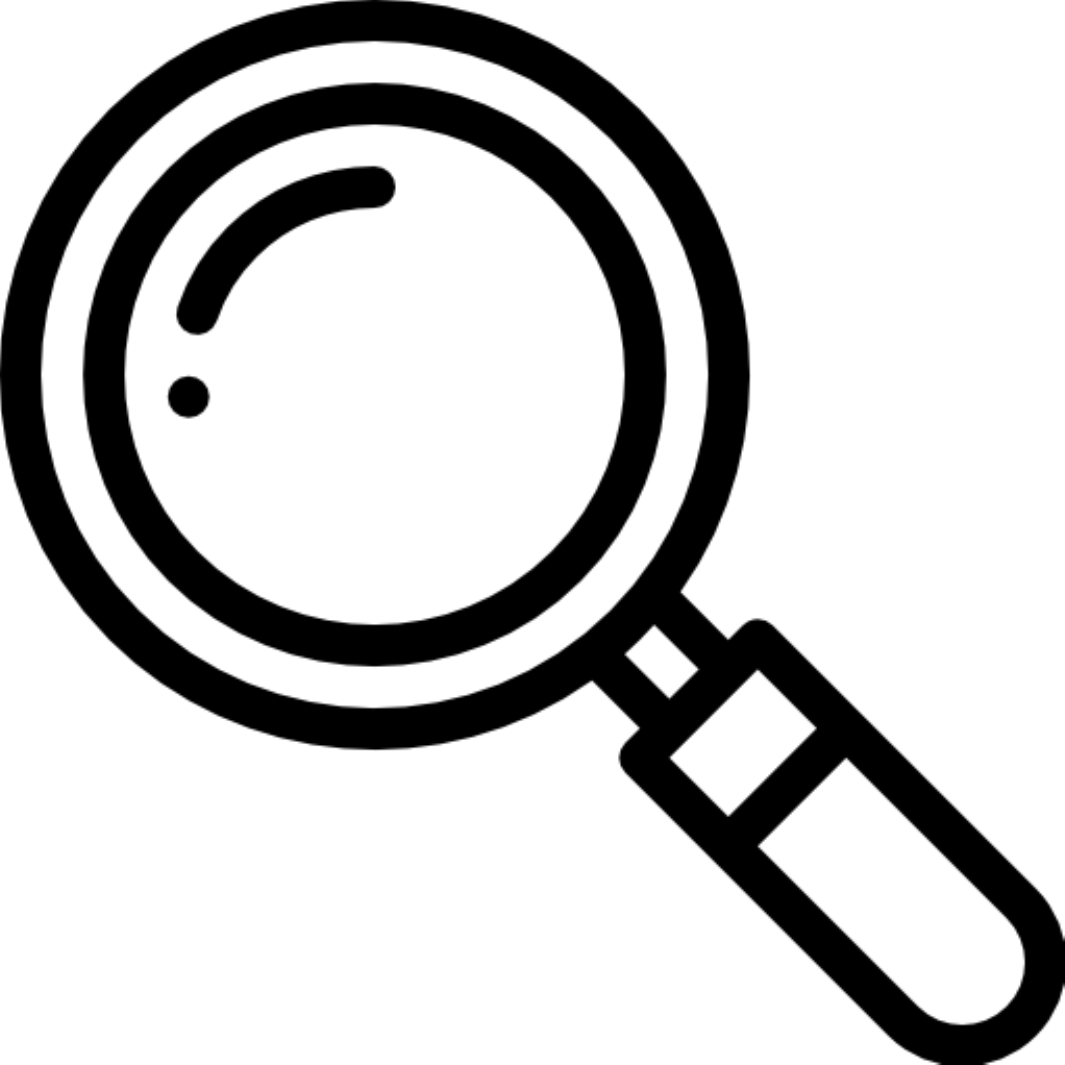}\;\textbf{\ul{Knowledge Retrieval}} 
is crucial for integrating knowledge with multi-modal reasoning tasks, focusing on extracting pertinent knowledge from various external sources, including not only KGs but also non-KG organizations like document collections (e.g., Wikipedia~\cite{denoyer2006wikipedia}).
These techniques has evolved from early matching-based and dense embedding similarity approaches to learnable retrieval and Pre-trained Language Model (PLM) generation techniques, broadening the scope and efficiency of knowledge integration.

\textbf{\textit{(\rmnum{1})} Matching-based Retrieval} generally employs methods such as RDF query, Entity Linking, and BM25 to identify key concepts within images and questions, linking these to relevant data in extensive large-scale KBs like ConceptNet~\cite{speer2017conceptnet}. 
The extraction process from images might involve identifying spatial positions~\cite{DBLP:conf/aaai/ShahMYT19,DBLP:conf/ijcai/ZhuYWS0W20,DBLP:journals/pr/YuZWZHT20,DBLP:conf/emnlp/GarderesZAL20,DBLP:conf/cvpr/MarinoCP0R21,DBLP:conf/nips/LinX0X0Y22},  visual object sizes and names~\cite{DBLP:conf/ijcai/WangWSDH17,DBLP:journals/pami/WangWSDH18,DBLP:conf/eccv/NarasimhanS18,DBLP:conf/ecir/WangLT19,DBLP:conf/nips/NarasimhanLS18,DBLP:conf/ijcai/ZhuYWS0W20,DBLP:conf/coling/ZiaeefardL20,DBLP:journals/pr/YuZWZHT20,DBLP:conf/mm/Li0020,DBLP:journals/corr/abs-2012-15484,DBLP:journals/tnn/ZhangLLZLSG21,DBLP:conf/emnlp/GarderesZAL20,DBLP:journals/pr/ZhengYCML021,DBLP:conf/acl/VickersAMB20,DBLP:conf/aaai/LiM22,DBLP:journals/corr/abs-2202-04306,DBLP:conf/cvpr/DingYLHC022,DBLP:conf/sitis/HussainMSF22,DBLP:journals/tip/HanYWWN23,DBLP:journals/pr/SongLLYSS23,DBLP:conf/wacv/RaviCSLS23,DBLP:conf/icmcs/YouYLL23,DBLP:conf/emnlp/KhademiYFZ23,yin2023multi,dong2024modalityaware}, 
high-level attributes like scene names, object parts, and human activities, using various pre-trained classifiers or APIs\footnote{\url{https://azure.microsoft.com/en-us/products/cognitive-services/vision-services}}
\cite{DBLP:conf/ijcai/WangWSDH17,DBLP:conf/cvpr/WuWSDH16,DBLP:conf/eccv/NarasimhanS18,DBLP:journals/pami/WangWSDH18,DBLP:conf/nips/NarasimhanLS18,DBLP:journals/pr/YuZWZHT20,DBLP:journals/corr/abs-2012-15484,DBLP:conf/cvpr/MarinoCP0R21,DBLP:conf/sitis/HussainMSF22,DBLP:journals/corr/abs-2202-04306,DBLP:conf/emnlp/LinB22,DBLP:conf/eacl/HeW23,DBLP:journals/pr/SongLLYSS23,DBLP:conf/icmcs/YouYLL23,DBLP:conf/emnlp/KhademiYFZ23}.
Additionally, image captions and OCR text strings can be generated for information supplement
\cite{DBLP:conf/cvpr/WuWSDH16,DBLP:conf/cvpr/SuZDCCL18,DBLP:conf/ijcai/ZhuYWS0W20,DBLP:journals/pr/YuZWZHT20,DBLP:journals/eswa/SalaberriaALSA23,DBLP:conf/jist/0007HCGFP0Z22,DBLP:conf/sitis/HussainMSF22,DBLP:conf/naacl/GuiWH0BG22,DBLP:conf/emnlp/LinB22,DBLP:conf/emnlp/WuM22,DBLP:conf/nips/LinX0X0Y22,DBLP:conf/icmcs/YouYLL23,DBLP:journals/corr/abs-2308-15851,DBLP:conf/acl/SiMLJW23,DBLP:conf/emnlp/KhademiYFZ23,dong2024modalityaware}.
And those question and captions can be parsed by NLP tools (e.g., NLTK~\cite{DBLP:books/daglib/0022921}, AllenNLP constituency parser~\cite{DBLP:journals/corr/abs-1803-07640}, Stanza~\cite{DBLP:conf/acl/QiZZBM20}, NLP Dependency Parser~\cite{DBLP:conf/emnlp/ChenM14}, Named Entity Recognizer~\cite{DBLP:conf/acl/FinkelGM05}, LLMs~\cite{dong2024modalityaware}) for syntax analysis~\cite{DBLP:conf/ecir/WangLT19,DBLP:conf/mm/Li0020,DBLP:conf/nips/SaqurN20,DBLP:journals/tnn/CaoLLWL22,DBLP:journals/tip/HanYWWN23,DBLP:conf/emnlp/WuM22,DBLP:conf/wacv/RaviCSLS23}, along with techniques like regular expressions (regex)~\cite{DBLP:conf/ijcai/WangWSDH17}, semantic graph parsing model~\cite{DBLP:conf/ijcai/ZhuYWS0W20,DBLP:journals/pr/YuZWZHT20,DBLP:conf/aaai/WuLSM22,DBLP:conf/eacl/HeW23,DBLP:conf/sitis/HussainMSF22}, SpanSelector~\cite{DBLP:conf/sigir/JainKKJRC21}, or pre-trained classiﬁer for query template selection~\cite{DBLP:journals/pami/WangWSDH18,DBLP:conf/eccv/NarasimhanS18,DBLP:conf/nips/NarasimhanLS18}.
During this stage, unimportant visual objects not present in the question or caption might be filtered out~\cite{DBLP:journals/tnn/ZhangLLZLSG21,DBLP:journals/corr/abs-1803-07640}.

After extracting initial concepts from $Q$ and $I$, two key mappings are established: the first links parsed objects in $Q$ to their visual counterparts in $I$, and the second associates these concepts with relevant entries in KBs. 
This is achieved using methods like greedy longest-string matching
\cite{DBLP:conf/ijcai/WangWSDH17,DBLP:conf/cvpr/SuZDCCL18,DBLP:journals/corr/abs-2101-06013}, template matching~\cite{DBLP:journals/pami/WangWSDH18}, and multi-modal entity linking methods~\cite{DBLP:journals/pr/ZhengYCML021,DBLP:conf/sigir/JainKKJRC21,DBLP:conf/emnlp/WuM22,DBLP:conf/icmcs/YouYLL23,DBLP:conf/mir/AdjaliGFGB23}.  Techniques like face identification algorithms ~\cite{DBLP:conf/aaai/ShahMYT19,DBLP:conf/acl/VickersAMB20,DBLP:conf/acl/HeoKCZ22,DBLP:conf/sigir/LernerFGBB0L22} and ViLBERT-multi-task~\cite{DBLP:conf/cvpr/LuGRPL20} serve as effective tools for linking objects.
Following this, fact triples can be collected by involving the first-order sub-KG from these identified concept nodes (sometimes will be three-hops in character KG~\cite{DBLP:conf/aaai/ShahMYT19}) or by identifying brief knowledge paths among the entities from $I$ and $Q$. This process requires constructing a temporary sub-KG specific to the current $Q$-$I$ pair~\cite{DBLP:conf/ecir/WangLT19,DBLP:conf/cvpr/SuZDCCL18,DBLP:conf/mm/Li0020}.
In addition to the construction of local sub-KG for individual $Q$-$I$ sample, KG-Aug~\cite{DBLP:conf/mm/Li0020} constructs a global sub-KG that links $Q$, $I$, and candidate answers in a unified knowledge-based semantic space.
Besides, KAN~\cite{DBLP:journals/tnn/ZhangLLZLSG21} presents a weighting system for each fact to indicate the reliability of the corresponding knowledge piece.
Heo et al.~\cite{DBLP:conf/acl/HeoKCZ22} develop a hypergraph from the KG, using a triplet as the basic unit to preserve the higher-order semantics inherent in the KG.

RDF query generation, such as SPARQL, often involves filling pre-defined templates with parsed question data, suitable for datasets with consistent question patterns~\cite{DBLP:conf/ijcai/WangWSDH17}.
The queries typically include both ``ASK'' and ``SELECT'' types, with ``ASK'' checking for a solution to the query pattern and ``SELECT'' returning variables from all matched solutions~\cite{DBLP:conf/ijcai/WangWSDH17}.

Term-based retrievers like TF-IDF and BM25 is another good choice, with their scoring reflecting the direct correlation between the query and fact triplets.
Luo et al.~\cite{DBLP:conf/emnlp/LuoZBB21} 
use image captions generated by a model, concatenating them with $Q$ as a query for BM25-based document retrieval.
LaKo~\cite{DBLP:conf/jist/0007HCGFP0Z22} 
presents a Stem-based BM25 approach, using word stems as the smallest semantic units to maximize knowledge extraction from limited VQA and KG resources. 
EnFoRe~\cite{DBLP:conf/emnlp/WuM22} 
utilizes entity-augmented queries to recall passages via BM25, measuring an entity's importance to the query by the relevance of these passages to the answer.

\textbf{\textit{(\rmnum{2})} Pruning.}
The pruning stage refines the coarse-grained sub-KG obtained from initial retrieval.
This involves re-ranking candidate facts and may include assigning weights to nodes based on corresponding visual object sizes~\cite{DBLP:conf/ecir/WangLT19}, ensuring each knowledge triple contains key elements from $Q$ or auto-generated captions~\cite{DBLP:conf/cvpr/SuZDCCL18,DBLP:conf/emnlp/WuM22}, or aligns with the relation type implied in $Q$~\cite{DBLP:conf/nips/NarasimhanLS18,DBLP:conf/ijcai/ZhuYWS0W20,DBLP:journals/pr/YuZWZHT20,DBLP:journals/corr/abs-2012-15484,DBLP:conf/sitis/HussainMSF22,yin2023multi}. 
Additionally, a learnable score function can assess the compatibility between a fact representation and the $Q$-$I$ representation~\cite{DBLP:conf/eccv/NarasimhanS18,DBLP:conf/sitis/HussainMSF22,DBLP:conf/wacv/RaviCSLS23}.

Furthermore, global KG-level pruning is also practiced. 
For example, KRISP~\cite{DBLP:conf/cvpr/MarinoCP0R21} gathers all symbolic entities from the VQA dataset, including questions, answers, and visual concepts recognized by visual systems, and incorporates only triples related to these concepts to the model training. 
LaKo~\cite{DBLP:conf/jist/0007HCGFP0Z22} streamlines the KG by creating a stem corpus specific to the VQA field, ensuring all KG triples contain at least one stem from this corpus.
KAT~\cite{DBLP:conf/naacl/GuiWH0BG22} extracts a subset from Wikidata~\cite{DBLP:journals/cacm/VrandecicK14} covering common real-world objects, and RR-VEL~\cite{DBLP:conf/icmcs/YouYLL23} only retains triples in the KG that include candidate answers and visually detected entities in the training set images.

\textbf{\textit{(\rmnum{3})} Dense Retrieval}~\cite{DBLP:conf/emnlp/KarpukhinOMLWEC20} methods typically retrieve the most relevant top-k facts for a given $Q$-$I$ pair.
This technique utilizes embedding similarities to match questions and visual concepts with pre-flattened concise fact sentences~\cite{DBLP:conf/nips/NarasimhanLS18,DBLP:conf/ijcai/ZhuYWS0W20,DBLP:journals/pr/YuZWZHT20,DBLP:conf/coling/ZiaeefardL20,DBLP:conf/aaai/WuLSM22,DBLP:conf/aaai/LiM22,DBLP:conf/cvpr/0013PTRWN22,DBLP:journals/corr/abs-2306-17675,DBLP:journals/kbs/LiuWHQC22,DBLP:conf/icmcs/YouYLL23,DBLP:conf/acl/SiMLJW23}, simplifying the retrieval process without complex rules. Sometimes, dense retrieval can also serve as a mechanism for KG pruning, selectively excluding information that is likely irrelevant.

Retrieval efficiency is frequently enhanced by employing open-source indexing engines like FAISS~\cite{DBLP:journals/tbd/JohnsonDJ21}, which facilitates the organization and indexing of large-scale dense embeddings. The architectures involved are generally symmetrical or siamese to support shared embedding spaces, while asymmetrical designs are adopted for Cross-Modal Retrieval scenarios (e.g., CLIP-based retrieval).
DMMGR~\cite{DBLP:conf/aaai/LiM22} ranks triplets based on the average cosine similarity between each word in the triplet and both the nouns in $Q$ and the objects detected in $I$, excluding pairs with a zero average similarity.
RR-VEL~\cite{DBLP:conf/icmcs/YouYLL23} assesses the similarity between $Q$ and key entities across various knowledge triples, using combined similarity scores to rank the candidate triples.
KAT~\cite{DBLP:conf/naacl/GuiWH0BG22} uses the CLIP model to encode patch-level image regions and knowledge entries for retrieval purposes.
MAVEx~\cite{DBLP:conf/aaai/WuLSM22} creates a concept pool for each $Q$-$A$ pair, selecting facts containing potential answers identified by other VQA models. These facts are encoded using a pre-trained BERT model for re-ranking. Its subsequent work EnFoRe~\cite{DBLP:conf/emnlp/WuM22} also prioritizes key entities in the $Q$-$I$ pair, enhancing the knowledge retrieval process by focusing on entities crucial for answering the question.
HKEML~\cite{DBLP:conf/kdd/ZhengYG021} applies 2D convolutional operations~\cite{DBLP:journals/pami/GaoCZZYT21} to align the head and relation patterns in knowledge queries with $Q$, effectively mining implicit connections within the KG pertinent to $Q$-$A$ pairs.

\textbf{\textit{(\rmnum{4})} Search Engine} is an atypical yet useful method in KG-based VQA for scenarios requiring open knowledge. 
Marino et al.~\cite{DBLP:conf/cvpr/MarinoRFM19} 
gather Wikipedia articles for each $Q$-$I$ pair and select sentences closely matching the query based on key word frequency. Their ArticleNet predicts the presence and positioning of correct answers in these articles. 
Jain et al.~\cite{DBLP:conf/sigir/JainKKJRC21} 
utilize Google’s search engine to retrieve the top-10 relevant snippets for a Machine Reading Comprehension (MRC) module based on a reformulated $Q$.
MAVEx~\cite{DBLP:conf/aaai/WuLSM22} 
enriches knowledge retrieval through Google APIs for category labels, OCR readings, and logo information,  collecting sentences from Wikipedia articles\footnote{\url{https://github.com/goldsmith/Wikipedia}} that contain candidate answers. It also uses Google Image Search with candidate $Q$-$A$ pairs to provide additional visual information. 
Luo et al.~\cite{DBLP:conf/emnlp/LuoZBB21} notice that snippet-level knowledge outperforms sentence-level, and select ten snippets for each $Q$-$A$ query.

\textbf{\textit{(\rmnum{5})} Learnable Retriever} 
refers to a trainable retrieval model that enhances the adaptability and compatibility in KG-based VQA settings~\cite{DBLP:conf/semweb/0007CGPYC21,DBLP:conf/emnlp/LuoZBB21,DBLP:conf/emnlp/LinB22,DBLP:conf/wacv/RaviCSLS23,wu2023resolving,DBLP:conf/mir/AdjaliGFGB23}.
The learnable retriever differs from dense retrieval in its adaptability to specific contexts, offering biased recall that emphasizes specific interactions between visual, textual, and knowledge elements.
This technology demands more rigorous training processes, requiring either labeled data for direct retriever training or the model's capability to autonomously generate pertinent labels.
For example, Chen et al.~\cite{DBLP:conf/semweb/0007CGPYC21} and Li et al.~\cite{DBLP:conf/mm/LiX0FZLZGW22} 
separately aligning the joint embedding of the $Q$-$I$ pair with the targets like relations in separate feature spaces.
The prediction for relation type in the sub-KG pruning operation~\cite{DBLP:conf/nips/NarasimhanLS18,DBLP:conf/ijcai/ZhuYWS0W20,DBLP:journals/pr/YuZWZHT20,DBLP:journals/corr/abs-2012-15484} mentioned before is similar. 
Furthermore,
VLC-BERT~\cite{DBLP:conf/wacv/RaviCSLS23} assigns similarity scores to inference facts for each $Q$, based on their overlap with human-annotated answers. These scores serve as weak signals for training the retriever, indicating the relevance of each fact to $Q$.
Luo et al.~\cite{DBLP:conf/emnlp/LuoZBB21} utilize DPR~\cite{DBLP:conf/emnlp/KarpukhinOMLWEC20} as a neural retriever, leveraging two BERT models for encoding the query and context. They further adapt DPR for visual domains with two variants: Image-DPR based on LXMERT~\cite{DBLP:conf/emnlp/TanB19}, and Caption-DPR, which modifies the DPR approach to suit visual content.
LaKo~\cite{DBLP:conf/jist/0007HCGFP0Z22} explores a differentiable KG retriever, leveraging cross-attention scores between the token of the prediction output and input facts for iterative reader-retriever training.
RAVQA~\cite{DBLP:conf/emnlp/LinB22} defines retrieval content as negative for a query if it doesn't aid in answer generation, using the rest as positive samples for training the retriever. 
It also utilizes the joint probability of retrieval probability and reader answer prediction for final result determination.
UnifER~\cite{DBLP:conf/mm/GuoNWLCK22} calculates a reader loss using only $Q$-$I$ inputs and compare it to the loss when incorporating retrieved knowledge, defining this difference as the \textit{loss gap}. A negative value in this gap signifies that the knowledge is counterproductive, adversely affecting question answering. 
By employing the \textit{loss gap} as a metric, the model is refined to prioritize beneficial knowledge, thereby iteratively training the retriever and reader for co-evolution.
Addressing the challenge of slow convergence and sub-optimal performance in learnable retrievers, DEDR~\cite{DBLP:conf/sigir/SalemiPZ23} employs a dual multi-modal encoder architecture with shared parameters for both $Q$-$I$ queries and knowledge content, starting from the same shared embedding space. It further explores both multi-modal and text-only retrievers, combining their results via an ensemble method. 
Training for these retrievers is based on a supervised multi-modal retrieval dataset from Qu et al.~\cite{DBLP:conf/sigir/QuZ0CL21}.
REVEAL~\cite{DBLP:conf/cvpr/HuI0WCSSRF23} integrates three data sources: WikiData KB~\cite{DBLP:journals/cacm/VrandecicK14}, Wikipedia-Image-Text (WIT)~\cite{DBLP:conf/sigir/Srinivasan0CBN21}, and the VQA2.0 dataset~\cite{DBLP:conf/iccv/AntolALMBZP15}. It utilizes a gating mechanism for optimal knowledge source selection and employs the perceiver architecture~\cite{DBLP:conf/icml/JaegleGBVZC21} to encode and compress knowledge items. This setup enables cascading multi-modal retrievers and joint reasoning, creating an end-to-end architecture.

Cold start issues can arise with asymmetric or randomly initialized retrievers, often leading to retrieving irrelevant items and providing inadequate feedback for iterative training. To address this, REVEAL~\cite{DBLP:conf/cvpr/HuI0WCSSRF23} creates an initial retrieval dataset with pseudo ground-truth knowledge, using a large-scale image-caption dataset~\cite{DBLP:conf/sigir/Srinivasan0CBN21}. 
For pre-training, REVEAL pairs passages with query images as pseudo ground-truth knowledge and, to align with VQA task formats, randomly truncates captions to predict the truncated content using the image and the remaining text.
LaKo~\cite{DBLP:conf/jist/0007HCGFP0Z22} initially employs a BM25-based retriever for knowledge retrieval in the first training phase, allowing for preliminary distillation to the differentiable retriever to mitigate the cold start problem.

\textbf{\textit{(\rmnum{6})} PLM Generation.}
Recent research has demonstrated that PLMs acquire factual knowledge, functioning as a KB when prompted appropriately ~\cite{DBLP:conf/emnlp/PetroniRRLBWM19}.
For instance,
KAT~\cite{DBLP:conf/naacl/GuiWH0BG22}  and TwO~\cite{DBLP:conf/acl/SiMLJW23} take GPT-3 to retrieve implicit textual knowledge with supporting evidence;
VLC-BERT~\cite{DBLP:conf/wacv/RaviCSLS23} uses COMET~\cite{DBLP:conf/aaai/HwangBBDSBC21}, a LM trained on commonsense KGs,  to generate contextual expansions instead of direct knowledge retrieval from KBs;
PROOFREAD~\cite{DBLP:journals/corr/abs-2308-15851} leverages ChatGPT for generating relevant $Q$ and knowledge entries for each $Q$-$I$ pair, storing them in a \textit{Demo Bank} for demonstration reuse while ensuring case diversity;
Wang et al.~\cite{DBLP:journals/corr/abs-2311-11598} utilize ChatGPT to decompose $Q$, alleviating the issue of unfocused and lacking detailed image features in image captioning;
MM-Reasoner~\cite{DBLP:conf/emnlp/KhademiYFZ23} leverages LLMs to create rationales from multi-aspect visual descriptions, integrating commonsense knowledge, external information, and supporting facts. These rationales, alongside $I$ and $Q$, are processed by a specifically fine-tuned Visual Language Model (VLM) designed to handle such enriched input.

Furthermore, several studies directly and implicitly harness the knowledge embedded in PLMs for knowledge-aware VQA reasoning, often skipping a separate knowledge retrieval step~\cite{DBLP:journals/eswa/SalaberriaALSA23,DBLP:conf/aaai/YangGW0L0W22,DBLP:journals/corr/abs-2202-04306,DBLP:conf/cvpr/Shao0W023,DBLP:conf/acl/SubramanianNKYN23}. For example, CodeVQA~\cite{DBLP:conf/acl/SubramanianNKYN23} involves a knowledge query module that leverages a PLM to answer questions based on world knowledge, highlighting the PLM's role as an integral part of the reasoning process.

\includegraphics[scale=0.022]{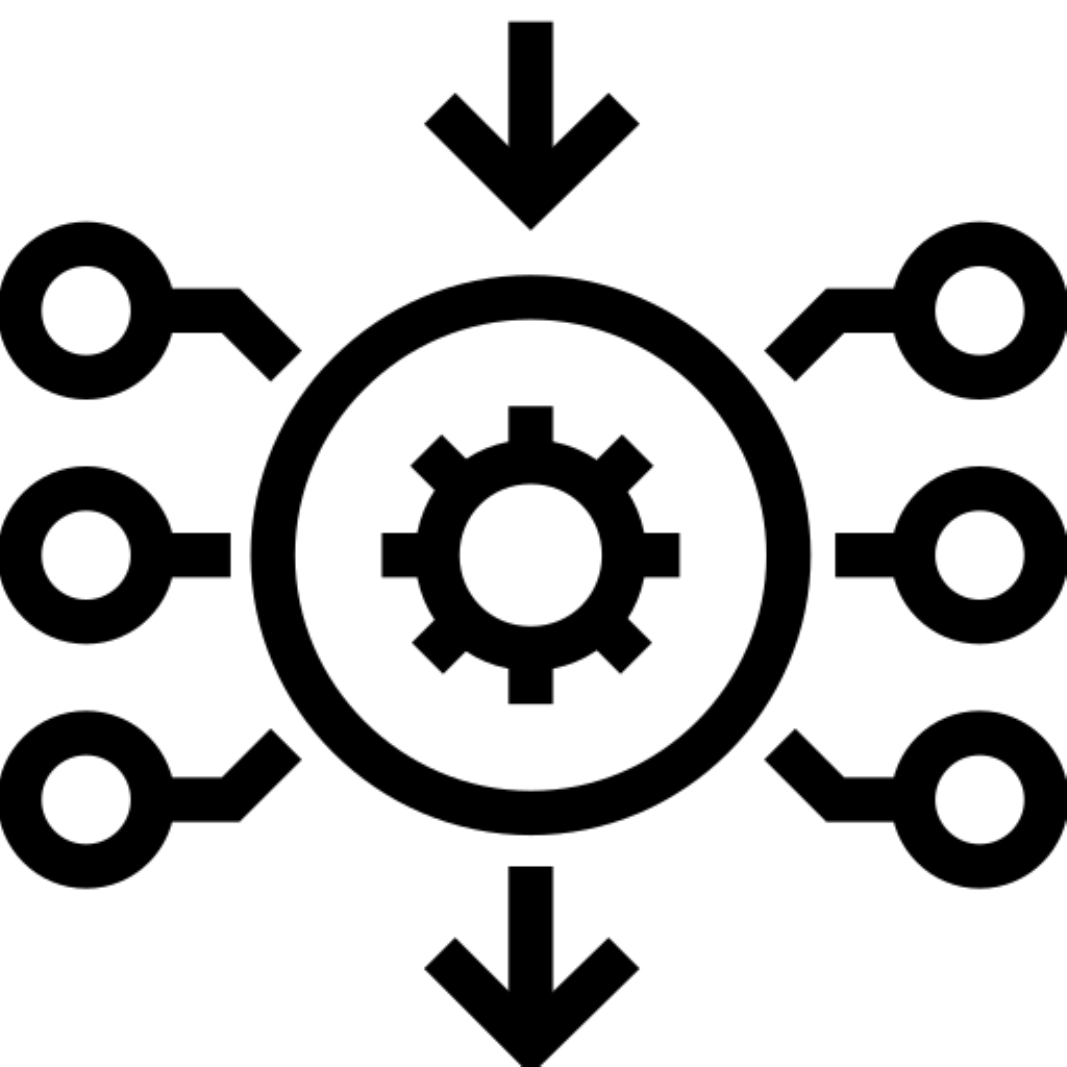}\;\textbf{\ul{Knowledge Representation}}
involves selecting the appropriate format for symbolic KGs to integrate with multi-modal models. This decision is crucial for effectively infusing knowledge into multi-modal reasoning tasks.

\textbf{\textit{(\rmnum{1})} Direct Text-to-Embedding Mapping.} 
Some research treats entities and relations in KGs as words, using embedding methods like Glove~\cite{DBLP:conf/emnlp/PenningtonSM14} to translate them into continuous vectors. This transformation enables the further compression of knowledge components (e.g., triples) into fixed-size vectors using Recurrent Neural Networks (RNNs)~\cite{DBLP:conf/ecir/WangLT19}, (V)PLMs~\cite{DBLP:conf/icmcs/YouYLL23,DBLP:conf/cvpr/HuI0WCSSRF23,DBLP:conf/emnlp/ChevalierWAC23}, or mean pooling~\cite{DBLP:conf/eccv/NarasimhanS18,DBLP:conf/nips/NarasimhanLS18,DBLP:conf/ijcai/ZhuYWS0W20,DBLP:journals/pr/YuZWZHT20,DBLP:conf/semweb/0007CGPYC21,DBLP:conf/cvpr/MarinoCP0R21,DBLP:conf/aaai/LiM22,DBLP:conf/aaai/WuLSM22,DBLP:conf/sitis/HussainMSF22}. 
Techniques such as stop-word removal in Word2Vec can further refine knowledge representation, reducing the noise from irrelevant words in mean pooling~\cite{DBLP:conf/nips/NarasimhanLS18,DBLP:journals/kbs/LiuWHQC22,DBLP:conf/jist/0007HCGFP0Z22}.
Some methods convert fact collections into natural language sentences via concatenating the relation and object entities~\cite{DBLP:conf/coling/ZiaeefardL20,DBLP:journals/tnn/ZhangLLZLSG21,DBLP:conf/icmcs/YouYLL23,DBLP:conf/cvpr/HuI0WCSSRF23}, allowing direct encoding into fixed-length vectors by PLMs. 

\textbf{\textit{(\rmnum{2})} Knowledge Graph Embedding} (KGE) 
offers a practical approach to embed facts and reveal semantic relationships among triples in an abstract space. This technology is valuable for setting up initial~\cite{DBLP:conf/cvpr/SuZDCCL18,DBLP:journals/corr/abs-2012-15484,DBLP:journals/pr/ZhengYCML021,DBLP:journals/tip/HanYWWN23} fact embeddings and gathering multi-modal knowledge~\cite{DBLP:conf/cvpr/DingYLHC022}.
During self-supervised training, signals from neighboring entities are embedded into each central entity's  unique representation. Such a process allows for the easy identification and integration of key entities into the training phase, thus efficiently simulating the retrieval of specific sub-KGs without the need for direct retrieval.
Cao et al.~\cite{DBLP:journals/tnn/CaoLLWL22} train the RotatE model on the entire KG to get entity and relation features, modifying a guided-attention block to fuse those knowledge embeddings with $I$ and $Q$ features.
Chen et al.~\cite{DBLP:conf/semweb/0007CGPYC21} evaluate various embeddings including TransE-based KG embeddings, BERT-based node representations of ConceptNet~\cite{DBLP:conf/eacl/YangDCC23,DBLP:conf/aaai/MalaviyaBBC20}, and GloVe embeddings, finding that Word2Vec representations excel at mapping answers in smaller datasets. 
RVL~\cite{DBLP:journals/corr/abs-2101-06013} utilizes the PyTorchBigGraph method~\cite{DBLP:conf/mlsys/LererWSLWBP19} for  embedding the Wikidata KG, while KVQAmeta~\cite{DBLP:conf/www/Garcia-OlanoOG22} employs Wikipedia2Vec for representing entities from Wikipedia, emphasizing KGE's versatility in representing different knowledge sources.

\textbf{\textit{(\rmnum{3})} Pure Context.} 
In many cases, KG triples are maintained in their original textual format for direct participation in multi-modal reasoning. This includes using sub-KGs for RDF query-based answer retrieval~\cite{DBLP:conf/ijcai/WangWSDH17} and serializing triples for joint reasoning with (V)PLMs~\cite{DBLP:conf/acl/VickersAMB20,DBLP:conf/jist/0007HCGFP0Z22,DBLP:conf/cvpr/0013PTRWN22,DBLP:conf/aaai/YangGW0L0W22,DBLP:conf/naacl/GuiWH0BG22,DBLP:conf/emnlp/LinB22,DBLP:conf/emnlp/WuM22,DBLP:conf/nips/LinX0X0Y22,DBLP:conf/acl/SiMLJW23,DBLP:conf/wacv/RaviCSLS23,DBLP:conf/icmcs/YouYLL23,DBLP:conf/cvpr/Shao0W023,DBLP:journals/corr/abs-2308-15851,DBLP:journals/corr/abs-2311-11598,DBLP:journals/corr/abs-2211-09699,DBLP:journals/corr/abs-2310-13570,DBLP:conf/emnlp/KhademiYFZ23,dong2024modalityaware}. 
To manage lengthy input sequences predominantly composed of facts, which might shift the model’s focus away from other crucial cues,
VLC-BERT~\cite{DBLP:conf/wacv/RaviCSLS23} 
summarizes the information from each inference sentence into a single token representation using SBERT~\cite{DBLP:conf/emnlp/ReimersG19};
Wang et al.~\cite{DBLP:journals/corr/abs-2311-11598} 
select only the contributive factual summaries (i.e., having higher contribution score than original $Q$) as caption supplements. 

\includegraphics[scale=0.45]{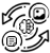}\;\textbf{\ul{Knowledge-aware Modality Interaction}} is the core of multi-modal reasoning with knowledge, mirroring human knowledge application in understanding the world.

\textbf{\textit{(\rmnum{1})} Concatenation.}
Directly merging multi-modal vectors through concatenation provides a straightforward yet effective approach for  modality fusion~\cite{DBLP:conf/eccv/NarasimhanS18}, combining different modality features into a singular representation. This unified feature is typically refined with a Multi-Layer Perceptron (MLP) to enhance modality interaction and integration. In multi-modal fusion models like MUTAN~\cite{DBLP:conf/iccv/Ben-younesCCT17}, BAN~\cite{DBLP:journals/titb/YuPFCS23}, and SAN~\cite{DBLP:conf/cvpr/YangHGDS16}, concatenation is a preliminary step before the MLP layer, crucial for sophisticated multi-modal analysis.

\textbf{\textit{(\rmnum{2})} Long Short-Term Memory} (LSTM) Network
is a foundational framework for integrating knowledge with multi-modal data. It typically employs an LSTM encoder to process semantic inputs from $I$ and $Q$, and an LSTM decoder for generating answers, initializing the hidden state with embeddings from attributes, captions, and external knowledge. $Q$ is tokenized  and fed into the system sequentially~\cite{DBLP:conf/cvpr/WuWSDH16}. Moreover, LSTMs also act as standalone encoders for textual data~\cite{DBLP:conf/eccv/NarasimhanS18,DBLP:conf/nips/NarasimhanLS18,DBLP:journals/pr/YuZWZHT20,DBLP:conf/ijcai/ZhuYWS0W20,DBLP:conf/mm/Li0020,DBLP:journals/corr/abs-2012-15484,DBLP:journals/tnn/ZhangLLZLSG21,DBLP:conf/aaai/LiM22}, employing Glove~\cite{DBLP:conf/emnlp/PenningtonSM14,DBLP:journals/tip/HanYWWN23} or PLMs~\cite{DBLP:conf/naacl/DevlinCLT19,DBLP:conf/iclr/LanCGGSS20} for token embedding initialization. The output embeddings aid in subsequent stages of modality fusion, giving LSTM a pivotal role similar to those methods in \textit{Text-to-Embedding Mapping} paradigm.

\textbf{\textit{(\rmnum{4})} Graph Neural Networks} (GNNs) 
emphasize the connection of concepts in VQA by integrating representations from $I$, $Q$, and entities into cohesive networks, where each node (entity) is represented by an embedding that is a concatenation of different modalities~\cite{DBLP:conf/nips/NarasimhanLS18}.
GNNs process entity representations over multiple iterative steps, with the final learned entity representations fed into an MLP, assigning a binary label to each entity to indicate its relevance as an answer.
Mucko~\cite{DBLP:conf/ijcai/ZhuYWS0W20} diverges from traditional modality embedding concatenation by independently processing distinct modalities' KGs. This involves isolating and separately analyzing the visual scene KG, the semantic KG from image captions, and the common sense KG, supporting precise answer determination through $Q$-guided attention and cross-KG convolution.
The method of $Q$-guided KG node weighting has seen similar implementations in other studies~\cite{DBLP:journals/pr/YuZWZHT20,DBLP:conf/mm/Li0020,DBLP:conf/nips/SaqurN20,DBLP:conf/coling/ZiaeefardL20,DBLP:conf/aaai/LiM22,DBLP:journals/kbs/LiuWHQC22,DBLP:conf/sitis/HussainMSF22,DBLP:journals/corr/abs-2205-11501}.
KG-Aug~\cite{DBLP:conf/mm/Li0020} uses GCN to generate entity representations, which are then used to embed knowledge into the features of both $Q$ and $I$.
KRISP~\cite{DBLP:conf/cvpr/MarinoCP0R21} applies a RGCN~\cite{DBLP:conf/esws/SchlichtkrullKB18}  for symbolic knowledge reasoning, enhancing each entity with four inputs: \textit{a)} A binary indicator for concept presence in $Q$; \textit{b)} Classifier probabilities for the concept's node, or zero if not detected in $I$, using various classifiers and detectors; \textit{c)} A GloVe pooling representation of the concept; \textit{d)} An implicit knowledge representation derived from a multi-modal pre-trained model~\cite{DBLP:journals/corr/abs-1908-03557}.
VQA-GNN~\cite{DBLP:journals/corr/abs-2205-11501} 
employs a multi-modal GNN with bidirectional fusion to update concept and scene graph nodes for answer prediction through inter-modal message passing.

\textbf{\textit{(\rmnum{3})} Dynamic Memory Networks} (DMNs)~\cite{DBLP:conf/icml/KumarIOIBGZPS16} utilize a attention-based mechanism for filtering critical information from localized small-scale knowledge triple embeddings, achieved by modeling interactions across multiple data channels~\cite{DBLP:conf/ecir/WangLT19,DBLP:conf/aaai/ShahMYT19,DBLP:journals/tip/HanYWWN23,yin2023multi}. 
Through \textit{triple replication}, VKMN~\cite{DBLP:conf/cvpr/SuZDCCL18} deconstructs each knowledge triple into three Key-Value pairs, for instance,  ($h$, $r$) as the key and $t$ as the value, reducing interference caused by using only head and tail entities as keys for retrieval thereby improving reasoning accuracy.
DMMGR~\cite{DBLP:conf/aaai/LiM22} follows this setting and further 
refines knowledge triple composition by using the average embedding of a triple as a key and its individual elements as values for enhanced relevance assessment. 
These networks apply a multi-scale attention mechanism that initially evaluates the overall relevance of a triplet's embedding, then assess the importance of each element, leading to more accurately recalled dynamic memories.
GRUC~\cite{DBLP:journals/pr/YuZWZHT20} uses visual and semantic scene graphs as knowledge sources for external memory, iteratively updating multi-modal memories and employing a GRU module to refresh factual entity representations, incorporating inputs from previous entities and memory from the last time step.
SUPER~\cite{DBLP:journals/tip/HanYWWN23} further advances this field by integrating a memory augmented component to retain and adjust key clues for answering questions, a method named \textit{memory reactivation}.
REVEAL~\cite{DBLP:conf/cvpr/HuI0WCSSRF23} unifies multi-modal data by compressing each entry into a set number of value embeddings and a single key embedding for memory storage, achieving synchronous and stable updates between the memory encoder and main framework by re-encoding a portion (10\%) of the retrieved knowledge items in each training iteration.

\textbf{\textit{(\rmnum{5})} Guided-Attention \& Transformer.}
The Transformer architecture, with its multi-head attention, layered stacking, and residual connections,  is widely used in multi-modal fusion~\cite{DBLP:conf/nips/VaswaniSPUJGKP17}. It allows for ample interaction between multi-modal information, where knowledge embeddings interact on par with other modalities.
Many studies~\cite{DBLP:journals/corr/abs-2012-15484,DBLP:conf/emnlp/GarderesZAL20,DBLP:journals/tnn/ZhangLLZLSG21,DBLP:journals/tnn/CaoLLWL22,DBLP:conf/aaai/WuLSM22,DBLP:conf/acl/HeoKCZ22}
have adopted a guided-attention mechanism to merge knowledge embeddings with visual and textual features. 
Unlike self-attention, guided-attention uses a separate set of features to steer the attention learning process, allowing for diverse integrations such as knowledge-guided visual/textual embeddings or $Q$-guided visual/knowledge embeddings.

\textbf{\textit{(\rmnum{6})} PLM \& VLM Reasoning.}
Integrating PLMs and VLMs in multi-modal knowledge fusion is an emerging trend that emphasizes efficient knowledge-aware modality interaction and answer reasoning.  This allows researchers to concentrate on the organization of input data and the design of training objectives without extensively modifying the core model structure. By doing so, it effectively utilizes the inherent knowledge in pre-trained models, simplifying the  development of processing pipelines.
LM-based reasoning primarily falls into two categories: 

\textit{a)} \textbf{Embedding-Based Visual Information Integration}:
This category includes methods that convert visual data into embeddings compatible with the input specifications of (V)PLMs~\cite{DBLP:conf/cvpr/DouXGWWWZZYP0022}.
It involves techniques that restructure visual inputs into embeddings which seamlessly integrate with the model's existing architecture, such as  compressing patch or local object features into fixed-length embedding sets~\cite{DBLP:conf/icml/JaegleGBVZC21,DBLP:conf/cvpr/HuI0WCSSRF23} or applying adapters or projection heads for cross-modal feature space alignment~\cite{DBLP:conf/nips/LinX0X0Y22,DBLP:journals/corr/abs-2306-06687}. These visual embeddings, combined with textual inputs, are processed in the embedding layers of (V)PLMs~\cite{DBLP:conf/icml/JaegleGBVZC21,DBLP:journals/corr/abs-2306-17675}. 
Some studies~\cite{DBLP:conf/acl/VickersAMB20,DBLP:conf/emnlp/LuoZBB21,DBLP:conf/mm/GuoNWLCK22,DBLP:conf/wacv/RaviCSLS23,DBLP:conf/sigir/SalemiPZ23} integrate retrieved knowledge content and questions with image regions of interest, subsequently fine-tuning VLMs end-to-end on the VQA dataset using ground truth answers for optimization.
Typically, VLMs such as UNITER~\cite{DBLP:conf/eccv/ChenLYK0G0020}, ViLT~\cite{DBLP:conf/icml/KimSK21}, VL-BERT~\cite{DBLP:conf/iclr/SuZCLLWD20}, LXMERT~\cite{DBLP:conf/emnlp/TanB19} and VL-T5~\cite{DBLP:conf/icml/ChoLTB21} can be categorized into two paradigms: two-stream and single-stream, whose main distinction lie on the moment selection for performing intra-modal fusion.
RVL~\cite{DBLP:journals/corr/abs-2101-06013} and KVQAmeta~\cite{DBLP:conf/www/Garcia-OlanoOG22} inject the knowledge into the VLMs via aligning the KG embedding with the corresponding textual phrase representations derived from the output summations of PLM's embedding layers.
MuKEA~\cite{DBLP:conf/cvpr/DingYLHC022} uses the visual and language output sides of the LXMERT as the head and relation of a triple, respectively, pairing these with the ground truth answer as the tail entity. 
This association, aroused through the KGE method (e.g., TransE), leverages implicit knowledge within VLMs for reasoning.
VLC-BERT~\cite{DBLP:conf/wacv/RaviCSLS23} uses a $Q$-guided multi-head attention block to fuse multiple knowledge representation vectors before feeding them into the VLM.
He et al.~\cite{DBLP:conf/eacl/HeW23} propose a graph-involved $Q$-attention mechanism, where $V$-$Q$ guided graphs are built to direct VLM training by integrating a graph-aware mask matrix into the Transformers' attention matrix.
Pang et al.~\cite{DBLP:journals/corr/abs-2310-12973}  enhance a VLM's ability for parametric knowledge injection by integrating the frozen Transformer layer of the LLM (LLaMA ~\cite{DBLP:journals/corr/abs-2302-13971}) between its cross-modal fusion and decoder modules.

\textit{b)} \textbf{Textual Conversion of Visual Data:}
This category involves converting all visual information into a textual format, like captions, enabling the application of PLM reasoning to a uniform textual dataset that includes background knowledge, questions, and images~\cite{DBLP:journals/eswa/SalaberriaALSA23,DBLP:conf/jist/0007HCGFP0Z22,DBLP:conf/emnlp/LuoZBB21,DBLP:journals/corr/abs-2202-04306,DBLP:conf/aaai/YangGW0L0W22,DBLP:conf/cvpr/0013PTRWN22,DBLP:conf/acl/SiMLJW23,DBLP:conf/icmcs/YouYLL23,DBLP:journals/corr/abs-2308-15851,DBLP:journals/corr/abs-2211-09699,DBLP:journals/corr/abs-2310-13570}. 
These works usually hold that 
text-only PLMs can effectively infer answers, even compensating for the loss of fine-grained visual features in image captions.
Chen et al.~\cite{DBLP:conf/jist/0007HCGFP0Z22} 
illustrate how encoder-decoder PLMs address the long-tail problem in answers and discrepancies between training and testing sets, while avoiding span prediction and directly generating free-form answers.
Jain et al.~\cite{DBLP:conf/sigir/JainKKJRC21} 
reframe VQA as a MRC task, integrating search engines for additional context.
TRiG~\cite{DBLP:conf/cvpr/0013PTRWN22} and TwO~\cite{DBLP:conf/acl/SiMLJW23} 
expand this approach to include object-level (e.g., object, attribute, and OCR labels) information alongside captions.
Utilizing LLMs like GPT-3 with image captions,
PICa~\cite{DBLP:conf/aaai/YangGW0L0W22} 
reveals that pure PLMs can achieve impressive performance in zero-shot and few-shot learning scenarios.
KAT~\cite{DBLP:conf/naacl/GuiWH0BG22} further queries GPT-3 for providing reasoning evidence,
aiming to extract deeper insights and implicit knowledge from GPT-3's outputs to bolster the reasoning process.
REVIVE~\cite{DBLP:conf/nips/LinX0X0Y22} 
employs a Transformer encoder as an adapter to utilize fine-grained regional visual information.
PROOFREAD~\cite{DBLP:journals/corr/abs-2308-15851} utilizes XGBoost~\cite{DBLP:conf/kdd/ChenG16}, a gradient-boosted decision tree model, as a knowledge perceiver to classify knowledge entries based on their contribution scores across various dimensions.
The Fusion-in-Decoder (FiD) approach~\cite{DBLP:conf/eacl/IzacardG21}, where knowledge is individually compressed in the encoder and then jointly utilized in the decoder for reasoning, is adopted by various studies~\cite{DBLP:conf/naacl/GuiWH0BG22,DBLP:conf/cvpr/0013PTRWN22,DBLP:conf/jist/0007HCGFP0Z22,DBLP:conf/emnlp/WuM22,DBLP:conf/nips/LinX0X0Y22,DBLP:conf/sigir/SalemiPZ23,DBLP:conf/acl/SiMLJW23}. This allows for the simultaneous input of a large corpus of uni-modal or multi-modal background knowledge into the (V)PLMs.

To mitigate the loss of fine-grained visual details in caption-based conversion,
Wang et al.~\cite{DBLP:journals/corr/abs-2311-11598} leverage the LLM's reasoning capabilities to spotlight critical image details that that might be overlooked in captions.
By decomposing the main $Q$ into sub-questions and obtaining answers via a pre-trained VQA model, they identify and select those factual summaries with higher contribution scores than the original $Q$, supplementing the initial captions with these key details.
This is similar to KAT~\cite{DBLP:conf/naacl/GuiWH0BG22} and TwO~\cite{DBLP:conf/acl/SiMLJW23}, 
which apply In-Context Learning (ICL) in GPT-3 which employs a combination of $Q$, caption, and object labels as the prompt to generate implicit textual knowledge;
PromptCap~\cite{DBLP:journals/corr/abs-2211-09699} introduces $Q$-guided caption generation to cover the visual details required by $Q$;
ASB~\cite{DBLP:journals/corr/abs-2310-13570}
identifies the image patches most relevant to $Q$ and generates informative captions from these patches only;
Cola-FT~\cite{DBLP:journals/corr/abs-2310-15166} prompts VLMs to generate captions and plausible answers separately, which are then concatenated with the instruction prompt, $Q$, and choices, forming a holistic context for LLMs to logically deduce the answer.

\includegraphics[scale=0.45]{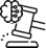}\;\textbf{\ul{Knowledge-aware Answer Determination}} 
plays a crucial role in generating and predicting answers, often overlapping with \textit{Knowledge-aware Modality Interaction}. Certain methods uniquely address both these aspects simultaneously, highlighting their intertwined nature.

\textbf{\textit{(\rmnum{1})} Information Extraction.} 
This category includes methods that retrieve or extract specific entities from KGs or documents to serve as answers.
Many query-based methods~\cite{DBLP:conf/ijcai/WangWSDH17} obtain the final answer  through inference on sub-KGs, offering benefits such as higher matching accuracy, relevance, and interpretability of answers, which are not limited by the training set's scope.
Their effectiveness, however, depends on the model’s capability to parse queries and the completeness of the KG. 
Challenges arise with non-unique or difficult-to-find answers.
To further rank the potential answers, some approaches implement heuristic rules, like matching score calculation~\cite{DBLP:journals/pami/WangWSDH18,DBLP:conf/eccv/NarasimhanS18} and answer frequency assessment~\cite{DBLP:journals/pami/WangWSDH18}.
Another technique, inspired by the MRC models, involves extracting a specific text span from KBs to answer queries.
For example, Luo et al.~\cite{DBLP:conf/emnlp/LuoZBB21} 
employ a RoBERTa-based text encoder~\cite{DBLP:journals/corr/abs-1907-11692} with a unique input structure, which uses two linear layers to determine the start and end of the answer span. They also implement a strategy where the model predicts an ``{\tt unanswerable}'' tag if the knowledge is insufficient, addressing the potential noise in retrieved knowledge.

\textbf{\textit{(\rmnum{2})} Discrimination.} 
Particularly suited in multi-choice VQA tasks, these methods integrate candidate answers with knowledge, questions, and images, feeding them through a discriminator for final selection. Such methods  are effective when narrowing down potential answers within a certain range, often using GNN-alike models~\cite{DBLP:conf/nips/NarasimhanLS18,DBLP:journals/kbs/LiuWHQC22,DBLP:conf/sitis/HussainMSF22} as the backbones. 
Furthermore,
discriminators can be either MLP-based~\cite{DBLP:conf/ecir/WangLT19,DBLP:conf/nips/NarasimhanLS18,DBLP:conf/ijcai/ZhuYWS0W20,DBLP:journals/pr/YuZWZHT20,DBLP:journals/kbs/LiuWHQC22} or rule-based~\cite{DBLP:conf/eccv/NarasimhanS18}. A notable limitation of this approach is time consumption, especially when dealing with extensive answer vocabularies.

\textbf{\textit{(\rmnum{3})} Classification.} 
In many VQA tasks, the range of possible answers is pre-determined, typically constrained either by their frequency range or a minimum occurrence threshold set during training. Consequently, many studies reformulate the question-answering process as a classification problem, often employing a fully connected (FC) or MLP layer for answer prediction~\cite{DBLP:conf/cvpr/SuZDCCL18,DBLP:conf/cvpr/MarinoRFM19,DBLP:conf/aaai/ShahMYT19,DBLP:conf/mm/Li0020,DBLP:conf/coling/ZiaeefardL20,DBLP:conf/nips/SaqurN20,DBLP:journals/tnn/ZhangLLZLSG21,DBLP:conf/emnlp/GarderesZAL20,DBLP:journals/pr/ZhengYCML021,DBLP:conf/aaai/LiM22,DBLP:conf/mm/GuoNWLCK22,DBLP:journals/tip/HanYWWN23,DBLP:conf/acl/HeoKCZ22,DBLP:conf/mm/LiX0FZLZGW22,DBLP:journals/corr/abs-2205-11501,DBLP:journals/pr/SongLLYSS23}, where the output dimension corresponds to the pre-defined number of answer candidates.
Chen et al.~\cite{DBLP:conf/semweb/0007CGPYC21} introduce an answer masking strategy that imposes direct knowledge-based constraints on the classifier's predicted answer probabilities, thereby limiting the range of potential answers. This method parallels KRISP~\cite{DBLP:conf/cvpr/MarinoCP0R21}, which employs late fusion to integrate the implicit and symbolic components of the model, selecting the highest-scoring answer from the combined answer vectors.
MAVEx~\cite{DBLP:conf/aaai/WuLSM22} introduces an answer validation module that leverages knowledge features from retrieved $I$, ConceptNet, and Wikipedia for answer candidate validation.
For (V)PLM-based methods, a classification head is typically appended to the output {\tt [CLS]} embedding~\cite{DBLP:journals/corr/abs-2101-06013,DBLP:conf/emnlp/LuoZBB21,DBLP:journals/eswa/SalaberriaALSA23,DBLP:conf/www/Garcia-OlanoOG22,DBLP:conf/mm/GuoNWLCK22,DBLP:conf/cvpr/DingYLHC022,DBLP:conf/eacl/HeW23,DBLP:conf/wacv/RaviCSLS23}, often utilizing encoder-based backbones like LXMERT~\cite{DBLP:conf/emnlp/TanB19} and BERT~\cite{DBLP:conf/naacl/DevlinCLT19}.

However, a significant trade-off common to classification-based approaches still exists, as noted by Chen et al.~\cite{DBLP:conf/jist/0007HCGFP0Z22}: the necessity to balance answer coverage and error rate, which hinges on pre-defining the answer candidate set according to its occurrence frequency.

\textbf{\textit{(\rmnum{4})} Generation.} 
Textual Generative Models have become increasingly important in VQA tasks, particularly for addressing questions with answers outside pre-defined vocabularies. Traditional LSTM-based methods generate answers sequentially, often  encoding each word as a one-hot vector from a word dictionary~\cite{DBLP:conf/cvpr/WuWSDH16}.
Considering the emergent capabilities of LMs that grow with the expansion of parameters and the scale of pre-training data~\cite{DBLP:journals/corr/abs-2304-15004}, the accuracy of generated answers by these models now compensates for the loss imposed by exact match criteria. As a result, generative (V)PLM-based methods are now increasingly supplanting traditional classification-based approaches~\cite{DBLP:conf/jist/0007HCGFP0Z22,DBLP:conf/aaai/YangGW0L0W22,DBLP:conf/cvpr/0013PTRWN22,DBLP:conf/naacl/GuiWH0BG22,DBLP:conf/emnlp/LinB22,DBLP:conf/emnlp/WuM22,DBLP:journals/corr/abs-2202-04306,DBLP:conf/nips/LinX0X0Y22,DBLP:conf/sigir/SalemiPZ23,DBLP:conf/icmcs/YouYLL23,DBLP:conf/acl/SiMLJW23,DBLP:conf/cvpr/HuI0WCSSRF23,DBLP:conf/cvpr/Shao0W023,DBLP:journals/corr/abs-2308-15851,DBLP:journals/corr/abs-2306-17675,DBLP:journals/corr/abs-2311-11598,DBLP:journals/corr/abs-2211-09699,DBLP:journals/corr/abs-2310-13570,DBLP:journals/corr/abs-2310-20159,DBLP:conf/emnlp/KhademiYFZ23}.
As illustrated in Table \ref{tab:kvqa-bm}, a noticeable increase in text-generation-based VQA methods is observed in the last two years. This trend can also be attributed to the limitations of Exact Match answer evaluation manner in VQA benchmarks, which historically do not provide an advantage in evaluating the performance of open-ended generative models~\cite{DBLP:journals/corr/abs-2304-15004}.  

These methods, using decoder-based or encoder-decoder models like GPT-3~\cite{DBLP:conf/nips/BrownMRSKDNSSAA20}, T5~\cite{DBLP:journals/jmlr/RaffelSRLNMZLL20}, VL-T5~\cite{DBLP:conf/icml/ChoLTB21}, and BLIP-2~\cite{DBLP:conf/icml/0008LSH23}, input constructed prompts to implicitly retrieve knowledge and perform analytical reasoning. Answer generation often relies on greedy decoding or beam search strategies~\cite{DBLP:conf/cvpr/0013PTRWN22,DBLP:conf/wacv/RaviCSLS23,DBLP:journals/corr/abs-2311-06411}, with the former selecting the most probable token at each step and the latter maintaining a fixed-size beam to  produce a list of ranked answer candidates.
To improve few-shot learning performance in models with large parameters, such as GPT-3, strategies like incorporating high-quality ICL examples~\cite{DBLP:conf/cvpr/Shao0W023,DBLP:journals/corr/abs-2311-11598,DBLP:journals/corr/abs-2211-09699,DBLP:journals/corr/abs-2310-13570} and employing multi-query ensembles~\cite{DBLP:conf/aaai/YangGW0L0W22,DBLP:journals/corr/abs-2310-13570} are effective. Prophet~\cite{DBLP:conf/cvpr/Shao0W023} enhances this process by first generating candidate answers using a standard VQA model, subsequently refined through GPT-3. Meanwhile, Cola-FT~\cite{DBLP:journals/corr/abs-2310-15166} prompts VLMs to generate captions and plausible answers separately, then integrating them with the instructional prompt, question, and candidate options for LLM-based reasoning.

Recent advancements include CodeVQA~\cite{DBLP:conf/acl/SubramanianNKYN23}, a training-free method prompting Codex~\cite{DBLP:journals/corr/abs-2107-03374} with in-context examples to break down $Q$ into Python code. This method leverages pre-defined visual modules in pre-trained VLMs, utilizing conditional logic and arithmetic. In line with NLP findings that LLMs improve performance at reasoning tasks when solving problems step-by-step~\cite{DBLP:conf/nips/Wei0SBIXCLZ22,DBLP:journals/corr/abs-2306-06687}, VQA performance improves by decomposing $Q$ and answering sub-questions sequentially. Khandelwal et al.~\cite{DBLP:journals/corr/abs-2311-06411} propose \textit{Successive Prompting}, where a LLM generates and resolves follow-up questions one at a time using a VLM, culminating in an answer to the original $Q$.

\textbf{METRICS:}
In VQA performance evaluation, \textit{Accuracy} (Acc), defined as the proportion of correctly answered test questions, is a predominant metric. 
A standard technique for computing Acc is recommended in the VQA challenge~\cite{DBLP:conf/iccv/AntolALMBZP15}:
\begin{equation}
	\operatorname{Acc}(ans)=\min (1, \frac{\#\{{ human~that~said~that~ans }\}}{3})\,. 
\end{equation}
This metric assigns a soft score (ranging from 0 to 1) to each answer, based on a voting mechanism among multiple annotators. In contrast, the Exact Match (EM) metric treats all annotated answers as ground truth (GT), offering a less stringent evaluation criterion~\cite{DBLP:conf/cvpr/0013PTRWN22}.
Additionally, the WuPalmer similarity (WUPS)~\cite{wu1994verb} calculates the similarity between words based on their common sub-sequences in a taxonomy tree. A candidate answer is considered correct if its similarity to a reference word exceeds a specified threshold. Chen et al.~\cite{DBLP:conf/jist/0007HCGFP0Z22} introduce Inclusion-based and Stem-based Acc metrics. The former considers an answer $A$ correct if it includes or is included by a GT answer after normalization. The latter assesses correctness based on the intersection of stems between $A$ and the GT $A$ (e.g., the stem of ``\textit{happy}'' and ``\textit{happiness}'' is ``\textit{happi}'').
Not that other NLP automatic evaluation metrics, beyond assessing answer correctness, can also evaluate the model’s explanation quality. For example, generative metrics such as BLEU~\cite{DBLP:conf/acl/PapineniRWZ02}, CIDEr~\cite{DBLP:conf/cvpr/VedantamZP15}, and METEOR measure the linguistic quality and relevance of rationale statements against a reference set. Originally developed for machine translation, these metrics provide insights into the generated explanations' coherence and fluency, complementing the evaluation of answer correctness.

Given the limitations of lexical matching metrics in evaluating open-domain VQA predictions from generative models, where entirely different words may convey the same meaning, ~\cite{DBLP:journals/corr/abs-2311-06411}, Kamalloo et al.~\cite{DBLP:conf/acl/KamallooDCR23} further propose an evaluation metric that leverages InstructGPT~\cite{DBLP:conf/nips/Ouyang0JAWMZASR22}, prompting it with $Q$ and candidate answers to  determine correctness:
\lstset{%
  basicstyle=\ttfamily\small,
  columns=fullflexible,
  frame=single,
  breaklines=true,
  keepspaces=true,
  xleftmargin=.05\textwidth, 
  xrightmargin=.05\textwidth
}
\begin{lstlisting}
Question: What is he doing?
Answer: horseback riding
Candidate: riding a horse
Is the candidate correct? [yes/no]
\end{lstlisting}

\textbf{KNOWLEDGE BASE:}\label{sec:kbtype}
The background KBs for knowledge-aware multi-modal reasoning frequently involves multiple KGs, each bringing its unique and complementary insights to the reasoning process.
\textbf{Trivia} knowledge, such as DBpedia, provides facts about famous people, places, and events. \textbf{Commonsense} knowledge, represented by sources like ConceptNet, offers insights into basic concepts like the composition of houses or parts of a wheel. \textbf{Scientific} knowledge, found in databases like hasPart KB, details classifications and properties, such as the genus of dogs or types of nutrients. Lastly, \textbf{situational} knowledge from resources like Visual Genome offers contextual data, e.g., typical locations of cars or common contents found in bowls.

\textbf{\textit{(\rmnum{1})} ConceptNet}~\cite{speer2017conceptnet}  encapsulates human commonsense knowledge, containing various relations including \textit{usedFor}, \textit{createdBy}, and \textit{isA}, primarily generated from the Open Mind Common Sense (OMCS) project;
\textbf{\textit{(\rmnum{2})} DBpedia}~\cite{DBLP:conf/semweb/AuerBKLCI07}, 
constructed from Wikipedia, spans multiple fields relevant to daily life. In this KG, concepts are connected through categories and super-categories in accordance with the SKOS\footnote{\url{http://www.w3.org/2004/02/skos/}} Vocabulary;
\textbf{\textit{(\rmnum{3})} WebChild}~\cite{DBLP:conf/wsdm/TandonMSW14} connects nouns with adjectives through fine-grained relations, such as \textit{hasShape}, \textit{faster}, \textit{bigger}. This information is automatically extracted from the Web;
\textbf{\textit{(\rmnum{4})} Wikidata}~\cite{DBLP:journals/cacm/VrandecicK14} 
offers extensive factual knowledge, including a broad range of topics about the world;
\textbf{\textit{(\rmnum{5})} hasPart KB}~\cite{DBLP:journals/corr/abs-2006-07510} 
documents relationships between objects, both common and scientific, such as \textit{(Dog, hasPart, Whiskers)} and \textit{(Molecules, hasPart, Atoms)};
\textbf{\textit{(\rmnum{6})} Visual Genome (VG)}~\cite{DBLP:journals/ijcv/KrishnaZGJHKCKL17} 
gathers scene graphs from real-life situations, focusing on spatial relationships, e.g., \textit{(Boat, isOn, Water)}, and common affordances, e.g., \textit{(Person, sitsOn, Couch)};
\textbf{\textit{(\rmnum{7})} ATOMIC}~\cite{DBLP:conf/aaai/HwangBBDSBC21} consists of over 1M knowledge triplets covering a range of topics, including physical-entity relations, event-centered relations, and social interactions.
\textbf{\textit{(\rmnum{8})} CSKG}~\cite{DBLP:conf/esws/IlievskiSZ21} is a large consolidated source that integrates commonsense knowledge from seven diverse and disjoint sources, including ConceptNet, Wikidata, ATOMIC, VG, Wordnet~\cite{miller1995wordnet}, Roget~\cite{kipfer1992roget} and FrameNet~\cite{DBLP:conf/acl/BakerFL98}.

\textbf{BENCHMARKS:}
We select FVQA~\cite{DBLP:journals/pami/WangWSDH18} and OKVQA~\cite{DBLP:conf/cvpr/MarinoRFM19} as our primary datasets due to their critical contributions to advancing knowledge-aware VQA and their significant impact on the development of subsequent datasets. Table~\ref{tab:kvqa-bm} presents a chronological analysis of relevant methods, detailing their performance, model paradigms, and design principles.
To further aid in understanding the field's development, we also include an analysis of knowledge-based VQA methods that do not rely on KGs, indicated by the {\small \faToggleOff} icon.

\input{tab/kgvqa-bm}

\textbf{RESOURCES:}
In analyzing the evolution of KG-aware VQA datasets, we categorize the developments into three main groups: FVQA-type, OKVQA-type, and others.

\textbf{\textit{(\rmnum{1})} FVQA~\cite{DBLP:journals/pami/WangWSDH18}:} The \textbf{KB-VQA} dataset~\cite{DBLP:conf/ijcai/WangWSDH17} first evaluates VQA algorithms' ability to leverage external knowledge for answering complex image-based questions. 
It consists of multiple $Q$-$A$ pairs per image, crafted by five questioners using predefined templates. These pairs aim to probe knowledge levels that surpass mere visual observation by leveraging DBpedia as the knowledge source.
Expanding on KB-VQA, FVQA~\cite{DBLP:journals/pami/WangWSDH18} includes more questions, images and integrates additional KGs such as ConceptNet and Webchild.  Notably, FVQA is the first VQA dataset to provide supporting facts for each question (i.e., external knowledge facts, rather than visual relation facts in R-VQA~\cite{DBLP:conf/kdd/LuJZDZW18}), paving the way for developing more knowledgeable VQA systems.
\textbf{Variants:} 
\textbf{ZS-F-VQA}~\cite{DBLP:conf/semweb/0007CGPYC21} targets Zero-shot VQA, designed to prevent overlap between training and testing answers, paying attention on answer bias and Out Of Vocabulary (OOV) issues.
\textbf{KRVQA}~\cite{DBLP:journals/tnn/CaoLLWL22} imposes constraints to promote image context engagement over mere knowledge fact memorization;
\textbf{FVQA 2.0}~\cite{DBLP:conf/eacl/LinWB23} increases dataset size and introduces adversarial question variants to balance the original dataset’s answer distribution;

\textbf{\textit{(\rmnum{2})} OKVQA~\cite{DBLP:conf/cvpr/MarinoRFM19}: }
Different from FVQA, 
OKVQA dataset focuses on open-world VQA, involving questions that implicitly require external knowledge without specifying a direct KB link or providing explicit KG triplets. Its broad knowledge scope makes it a benchmark alongside the VQA2.0 dataset~\cite{DBLP:conf/iccv/AntolALMBZP15}.
\textbf{Variants: } 
\textbf{OKVQA$_{S3}$} and \textbf{S3VQA}~\cite{DBLP:conf/sigir/JainKKJRC21} 
enhance the original OK-VQA by incorporating questions that require object detection within images, with subsequent substitution of the detected object in the query and employing web searches to find answers; 
\textbf{A-OKVQA}~\cite{DBLP:conf/eccv/SchwenkKCMM22} introduces a greater diversity of world knowledge and more reasoning steps to extend OK-VQA,  further providing rationales for each question to aid in training explainable VQA models.  
\textbf{OKVQA2.0}~\cite{DBLP:conf/icassp/ReichmanSRZCSTGSCPJH23} refines OK-VQA with corrections and attaching Wikipedia sources to $Q$-$I$ pairs;
\textbf{ConceptVQA}~\cite{DBLP:journals/corr/abs-2310-08148} enriches OK-VQA with entity-level annotations aligned with ConceptNet entities and presents a unique challenge by ensuring its testing split features non-overlapping answers with the training set, similar to ZS-F-VQA~\cite{DBLP:conf/semweb/0007CGPYC21}.

\textbf{\textit{(\rmnum{3}) Others: }}
Li et al.~\cite{DBLP:journals/corr/abs-1712-00733} develop \textbf{Visual7W+KB} from the Visual7W test split images~\cite{DBLP:conf/cvpr/ZhuGBF16}, automating question creation using predefined templates and ConceptNet~\cite{speer2017conceptnet} for guidance;
\textbf{KVQA}~\cite{DBLP:conf/aaai/ShahMYT19} incorporates world knowledge about named entities like \textit{Barack Obama} and the \textit{White House} from Wikidata~\cite{DBLP:journals/cacm/VrandecicK14}, also employing face identification technology in image analysis;
\textbf{ViQuAE}~\cite{DBLP:conf/sigir/LernerFGBB0L22} 
extends KVQA’s scope to include a broader range of entity types beyond just persons;
\textbf{VCR}~\cite{DBLP:conf/cvpr/ZellersBFC19} 
targets understanding human intentions in movie scenes with questions such as ``\textit{why is {\tt [PERSON]} doing this?}'';
\textbf{AI-VQA}~\cite{DBLP:conf/mm/LiX0FZLZGW22} 
utilizes Visual Genome scene graphs and ATOMIC KG~\cite{DBLP:conf/aaai/HwangBBDSBC21} event knowledge, enriched by including volunteer-annotated QA pairs and  detailed scene/object descriptions;
\textbf{DANCE}~\cite{DBLP:conf/cvpr/YeX0XY0023}
re-formats knowledge triples as natural language riddles paired with images, aiming to infuse visual language models with commonsense knowledge;
Gao et al.~\cite{gao2023lora} introduce \textbf{LoRA}, a dataset focusing on formal and complex description logic reasoning in VQA. Centered around a KB related to food and kitchen scenarios, LoRA aims to enhance the logical reasoning capabilities of VQA models, which are not adequately assessed by existing VQA datasets;
\textbf{ScienceQA}~\cite{DBLP:conf/nips/LuMX0CZTCK22}, sourced from elementary and high school science curricula, includes $21,208$ items along with lectures and explanations. It challenges models to generate coherent explanations across a wide range of subjects, setting it apart from OKVQA.  Despite not incorporating a KG in its design, ScienceQA is pivotal for advancing knowledge-intensive multi-modal models, which marks a significant step in the evolution of future KG-aware VQA methods.

In addition, KG-aware VQA can also extend to various scenarios beyond traditional settings.
For example,
\textbf{KnowIT} VQA~\cite{DBLP:conf/aaai/GarciaOCN20} 
contains video clips from ``\textit{The Big Bang Theory}'' with associated knowledge-based QA pairs, annotated by those dedicated fans well-versed in the show’s content.
\textbf{K-EQA}~\cite{DBLP:journals/pami/TanGGLS23} employs a KB and 3D scene graphs, enabling an AI agent to navigate environments and answer environment-aware natural language queries.

\begin{discussion}
VQA datasets vary in their answer formats, ranging from multiple-choice, where models select from provided options, to open-ended formats that test a model's understanding, reasoning, and independent answer generation or retrieval capabilities.
Beyond answer formats, an important consideration in these datasets is the use of a Ground Truth (GT) set of facts for answering questions. Datasets like those in the FVQA series come with their own GT facts, while those in the OKVQA series do not. These facts should ideally be employed not for training purposes (such as pre-training a relation classifier) but for assessing the model's proficiency in KG fact retrieval. Besides, selecting appropriate knowledge sources and methods for knowledge filtering is also crucial for model performance.

Furthermore, as indicated in Table~\ref{tab:kvqa-bm}, the comparison of VQA works can be influenced due to varying background KG sources and backbone models. Ensuring consistency in these aspects is essential for fair comparative analysis. 
It's important for researchers to distinguish whether improvements are due to the quality of the KB, the method of KG integration, or the backbone's inherent capabilities.
These distinctions, often overlooked, are crucial to understanding genuine progress in the field.
Relying solely on sophisticated visual, language, or multi-modal backbones to claim SOTA results, without addressing the uniformity of model parameters and ensuring fair comparisons, may compromise the credibility of the findings.
Given VQA's practical applications, additional factors such as time, space complexity, real-time consumption, and GPU requirements are also significant for a comprehensive evaluation of these models.
\end{discussion}

\subsubsection{Visual Question Generation}\label{sec:kgvqg}
VQG~\cite{DBLP:journals/tcsv/XieFCHL22,DBLP:conf/mm/ChenGX0023,DBLP:conf/ictir/SalemiRZ23}  leverages visual cues to generate questions, diverging from traditional VQA by prioritizing question creation. This process is crucial in educational applications, such as engaging children with questions about images to support learning.
Early VQG models~\cite{DBLP:conf/acl/MostafazadehMDM16} utilize RNNs to generate questions based solely on images, leading to questions that often lack specific focus. 
In the KG-aware VQG domain, volunteers create the K-VQG dataset~\cite{DBLP:conf/wacv/UeharaH23} by integrating external knowledge from resources like ConceptNet and Atomic~\cite{DBLP:conf/aaai/HwangBBDSBC21} with image content, using partially masked commonsense triplets to enrich questions with knowledge.
Xie et al.~\cite{DBLP:journals/tcsv/XieFCHL22} develop a pipeline comprising a visual concept feature extractor, knowledge representation extractor, target object extractor, and a decoder. This setup, aligned with the process outlined in Fig.~\ref{fig:kg4mmr}, integrates non-visual knowledge into VQG and employs FVQA for its evaluation.
KECVQG~\cite{DBLP:conf/mm/ChenGX0023} utilizes a causal graph to analyze and correct spurious correlations in VQG by linking unbiased features with external knowledge, thereby disentangling visual features to lessen the impact of these correlations.

Unlike VQA, VQG methods prioritize evaluation on meaningfulness, logical soundness, and consistency with target knowledge over strict correctness~\cite{DBLP:conf/wacv/UeharaH23}, often using NLP-style metrics like BLEU and CIDEr for assessment.

\begin{discussion}
Evolving intelligent dialogues with chatbots  remains a critical objective in VQG, particularly in empowering robots to formulate precise, knowledge-enriched questions that boost problem-solving capabilities for future advancements. 
Equally important is the progression towards more interactive KG-aware VQG systems, which can dynamically adapt their questioning strategies based on user interactions and feedback, marking a significant direction for future research. Moreover, with the ongoing rapid developments in VQA, transferring and adapting common problems and methods from VQA to VQG can catalyze further innovative breakthroughs in question generation technologies.
\end{discussion}

\subsubsection{Visual Dialog}\label{sec:kgvd}
VD~\cite{DBLP:journals/corr/abs-2207-07934} extends the VQA task  by adopting a multi-round format where a continuous series of $Q$-$A$ pairs revolves around a single image. This setting shifts from the single-question focus of VQA to a dynamic, conversational interaction about the image, posing a challenge for agents to adaptively interpret evolving relationships among visual elements based on the dialogue context. 

VD methods typically leverage historical dialogue information as background knowledge~\cite{DBLP:conf/cvpr/GuoWZZW20,DBLP:conf/mm/JiangDQSY20,DBLP:conf/emnlp/KangPLZK21,DBLP:conf/mm/WangWJ22,DBLP:journals/tcsv/ZhaoLGRSS23}, employing visual graph construction, query-guided relation selection and GNN propagation for dialog reasoning. 
Guo et al. \cite{DBLP:conf/cvpr/GuoWZZW20,DBLP:journals/pami/GuoWW22} introduce $Q$-conditioned attention to aggregate textual context from dialogue history, constructing a context-aware object graphs for $Q$-guided message passing. Similarly, KBGN~\cite{DBLP:conf/mm/JiangDQSY20}  uses cross-modal GNNs to bridge modal gaps and capture inter-modal semantics, retrieving information relevant  to the current question from both vision and text sources.

Addressing the limitations of relying solely on internal knowledge from images and dialog history, some approaches integrate commonsense knowledge for enhanced conversational depth. These methods all align with the \textit{KG-aware Understanding and Reasoning} paradigms we have previously outlined (Fig.~\ref{fig:kg4mmr}). 
For example,
\textbf{\textit{(\rmnum{1})} Knowledge Retrieval}: SKANet~\cite{DBLP:conf/icmcs/ZhaoG0SS21} 
integrates commonsense knowledge from ConceptNet into VD by using concept recognition and n-gram matching techniques to build a sub-KG. 
\textbf{\textit{(\rmnum{2})} Knowledge Representation}:
KACI-Net~\cite{DBLP:conf/mir/Zhang0L23} 
selects triplets with at least two entities or relations mentioned in a questionand transforms them into textual format for subsequent processing.
\textbf{\textit{(\rmnum{3})} Knowledge-aware Modality Interaction}: RMK~\cite{DBLP:conf/cvpr/ZhangJYWQ22}
utilizes caption-based dense retrieval to fetch relevant facts from ConceptNet, injecting knowledge into the dialogues through sentence-level and graph-level cross-modal attention and embedding concatenation.
\textbf{\textit{(\rmnum{4})} Knowledge-aware Answer Determination}: Acknowledging the issue of spurious correlations from unobserved confounders in retrieved knowledge, Liu et al. \cite{liu2023counterfactual} construct a counterfactual commonsense-aware VD causal graph. This graph applies counterfactual reasoning to mitigate commonsense bias, reducing the effect of misleading or inaccurate commonsense in answer derivation.

\begin{discussion}
Currently, the emphasis in knowledge-based VD mainly lies in using external commonsense knowledge, while other knowledge types, such as scientific and situational, have been relatively underexplored.  However, the rise of LLMs is diminishing the distinction between VD and VQA, with In-context Learning techniques in VQA starting to overshadow the traditional role of context in dialogues.  This shift prompts a need to reassess VD's unique contribution and its path forward.
As the boundaries between VD and VQA continue to merge, identifying and articulating the distinct potential of VD becomes imperative.
\end{discussion}

%% file: tab/kgvqa-bm.tex

\begin{table*}[!htbp]
    \centering 
    \renewcommand\arraystretch{1.0}
        \caption{Comparison of Knowledge-based VQA accuracy results on OKVQA \cite{DBLP:conf/cvpr/MarinoRFM19} and FVQA \cite{DBLP:journals/pami/WangWSDH18}.  The icon {\small \faToggleOn} represents KG-based VQA methods; {\small \faToggleOff} indicates methods without KG utilization. The $\dagger$ symbol signifies methods pre-trained on VQA2.0 or similar datasets. $\star$ indicates results reported on dataset version 1.1, which differs from version 1.0 in answer stemming methods.  Abbreviations used: Q (Question), V (Visual), w/ (with), KG (Knowledge Graph), CN (ConceptNet), WP (Wikipedia), WC (WebChild), WD (Wikidata), DBP (DBpedia), VG (VisualGenome), YG (YAGO), HP (hasPart KB), AT (ATOMIC \cite{DBLP:conf/aaai/SapBABLRRSC19}), AS (Ascent \cite{DBLP:conf/www/NguyenRW21}), VLM (Visual-Language Model),  GNN (Graph Neural Network), GAT (Graph Attention Network), MRC (Machine Reading Comprehension), MHA (Multi-head Attention), DMN (Dynamic Memory Network), DPR (Dense Passage Retriever), FiD (Fusion-in-Decoder), In-context Learning (ICL), GI (Google Image), GS (Google Search), Enc.Dec.(Encoder-Decoder), DC (Discrimination), IE (Information Extraction), CLS (Classification), TG (Text Generation), WIT (Wikipedia-Image-Text \cite{DBLP:conf/sigir/Srinivasan0CBN21}). 
        For methods employing both PLM intrinsic knowledge and external KB, only the external KB is listed in the knowledge source.
    }
    \label{tab:kvqa-bm}
    \resizebox{1.0\linewidth}{!}{
    \begin{NiceTabular}{@{}l|l|c|c|c|c|c}
    \CodeBefore
    \rowcolors{2}{gray!15}{white}
    \columncolor{gray!1}{1}
    \rowcolor{gray!40}{1}
    \Body
        \toprule[0.8pt]
        & \multicolumn{1}{c}{\textbf{Methods}} & \textbf{Approaches (Paradigms)} & \textbf{Key Idea} & \textbf{Knowledge Source} & \textbf{FVQA} & \textbf{OKVQA}  \\
        \midrule[0.8pt]
        \parbox[b]{2mm}{\multirow{17}{*}{\rotatebox[origin=c]{90}{ 2018$\sim$2021}}}
        & {\small \faToggleOn} QQmaping \cite{DBLP:journals/pami/WangWSDH18} & RDF Query (IE) & Question-Query Mapping & DBP / CN / WC & 56.91 & -  \\
        & {\small \faToggleOn} STTF \cite{DBLP:conf/eccv/NarasimhanS18} & Relation Query (IE)  & Scoring the Facts & DBP / CN / WC & 62.20 & -  \\
        & {\small \faToggleOn} OB \cite{DBLP:conf/nips/NarasimhanLS18} & Fact Retrieval + GNN (DC)  & Entity Graph + GCN & DBP / CN / WC & 69.35 & -  \\
        & {\small \faToggleOff} Marino et al. \cite{DBLP:conf/cvpr/MarinoRFM19} & Retrieval + ArticleNet (IE)  & Web Retrieval + Knowledge Span Prediction  & WP & - & 27.84  \\
        & {\small \faToggleOn} KG-Aug \cite{DBLP:conf/mm/Li0020} & Fact Retrieval + GNN (CLS)  & Augment Q\&V Features w/ Entity Embedding & WD / CN & 38.58 & 26.71  \\
        & {\small \faToggleOn} Chen et al. \cite{DBLP:conf/semweb/0007CGPYC21} & Alignment + Re-rank (CLS)  & KG-aware Answer Masking for Validation & DBP / CN / WC & 58.27 & -  \\
        & {\small \faToggleOn} ERMLP \cite{DBLP:journals/corr/abs-2012-15484} & KGE + Attention (IE)  & Knowledge-guided Co-attention & DBP / CN / WC & 60.82 & -  \\
        & {\small \faToggleOn} Liu et al. \cite{DBLP:journals/kbs/LiuWHQC22} & Fact Retrieval + GNN (DC)  & Q-V Guided Cross-modal GAT & DBP / CN / WC & 63.56 & 29.43  \\
        & {\small \faToggleOn} KAN \cite{DBLP:journals/tnn/ZhangLLZLSG21} & Retrieval + Attention (CLS)  & Question-guided MHA & CN & 66.39 & -  \\
        & {\small \faToggleOn} Mucko \cite{DBLP:conf/ijcai/ZhuYWS0W20} & Fact Retrieval + GNN (DC) & Question-guided Attention + Cross-KG GAT & DBP / CN / WC & 73.06 & 29.20 \\
        & {\small \faToggleOn} GRUC \cite{DBLP:journals/pr/YuZWZHT20} & Fact Retrieval + DMN (DC) & Fact-centered DMN + GRU & DBP / CN / WC & 79.63 & 29.87 \\
        & {\small \faToggleOn} ConceptBert \cite{DBLP:conf/emnlp/GarderesZAL20} & GCN + Attention (CLS)  & Compact Trilinear Interaction + MHA & CN & - & 33.66 \\
        & {\small \faToggleOn} KRISP$\dagger$ \cite{DBLP:conf/cvpr/MarinoCP0R21} & GCN + VLM (CLS)  & Global KG + RGCN + VisualBERT & HP / DBP / CN / VG & - & 38.90$^\star$ \\
        & {\small \faToggleOn} MAVEx$\dagger$ \cite{DBLP:conf/aaai/WuLSM22} & Multi-retrieval + Re-rank (CLS)  & Web Retrieval + VilBERT + Answer Validation & WP / CN / GI & - & 38.70$^\star$  \\
        & {\small \faToggleOn} RVL$\dagger$ \cite{DBLP:journals/corr/abs-2101-06013} & KGE Alignment + VLM (CLS)  & Aligning VLM Text Embedding w/ KGE & WD / CN / PLM & 54.27 & 39.04 \\  
        & {\small \faToggleOff} Luo et al. $\dagger$ \cite{DBLP:conf/emnlp/LuoZBB21} & Dense Retriever + PLM (IE)  & RoBERTa + DPR  Learning + MRC  & GS & - & 39.20$^\star$  \\
        & {\small \faToggleOn} PGVQA \cite{DBLP:journals/pr/SongLLYSS23} & Retrieval + Re-rank (CLS)  & KG-aware Answer Refinement & DBP / CN / WC & - & 41.07  \\
        \midrule
        \parbox[b]{2mm}{\multirow{15}{*}{\rotatebox[origin=c]{90}{ 2022}}}
        & {\small \faToggleOn} SUPER \cite{DBLP:journals/tip/HanYWWN23} & Multi-modules + DMN (CLS) & Q-V Guided Modular Routing + DMN & CN & 48.90 & 30.46  \\
        & {\small \faToggleOn} MKRE \cite{DBLP:conf/sitis/HussainMSF22} & Fact Retrieval + GNN (DC)  & Question Guided Attention + Cross-KG GAT & DBP / CN / WC & 73.06 & -  \\
        & {\small \faToggleOn} DMMGR \cite{DBLP:conf/aaai/LiM22} & Retrieval + DMN + GNN (CLS) & Caption + Multi-scale DMN + GAT & DBP / CN / WC & 81.20 & -  \\
        & {\small \faToggleOff} CBM-BERT$\dagger$  \cite{DBLP:journals/eswa/SalaberriaALSA23} & Caption + PLM (CLS)  & Caption + BERT + Ensemble & PLM & - & 36.00$^\star$  \\
        & {\small \faToggleOff} CBM-T5$\dagger$  \cite{DBLP:journals/eswa/SalaberriaALSA23} & Caption + PLM (TG)  & Caption + T5 + Ensemble & PLM & - & 40.80$^\star$  \\
        & {\small \faToggleOn} UnifER$\dagger$ \cite{DBLP:conf/mm/GuoNWLCK22} & Fact Retrieval + VLM (CLS) & Loss Gap Driven DPR Learning + ViLT & CN & - & 42.13$^\star$  \\
        & {\small \faToggleOff} MuKEA$\dagger$ \cite{DBLP:conf/cvpr/DingYLHC022} & VLM + KGE  & LXMERT + Multi-modal TransE & VLM & - & 42.59$^\star$  \\
        & {\small \faToggleOff} PICa$\dagger$ \cite{DBLP:conf/aaai/YangGW0L0W22} & Caption + Decoder (TG) & Caption + Tag + ICL + GPT3 & GPT3 & - & 43.30$^\star$  \\
        & {\small \faToggleOn} KVQAmeta$\dagger$ \cite{DBLP:conf/www/Garcia-OlanoOG22} & KGE Alignment + VLM (CLS)  & Aligning VLM Embedding w/ Wikipedia2Vec & WP & - & 43.67$^\star$  \\
        & {\small \faToggleOn} LaKo$\dagger$ \cite{DBLP:conf/jist/0007HCGFP0Z22} & Retrieval + Enc.Dec. (TG)  & Caption + Fact DPR + T5 + FiD  & HP / DBP / CN / WC & - & 47.01$^\star$  \\
        & {\small \faToggleOff} TRiG$\dagger$ \cite{DBLP:conf/cvpr/0013PTRWN22} & Retrieval + Enc.Dec. (TG) & Caption + Tag + DPR + FiD & WP & - & 49.24$^\star$  \\
        & {\small \faToggleOn} KAT$\dagger$ \cite{DBLP:conf/naacl/GuiWH0BG22} & Retrieval + Enc.Dec. (TG)  & Caption + Tag + ICL + GPT3 + DPR + FiD & WD / GPT3 & - & 53.09 $^\star$ \\
        & {\small \faToggleOn} EnFoRe$\dagger$ \cite{DBLP:conf/emnlp/WuM22} & Retrieval + Enc.Dec. (TG)  & KAT + Entity Focused DPR & WD / GPT3 & - & 54.35$^\star$  \\
        & {\small \faToggleOff} RAVQA$\dagger$ \cite{DBLP:conf/emnlp/LinB22} & Retrieval + Enc.Dec. (TG) & DPR Learning + Ensemble &  GS & - & 54.48$^\star$  \\
        & {\small \faToggleOn} REVIVE$\dagger$ \cite{DBLP:conf/nips/LinX0X0Y22} & Retrieval + Enc.Dec. (TG) & KAT + Regional Visual & WD / GPT3 & - & 56.604$^\star$  \\
        \midrule
        \parbox[b]{2mm}{\multirow{15}{*}{\rotatebox[origin=c]{90}{ 2023}}}
        & {\small \faToggleOn} VLC-BERT$\dagger$ \cite{DBLP:conf/wacv/RaviCSLS23} & Fact Generation + VLM (CLS)  & COMET Generation + MHA + VL-BERT & CN / AT & - & 43.14$^\star$  \\
        & {\small \faToggleOff} DEDR$\dagger$ \cite{DBLP:conf/sigir/SalemiPZ23} & Retrieval + Enc.Dec. (TG)  & Mutual Retriever Distillation + VL-T5 + FiD & WP & 61.80 & 44.57$^\star$  \\
        & {\small \faToggleOn} RR-VEL$\dagger$ \cite{DBLP:conf/icmcs/YouYLL23} & EL + Retrieval + Enc.Dec. (TG) & Ground Truth Referent in Q + T5 & HP / CN / Ascent & 65.59 & 49.48$^\star$  \\
        & {\small \faToggleOn} MCR-MemNN \cite{yin2023multi} & Fact Retrieval + DMN (CLS) & Multi-clue Reasoning + Fact-centered DMN & DBP / CN / WC & 70.92 & -  \\
        & {\small \faToggleOff} CodeVQA$\dagger$ \cite{DBLP:conf/acl/SubramanianNKYN23} & Code Generation + PLM (TG) & ICL + Codex + Modular Combination & Codex / VLM & - & 53.50$^\star$  \\
        & {\small \faToggleOff} TwO$\dagger$ \cite{DBLP:conf/acl/SiMLJW23} & Retrieval + Enc.Dec. (TG) & KAT + OFA Multi-modal Knowledge  & WP / GPT3 & - & 57.57$^\star$  \\
        & {\small \faToggleOff} PROOFREAD$\dagger$ \cite{DBLP:journals/corr/abs-2308-15851} & Decoder + Re-rank (TG) & Answer-aware Knowledge Generation \& Filter & ChatGPT & - & 57.60$^\star$  \\
        & {\small \faToggleOn} REVEAL$\dagger$ \cite{DBLP:conf/cvpr/HuI0WCSSRF23} & Retrieval + Enc.Dec. (TG)  & Multi-modal Retrieval + Gate + DMN + T5 & WIT / WD / VQA2.0 &  -  & 59.10$^\star$  \\
        & {\small \faToggleOn} MM-Reasoner$\dagger$ \cite{DBLP:conf/emnlp/KhademiYFZ23} & Fact Generation + Enc.Dec. (TG) & Vision APIs + ICL Rationales Generation & GPT-4 &  61.10  & 59.20$^\star$  \\
        & {\small \faToggleOff} Wang et al.$\dagger$ \cite{DBLP:journals/corr/abs-2311-11598} & Q-decomposition + Decoder (TG)  & Q Decomposition + Fact Refinement + PICa  & ChatGPT & - & 59.34$^\star$  \\
        & {\small \faToggleOff} PromptCap$\dagger$ \cite{DBLP:journals/corr/abs-2211-09699} & Caption + Decoder (TG) & Q-guided Caption Generation  + PICa & GPT-3 &  -  & 60.40$^\star$  \\
        & {\small \faToggleOff} Prophet$\dagger$ \cite{DBLP:conf/cvpr/Shao0W023} & Caption + Decoder (TG)  & MCAN + Answer Pruning + Answer-aware ICL & GPT3 & - & 61.10$^\star$  \\
        & {\small \faToggleOff} ASB$\dagger$ \cite{DBLP:journals/corr/abs-2310-13570} & Caption + Decoder (TG) & Q-guided Patch Caption Selection + PICa & LLaMA-13B & - & 61.20$^\star$  \\
        & {\small \faToggleOff} Cola-FT$\dagger$ \cite{DBLP:journals/corr/abs-2310-15166} & Decoder (TG)  & OFA + BLIP + LLM Answer Decision & FLAN-T5  & - & 62.40$^\star$  \\
        & {\small \faToggleOff} GPT-4V$\dagger$ \cite{DBLP:journals/corr/abs-2311-07536} & Decoder (TG)  & Prompt + GPT-4V & GPT-4V & - & 64.28$^\star$  \\
        \bottomrule[0.8pt]
    \end{NiceTabular}
    \vspace{-4pt}
    }
\end{table*}

%% file: 4.3-cls.tex
\subsection{Classification Tasks}\label{sec:kgcls}
This section focuses on classification tasks that leverage multi-modal inputs, particularly text and image combinations, with a focus on knowledge-based Image Classification.  Discussions also extend to related multi-modal tasks like Fake News Detection and Movie Genre Classification, highlighting the diversity and wide-ranging applications within the field.

\begin{defn}{\textbf{KG-aware Classification.}}\label{def:kgcls}
Consider a set of labeled training samples $\mathcal{D}_{tr}=\{(x,y)| x \in \mathcal{X}, y \in \mathcal{Y}\}$ with a background KG defined as $\mathcal{G} = \{\mathcal{E},\mathcal{R}, \mathcal{T}\}$,
a classifier aims to approximate a function $f: x \rightarrow y$ from input $x$ to output label $y$ with the assist of $\mathcal{G}$. This function should accurately predict the labels of samples in a testing set $\mathcal{D}_{te} = \{(x,y)| x \in \mathcal{X}', y \in \mathcal{Y}\}$, with $\mathcal{X} \cap \mathcal{X}' = \emptyset$. 
\end{defn}

\subsubsection{Image Classification} \label{sec:kgimgc}

\begin{figure*}[!htbp]
  \centering
   \vspace{-1pt}
\includegraphics[width=0.85\linewidth]{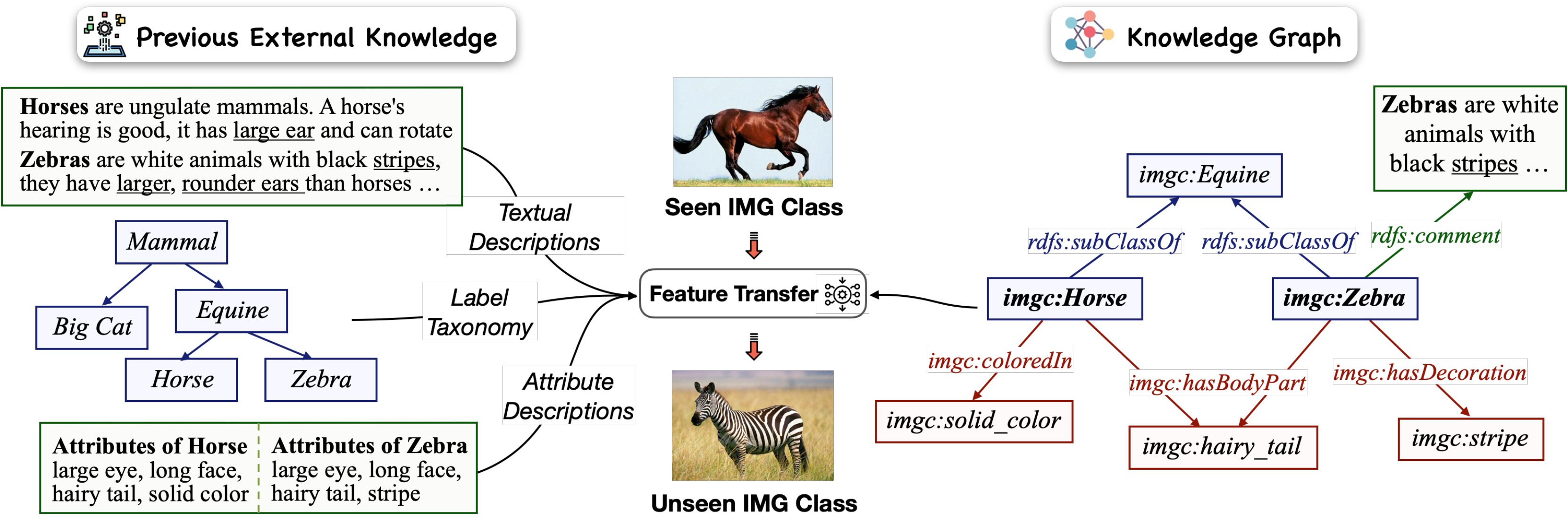}
  \caption{Comparison of previously used external knowledge (Left) and KG (Right) in Zero-Shot Image Classification task~\cite{DBLP:conf/www/GengC0PYYJC21}.}
  \label{fig:IMGC_KG_example}
  \vspace{-10pt}
\end{figure*}

Image Classification (IMGC) aims to identify objects within images and, with deep learning advancements, has even surpassed human performance in challenges like ImageNet ILSVRC \cite{russakovsky2015imagenet}. 
Traditional IMGC follows a closed-world assumption, requiring extensive labeled images for both training and testing within known classes, i.e., $\mathcal{Y}_{tr} = \mathcal{Y}_{te} = \mathcal{Y}$. However, it is not feasible for newly emerging classes due to the impracticality of continuously annotating and retraining models with sufficient images for these classes. Consequently, there is a growing interest in Zero-Shot Image Classification (ZS-IMGC), which supports classifying images of novel, unseen classes without the need for specific training images, i.e., $\mathcal{Y}_{tr} \cap \mathcal{Y}_{te} = \emptyset$.

To handle these unseen classes, most existing ZS-IMGC methods adopt a knowledge transfer strategy~\cite{DBLP:journals/pieee/ChenGCPHZHC23,DBLP:conf/ijcai/ChenG0HPC21}: transferring labeled images, image features or model parameters from the seen classes in training set to unseen classes,  guided by external knowledge that describes semantic relationships between classes.
For instance, as illustrated in the left part of Fig.~\ref{fig:IMGC_KG_example}, consider the description of a ``\textit{Zebra}'' as an animal with a horse-like body, tiger-like stripes, and black-and-white colors similar to a panda. Models can infer the appearance of a ``\textit{Zebra}'' by combining features of these seen animals,  even without direct exposure to its images.
Briefly, ZS-IMGC relies on data from observed classes and class-specific semantic knowledge, with the external knowledge frequently embodying a modality distinct from the image data. 
This section reviews KG-based ZS-IMGC efforts to illustrate the typical practice of multi-modal learning in IMGC.

Early ZS-IMGC works like \cite{frome2013devise,zhu2018generative} use textual class descriptions or names to model inter-class relationships. Others, such as \cite{xian2018feature,lampert2013attribute}, utilize class attributes, annotating each class with descriptive characteristics, thereby defining semantic relationships through shared attributes. However, these approaches sometimes face limitations in capturing complete semantics \cite{DBLP:journals/ws/GengCZCPLYC23}.
Recently, KGs have gained prominence in ZS-IMGC for their ability to encapsulate diverse and explicit class semantics. For example, studies like \cite{wang2018zero,kampffmeyer2018rethinking} integrate hierarchical relationships from WordNet, while \cite{roy2022improving,nayak2020zero,gao2019know} explore class knowledge from commonsense KGs such as ConceptNet. KGs, due to their compatibility, can unify various knowledge forms, including textual description and discrete attributes, into a single graph \cite{DBLP:journals/ws/GengCZCPLYC23,DBLP:conf/kdd/GengCZXCPHXC22,DBLP:conf/www/GengC0PYYJC21}, as depicted in the right part of Fig.~\ref{fig:IMGC_KG_example}. Furthermore, ontologies can be utilized to define complex class relationships~\cite{DBLP:conf/kr/ChenLGPC20}, such as \textit{disjointness}, enhancing classification performance significantly.
Pahuja et al. \cite{pahuja2023bringing} structure species classification within an MMKG, treating it as a link prediction task. They leverage multi-modal contexts, like visual cues and GPS coordinates, to improve efficiency in identifying unseen classes. For example, ``\textit{an image of a feline captured in Africa is more likely to be a lion}''.
In KG-aware ZS-IMGC, a KG is defined as $\mathcal{G} = {\mathcal{E},\mathcal{R}, \mathcal{T}}$, with $\mathcal{Y} \subset \mathcal{E}$. This paradigm, representing semantic hierarchical relationships among classes, is instrumental in augmenting classification performance and interpretability.

According to the ways of exploiting KGs to guide the feature transfer from seen to unseen classes, existing KG-driven ZS-IMGC approaches can be divided into three groups: \textit{Mapping-based}, \textit{Data Augmentation}, and \textit{Propagation-based}.

\textbf{\textit{(\rmnum{1})}} \textbf{\ul{Mapping-based}} methods focus on developing mapping functions that align image inputs and KG-based class semantics into a shared vector space. They typically identify the closest class to a test image in this space using similarity-related metrics.
For instance, HierSE~\cite{li2015zero} employs a linear projection to map image features into a class embedding space derived from word embeddings of the class and its superclasses, using cosine similarity for comparison. Chen et al. \cite{DBLP:conf/kr/ChenLGPC20} employ an OWL-based ontology for animal classes, encoding it via the OWL EL embedding method.
Akata et al. \cite{akata2015label,akata2013label} map initial class encodings into the image feature space, representing classes as multi-hot vectors of ancestors based on class hierarchies. 
There are also some joint mapping methods that map both the class encoding and the image features. For example, DUET~\cite{DBLP:conf/aaai/ChenHCGZFPC23}, an end-to-end Transformer-based ZSL method, leverages cross-modal PLMs for fine-grained visual characteristic reorganization and discrimination with structured KGs serialized as input.
In summary, mapping-based methods, employing linear or nonlinear networks for transformation, are relatively simple to implement but inherently biased towards seen classes when trained exclusively on their images. This bias becomes evident in generalized ZSL scenarios, where seen and unseen classes coexist, underlining a fundamental limitation of these methods.

\textbf{\textit{(\rmnum{2})}} \textbf{\ul{Data augmentation}} methods address the ZSL sample shortage by generating images or features for unseen classes, transforming ZS-IMGC into a supervised learning problem and mitigating bias issues. They mainly use generative models like Generative Adversarial Networks (GANs) and Variational Auto-encoders (VAEs).
OntoZSL~\cite{DBLP:conf/www/GengC0PYYJC21} generates image features conditioned on a attribute-and-species KG and combines class embeddings with a random noise vector to synthesize features for each class. This process is supervised by real features from annotated images, with an adversarial discriminator distinguishing between real and generated features.  DOZSL~\cite{DBLP:conf/kdd/GengCZXCPHXC22} further employs a disentangled KG embedding module to enhance the quality of synthesized image features. 
TGG~\cite{zhang2019tgg} generates few-shot samples, where a GAN-based generation module is applied to generate an image-level graph. Zhang et al.~\cite{DBLP:conf/ictai/ZhangSBH22} develop a MMKG by merging visual representations of classes with word embeddings, creating multi-modal class nodes and edges that explicitly model class correlations, thereby enhancing information transfer from seen to unseen class nodes.

\textbf{\textit{(\rmnum{3})}} \textbf{\ul{Propagation-based}} methods leverage KG-structured inter-class relationships for knowledge transfer, aligning seen and unseen classes with KG entities and using GNNs to propagate features from seen to unseen class nodes.
For single-labeled images, GNN models typically train to output a class-specific parameter vector as a classifier for each class, estimating unseen class classifiers by aggregating those of neighboring seen classes on the graph. A notable example is the work by Wang et al.~\cite{wang2018zero}, with subsequent studies adopting similar concepts but varying in graph propagation optimization~\cite{kampffmeyer2018rethinking,geng2020explainable}. Certain methods tackle KGs' diverse relation types by using multi-relational GCNs~\cite{chen2020zero} or dividing multi-relation KGs into single-relation graphs with parameter-shared GCNs for feature propagation \cite{DBLP:conf/kdd/GengCZXCPHXC22,wang2021zero,DBLP:conf/kdd/WuLZWLHYC23}. 
For multi-label images, where models assign a probability score per class, GNNs utilize KG-implied correlations to propagate these scores from seen to unseen class nodes, as done by Lee at al.~\cite{lee2018multi}.

\textbf{RESOURCES:} For ZS-IMGC research, particularly KG-aware ZS-IMGC, several open datasets and KG resources have been proposed:

\textbf{\textit{(\rmnum{1})}} \textbf{ImageNet} \cite{DBLP:conf/cvpr/DengDSLL009}:
A large-scale database with 14M images across 21K classes each aligned with a WordNet \cite{miller1995wordnet} entity. It leverages class hierarchies as KG-based knowledge, where the graph only contains one type of relation, i.e., \textit{subClassOf}. From \cite{xian2018zero}, a subset of 1K classes serves as seen classes, with unseen classes determined by their distance in the WordNet graph. Note that ImageNet is widely used for ZS-IMGC benchmarking, albeit with a single-relational KG limitation.

\textbf{\textit{(\rmnum{2})}} \textbf{ImNet-A} and \textbf{ImNet-O}:
Subsets of ImageNet by Geng et al.~\cite{DBLP:conf/www/GengC0PYYJC21,DBLP:journals/ws/GengCZCPLYC23}. ImNet-A contains 80 animal classes, and ImNet-O has 35 general object classes. Each comes with a KG combining multiple knowledge types, including class attribute, class name,  commonsense knowledge from ConceptNet, class hierarchy (taxonomy) from WordNet, and logical relationships such as \textit{disjointness}.

\textbf{\textit{(\rmnum{3})}} \textbf{AwA2} \cite{xian2018zero}:
A coarse-grained animal classification dataset with 50 animal classes and 37,322 images, plus 85 expert-annotated attributes. Its classes align with WordNet entities for taxonomy-based KGs. Geng et al. \cite{DBLP:journals/ws/GengCZCPLYC23} equip AwA2 with a KG similar to ImNet-A and ImNet-O. 
Chen et al. \cite{DBLP:conf/kr/ChenLGPC20} utilize OWL2 to model complex class semantics.

\textbf{\textit{(\rmnum{4})}} \textbf{NUS-WIDE} \cite{chua2009nus}:
A mutli-label image classfication dataset, where each image contains multiple objects.
In works like \cite{lee2018multi}, NUS-WIDE is accompanied by a KG with three types of label relations, including a super-subordinate correlation from WordNet, positive and negative correlations computed by label similarities such as WUP similarity. 

\input{tab/zs-imgc}

\textbf{BENCHMARKS:} 
Single-label image classification in ZS-IMGC includes two evaluation settings:
\textbf{\textit{(\rmnum{1})} Standard ZSL:} Focuses solely on unseen class samples, using \textit{Macro Accuracy}, which is calculated as the average of individual class accuracies (correct predictions to total samples ratio), as the metric.
\textbf{\textit{(\rmnum{2})} Generalized ZSL (GZSL):} Evaluates both seen and unseen class samples, hence more challenging. Here, two \textit{Macro Accuracies}, $Acc_s$ for seen classes and $Acc_u$ for unseen classes, are  measured. The key performance indicator is the harmonic mean  $H = (2 \times Acc_s \times Acc_u) / (Acc_s + Acc_u)$, ensuring a balance between the two.


\begin{discussion}
Incorporating diverse class semantics generally yield better ZS-IMGC results, even with basic methods.
For instance, as Table \ref{tab:IMGC_results} illustrates that some mapping-based methods \cite{frome2013devise,DBLP:conf/kr/ChenLGPC20}, employing straightforward linear mapping functions, achieve better performance by integrating various class semantics such as attributes, hierarchy, and names, compared to $\text{GCNZ}^{\dag}$, which relies solely on class hierarchy. Significantly, enriching $\text{GCNZ}^{\dag}$ with more class semantics, as demonstrated by $\text{GCNZ}^{\ddag}$, markedly enhances its effectiveness.
Furthermore, KG-based ZS-IMGC methods typically operate in a class transductive setting, where unseen classes are known during training, in contrast to the conventional inductive approach that utilizes only seen class knowledge.  These methods leverage a KG to bridge seen and unseen classes through semantic links.
Additionally, although the generalized ZSL setting is well-recognized by many researchers, some studies adopt their own definitions, specifically testing only unseen class images while classifying them within a combined pool of both seen and unseen classes. This variation requires careful consideration in future work.
\end{discussion}

\subsubsection{Fake News Detection}\label{sec:kgfnd}
FND, also termed Rumor Detection, addresses the proliferation of misleading multimedia content on social media to ensure the dissemination of trustworthy information. Unlike standard text classification, FND challenges involve discerning falsehoods across diverse subjects. While traditional deep learning approaches in FND emphasize text, they frequently neglect the significance of visual content and background knowledge. Thus, a comprehensive integration of text, visuals, and knowledge is essential for precise FND.

KMGCN~\cite{DBLP:conf/mir/WangQHFX20} employs Entity Linking to map entities from social media posts to concepts from Probase \cite{DBLP:conf/sigmod/WuLWZ12} and YAGO KGs. It constructs a graph with post words as nodes and incorporates visual words from images (detected by a pre-trained object detector~\cite{DBLP:journals/corr/abs-1804-02767}), weighting edges with Point-wise Mutual Information to emphasize word correlations. A GCN is then utilized to model semantic interactions, employing global mean pooling for the final binary classification of multimedia posts.
Extending these insights, KMAGCN~\cite{DBLP:journals/tomccap/QianHFX21} integrates the visual modality through a late fusion paradigm, employing feature-level attention to more accurately delineate the interplay between visual and textual content.
As a dual-consistency network, KDCN~\cite{DBLP:journals/tkde/SunZMXLY23} identifies inconsistencies in both cross-modal and content-knowledge aspects with Freebase as the reference KG. It finds that entities in rumor posts are more distantly connected within KGs compared to non-rumors, providing a clear marker for distinction.
EmoKnow~\cite{DBLP:conf/adma/ZhangSWYFZ23} advances COVID-19 FND by incorporating WiKiData5M \cite{DBLP:journals/tacl/WangGZZLLT21} as an external knowledge source. 
It uses PLMs for text analysis, extracts emotion features, and identifies relevant linked entities, utilizing TransE~\cite{DBLP:conf/nips/BordesUGWY13} for entity representation, with an MLP-based classifier to combine these multi-modal inputs.

\begin{discussion}
The progression of LLMs is transforming many classification tasks into ones focused on inference and directive question-answering, emphasizing the crucial role of selecting knowledge sources. 
Moreover, given the frequent association of fake news with political content, their urgency of timely news highlights the need for conducting research on knowledge updating and lifelong learning in FND.
\end{discussion}

\subsubsection{Movie Genre Classification}\label{sec:mmgc}
MMGC models integrate visual, textual, and metadata information to predict movie genres, representing each genre as an element in a binary vector for multi-genre classification of movies.

Traditional methods~\cite{DBLP:conf/accv/BainNBZ20,DBLP:conf/eccv/HuangXRWL20} primarily rely on features extracted from images and texts alone.  A recent work, IDKG~\cite{DBLP:conf/mm/LiQZ0TXT23}, integrates a domain-specific MMKG created from metadata fields such as titles, genres, casts, and directors. Its motivation stems from recognizing the relational patterns in metadata, like the tendency for ``\textit{Nolan to direct science fiction movies}'' or ``\textit{Emma to not often feature in comedies}''. It use a translation model to merge KG embeddings with other modality features, guided by attention mechanism.

\begin{discussion}
The success of a domain-specific KG largely depends on the metadata's quality and completeness, where addressing scalability is vital especially for large movie datasets.  Additionally, there is scope for creating interactive and personalized Movie Genre Classification systems. By integrating user feedback and preferences, the system can be tailored to individual tastes, offering personalized genre suggestions. Techniques such as reinforcement learning and user modeling could be utilized to customize the genre classification process, thus further enhancing user experience and satisfaction.
\end{discussion}

%% file: tab/zs-imgc.tex
\begin{table}[!htbp]
\renewcommand{\arraystretch}{0.95}
\renewcommand{\ttdefault}{pcr}
\caption{Comparison of ZS-IMGC results across various datasets. We use $Acc$ and $H$ as the evaluation metrics of Standard ZSL and Generalized ZSL, respectively. DeViSE here is implemented with a KG which covers the semantics of class hierarchy, class attributes, attribute hierarchy and class names (see \cite{DBLP:journals/ws/GengCZCPLYC23} for details). $\text{GCNZ}^{\dag}$ utilizes a KG with class hierarchy only, whereas $\text{GCNZ}^{\ddag}$ includes broader class semantics like attributes, as reported in \cite{DBLP:journals/ws/GengCZCPLYC23}.  DOZSL (GAN) and DOZSL (GCN) are variants of DOZSL \cite{DBLP:conf/kdd/GengCZXCPHXC22}, representing generation-based and propagation-based ZSL learners, respectively. ImageNet results are tested on 2-hops unseen classes.}
\centering
\resizebox{0.8\linewidth}{!}{
\begin{NiceTabular}{clcc}
\CodeBefore
\rowcolors{2}{gray!10}{white}
\columncolor{gray!1}{1}
\rowcolor{gray!30}{1}
\Body
\toprule[0.8pt]
\textbf{Dataset} & \multicolumn{1}{c}{\textbf{Methods}} & {$\bm{Acc}$} \textbf{(\%)} &{$\bm{H}$} \textbf{(\%)}  \\
\midrule[0.8pt]
\multirow{3}{*}{ImageNet} 
& GCNZ \cite{wang2018zero} & 19.8 & -- \\ 
& DGP \cite{kampffmeyer2018rethinking} & 26.6 & -- \\ 
& FGP \cite{DBLP:conf/kdd/WuLZWLHYC23} & 26.4 & --\\
\midrule
\multirow{6}{*}{ImNet-A} 
& DeViSE \cite{frome2013devise} & 33.62 & 26.01 \\
& OntoZSL \cite{DBLP:conf/www/GengC0PYYJC21} & 39.00 & 32.15 \\
& DOZSL (GAN) \cite{DBLP:conf/kdd/GengCZXCPHXC22} & 40.26 & 32.82 \\
& DOZSL (GCN) \cite{DBLP:conf/kdd/GengCZXCPHXC22} & 38.69 &  32.12 \\
 & $\text{GCNZ}^{\dag}$ \cite{wang2018zero,DBLP:journals/ws/GengCZCPLYC23}  & 33.95 & 26.68 \\
 & $\text{GCNZ}^{\ddag}$ \cite{wang2018zero,DBLP:journals/ws/GengCZCPLYC23}  & 36.64 & 31.38 \\
\midrule
\multirow{9}{*}{{AwA2}}   
& DeViSE \cite{frome2013devise} & 46.12 & 15.88 \\
& AHLE \cite{akata2015label} & 49.40 &-- \\
& OWL-based \cite{DBLP:conf/kr/ChenLGPC20}   & 58.90 & --      \\ 
& DUET \cite{DBLP:conf/aaai/ChenHCGZFPC23} & -- & 58.00  \\
& OntoZSL \cite{DBLP:conf/www/GengC0PYYJC21} & 63.31 & 56.06   \\
& DOZSL (GAN) \cite{DBLP:conf/kdd/GengCZXCPHXC22} & 66.36 & 57.62 \\
& DOZSL (GCN) \cite{DBLP:conf/kdd/GengCZXCPHXC22} & 63.88  &  52.74 \\
& $\text{GCNZ}^{\dag}$ \cite{wang2018zero,DBLP:journals/ws/GengCZCPLYC23}  & 37.44 & 14.34 \\
& $\text{GCNZ}^{\ddag}$ \cite{wang2018zero,DBLP:journals/ws/GengCZCPLYC23}  & 62.98 & 31.98 \\
& FGP \cite{DBLP:conf/kdd/WuLZWLHYC23} & 79.10 & 43.30 \\
\bottomrule[0.8pt]
\end{NiceTabular}}
\label{tab:IMGC_results}
\vspace{-10pt}
\end{table}

%% file: 4.1-gen.tex
\subsection{Content Generation Tasks}\label{sec:kggen}
Many of the reasoning methods previously discussed, including those used in VQA tasks, are based on generative approaches. This section focuses on tasks where content generation is strictly necessary for task completion, highlighting the crucial role KGs play in augmenting the generation process.

\begin{defn}{\textbf{KG-aware Generation Tasks.}}\label{def:kggen}
Given visual images ($x^\mathbbm{v}$) or textual descriptions ($x^\mathbbm{l}$), the objective is to generate, in a cross-modal manner, either a textual target $y^\mathbbm{l}$ (e.g., caption), a visual target $y^\mathbbm{v}$ (e.g., image), or a graph target (e.g., scene graph), leveraging the background KG $\mathcal{G}$ for foundational support.
\end{defn}

\subsubsection{Image  Captioning}\label{sec:kgic}
\begin{figure}[!htbp]
  \centering
   \vspace{-1pt}
\includegraphics[width=0.85\linewidth]{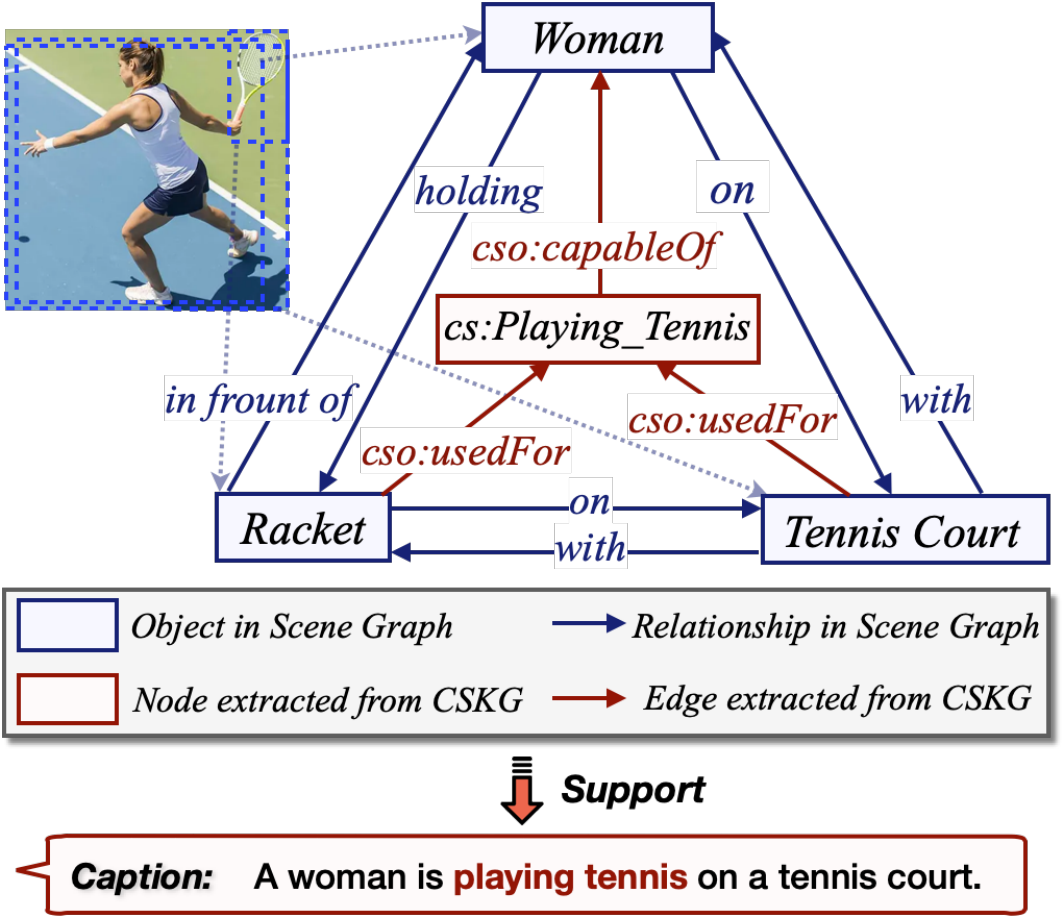}
  \caption{Scene Graph (SG) of an image equiped with CSKG~\cite{DBLP:conf/esws/IlievskiSZ21}. The SG (in blue) outlines objects and their relationships in the scene. Additional background knowledge from CSKG triples (in red) such as \textit{(Woman, capableOf, Playing\_Tennis)} and \textit{(Racket, usedFor, Playing\_Tennis)}, enriches the SG with off-scene\protect\footnotemark 
 knowledge~\cite{DBLP:conf/esws/KhanBC22}. This facilitates higher-level inferences, enabling more accurate caption deductions, e.g.,  ``\textit{A woman is playing tennis on a tennis count}''~\cite{DBLP:conf/aaai/HouWZQJL20}.}
 \vspace{-7pt}
  \label{fig:SG_example}
\end{figure}

Image Captioning (IC) \cite{DBLP:journals/csur/HossainSSL19-ICSurvey} is a pivotal multi-modal learning task, aiming to describe images in natural language. In IC, KGs can provide essential prior knowledge, including commonsense semantic correlations and constraints among objects, guiding the construction of semantic graphs for  meaningful caption generation, even when certain elements are not visually present (Fig.~\ref{fig:SG_example}). Furthermore, since each image in the training data typically comes with only a few ground truth captions, models often lack the cues necessary to uncover implicit intentions. KGs can significantly bridge this gap by offering essential fact-checking support.

\textbf{Rule-based} methods~\cite{DBLP:journals/corr/AdityaYBFA15,DBLP:conf/emnlp/LuWHJC18,DBLP:conf/cikm/HuangLCZM20}
primarily incorporate KG knowledge into caption models through Entity Linking and symbolic rules, often supplemented by inter-concept co-occurrence scores.
Aditya et al.~\cite{DBLP:journals/corr/AdityaYBFA15} pioneer the application of KG into IC, identifying relevant events from a KG based on detected visual concepts, then constructing a Scene Description Graph (SDG) with pre-defined rules, from which captions are generated using NLG tools.
Lu et al.~\cite{DBLP:conf/emnlp/LuWHJC18} use a CNN-LSTM model to create a template caption from the input image, followed by employing a KG-based collective inference algorithm to populate the template with specific named entities, sourced from hashtags.
Instead of directly integrating semantic knowledge into the neural network layers, Huang et al.~\cite{DBLP:conf/cikm/HuangLCZM20} input retrieved triples from ConceptNet during the word generation stage for next-word prediction, augmenting the probabilities of potential words identified within the semantic knowledge corpus.

\textbf{Embedding-based} methods~\cite{DBLP:journals/corr/abs-1906-01290,DBLP:conf/aaai/HouWZQJL20,DBLP:journals/tmm/LiJ19,DBLP:journals/corr/abs-2007-11690,DBLP:journals/prl/ZhangSMY21,zhong2023image-ic-add1} 
typically employ networks such as GNNs or RNNs to efficiently encode retrieved knowledge as vectors, subsequently incorporating these vectors into the caption generation process.
Hou et al.~\cite{DBLP:journals/corr/abs-1906-01290,DBLP:conf/aaai/HouWZQJL20} utilize human {commonsense knowledge} to support object relationship reasoning in IC, avoiding the need for pre-trained detectors. Using Visual Genome as an external KG, they map densely sampled regions from images into low-dimensional vectors, and then, guided by the KG, form a temporary semantic graph. This graph enhances GNN-based relational reasoning for captioning and iteratively refines the KG itself.
CNet-NIC~\cite{DBLP:conf/wacv/ZhouSH19-NICKG} connects ConceptNet entries with identified image objects to enrich descriptions and infer non-explicit visual information. This method enhances the semantic depth of object recognition module outputs, integrating the embeddings of knowledge terms and image features to initialize a RNN for IC generation.
Interpret-IC~\cite{DBLP:journals/corr/abs-2007-11690} selects local objects in an image based on human-interpretable rules, ensuring captions reflect only those objects of human interest. During training, entities not present in standard captions are masked to align the model with human preferences. 
Zhang et al.~\cite{DBLP:conf/aaai/ZhangWXYYX20} employ a chest abnormality KG with prior \textbf{chest X-ray knowledge} to support radiology report generation. In this KG, entity features are initialized with CNN-extracted features of frontal and lateral chest X-ray images, where the application of GCN mean pooling yields graph-level features that contributes to generating radiology reports.
Zhao et al.~\cite{DBLP:journals/corr/abs-2107-11970-boosting} utilizes an MMKG that associates visual objects with named entities for IC, incorporating  external \textbf{multi-modal knowledge} sourced from Wikipedia and Google Images. This MMKG, once processed through a GAT~\cite{DBLP:conf/iclr/VelickovicCCRLB18}, feeds its final layer's output into a Transformer decoder which enables entity-aware caption generation.
Nikiforova et al.~\cite{DBLP:journals/corr/abs-2210-04806-ic-add3} propose a dataset from the Geograph project\footnote{\url{http://www.geograph.org.uk/}}, including geographic coordinates of photo locations.  Concentrating on \textbf{encyclopedic knowledge}, they extract facts from DBpedia and use a retriever to prioritize facts for possible caption inclusion. These knowledge triples, combined with the image and geographic context, are then utilized in an encoder-decoder IC pipeline.

\footnotetext{Off-scene entities refer to those not part of the VG \cite{DBLP:journals/ijcv/KrishnaZGJHKCKL17} classes, as opposed to on-scene entities.}

\subsubsection{Visual Storytelling}\label{sec:kgvs}
Visual Storytelling (VST) transcends traditional Image Captioning by transforming a series of pictures into a cohesive narrative, demanding both the recognition of contexts within and across images and overcoming narrative monotony. KGs are crucial here, enhancing story diversity, rationality, and coherence.

KG-Story~\cite{DBLP:conf/aaai/HsuCHLLHK20} links concept terms from images across scenes using background KGs like FrameNet \cite{DBLP:conf/acl/BakerFL98} and Visual Genome, refined by a PLM for sequential image storytelling.
Yang et al. \cite{DBLP:conf/ijcai/YangLCLYHS19} develop a vision-aware directional encoding schema, integrating essential commonsense knowledge from ConceptNet for concept in each image. The enhanced snapshot representations, augmented with attentive knowledge, processed in a GRU-based framework for final VST.
Building upon this, MCSM~\cite{DBLP:conf/aaai/ChenHTN21} applies pruning rules and two concept selection modules
to refine commonsense knowledge facts and facilitate sentence generation for each image using a visual-adapter-equipped BART~\cite{DBLP:conf/acl/LewisLGGMLSZ20}.
Further, PR-VIST~\cite{DBLP:conf/acl/HsuCHK21} represents image sequences as story graphs to identify the best storyline path and develop a discriminator model for outputting story quality scores, aligning the narratives with human preferences.
IRW~\cite{DBLP:conf/aaai/XuYLSAX21} utilizes imaginary key concepts derived from each image for entity mention detection, retrieving candidate fact triples from ConceptNet to form a sub-KG. This sub-KG, along with the constructed scene and event graph for each image, is integrated using separate GCNs, adaptively contributing to the VST process.
KAGS~\cite{DBLP:journals/pami/LiWHC23} involves a knowledge-enriched attention network with a group-wise semantic model for globally consistent VST guidance.
\begin{discussion}
The advent of Multi-modal LLMs (MLLMs) has enriched the knowledge embedded in pre-trained models, often diminishing the need for KGs to supply coarse-grained commonsense knowledge for those IC and VST tasks. This development highlights the need for KGs offering finer-grained or specific commonsense knowledge to address model hallucination issues.
Moreover, for VST task, maintaining coherence between pictures and scenes is essential, where KGs are vital for linking disparate scenes and enriching scene transitions with background knowledge. Several methods have innovated with data-centric KG enhancements, such as deriving background KGs from story collections in training corpora~\cite{DBLP:conf/acl/HsuCHK21}, or creating event graphs through image selection from the training set that resemble the query image, subsequently using Information Extraction tools to construct events for each sentence associated with an image~\cite{DBLP:conf/aaai/XuYLSAX21}. While these strategies are pioneering, they introduce challenges in ensuring equitable model comparisons due to varied dependencies on external knowledge sources, suggesting the need for separate evaluation of such data-centric methods.
\end{discussion}

\subsubsection{Conditional Text-to-Image Generation}\label{sec:kgctg}
Conditional Text-to-Image Generation (cIG) aims to transform textual descriptions into visually realistic images, where KGs could supply detailed prior knowledge and commonsense elements not originally present in the datasets.
LeicaGAN~\cite{DBLP:conf/nips/QiaoZXT19} establishes a shared semantic space
that enables text embeddings to convey visual information, by integrating a text-image encoder for semantic, texture, and color understanding, alongside a text-mask encoder for shaping layout through segmentation masks.
During the image imagination phase, it merges the outputs of these encoders with added Gaussian noise to enhance diversity. Here, a cascaded attentive generator produces detailed and realistic images, ensuring semantic and visual coherence through adversarial learning.
Many following works~\cite{DBLP:conf/cvpr/ChengWTWT20,DBLP:journals/tcsv/ChengWTWT22,liu2023prior}  treat image-caption pairs in training sets as KB entries, enriching captions by selecting and refining relevant items from this KB, thereby aiding in feature extraction and enabling more accurate cIG.
Concretely,
KnHiGAN~\cite{DBLP:conf/iscid/GeZXZH21} and AttRiGAN~\cite{DBLP:conf/csae/ZhuGHZZZ21} present a Knowledge-enhanced Hierarchical GAN, employing a KG to enrich text descriptions for detailed generative input.  This task-specific KG is constructed from training sample attributes, formatted in RDF triples~\cite{DBLP:conf/www/GengC0PYYJC21,DBLP:journals/ws/GengCZCPLYC23}. 
For 3D cIG, T2TD~\cite{DBLP:journals/corr/abs-2305-15753} involves a text-3D KG that correlates text with 3D shapes and textual attributes, utilizing these elements as prior knowledge. During 3D generation, it retrieves the knowledge based on text descriptions and employs a causal module to select shape information relevant to the text. 

\begin{discussion}
While metrics like the Inception score \cite{DBLP:conf/nips/SalimansGZCRCC16} and R-precision \cite{DBLP:conf/cvpr/XuZHZGH018} are commonly used for evaluating the diversity of generated images and the semantic consistency between input text and generated images, current evaluation methods for generated images still lack critical assessment at the knowledge and commonsense level~\cite{DBLP:journals/corr/abs-2305-06152}. Bridging this gap presents a critical direction for future research.
\end{discussion}

\subsubsection{Scene Graph Generation}\label{sec:kgsgg}
Introduced by Johnson et al. \cite{DBLP:conf/cvpr/JohnsonKSLSBL15}, Scene Graphs (SGs) form a crucial data structure for scene understanding, cataloging object instances within a scene and delineating their interrelationships. These instances, ranging from people to places and objects, are described through attributes like shape, color, and pose \cite{DBLP:journals/pami/ChangR00C023}. The relationships between these instances, often action-based or spatial, are expressed as \textit{(subject, predicate, object)} triplets, paralleling the $(h, r, t)$ and $(e, a, v)$ triplets in KGs, often denoted as $(s, p, o)$. 
Scene Graph Generation (SGG) serves as an intermediary task, unlike other multi-modal tasks with specific end goals, providing enhanced understanding and reasoning to support downstream tasks~\cite{DBLP:journals/corr/abs-2305-06152,DBLP:conf/semweb/KonerLHDTG21,DBLP:conf/aaai/0010TYSTW021}.
 
Grasping all relationships in Scene Graph Generation (SGG) training data is challenging, yet crucial, and leveraging prior knowledge significantly aids in effectively learning relationship representations from limited data, thereby enhancing detection, recognition, and overall accuracy of SGG.
One effective approach is using \textbf{language priors}. By leveraging semantic word embeddings, these priors adjust relationship prediction probabilities, thus augmenting visual relationship identification.  For example, even with infrequent occurrences in training data, like interactions between \textit{people} and \textit{elephants}, language priors can assist the inference of similar relationships, such as ``\textit{a person riding an elephant}'', by studying more common examples like ``\textit{a person riding a horse}''~\cite{DBLP:journals/pami/ChangR00C023}. This also helps mitigate the long tail effect in visual relationships~\cite{DBLP:conf/ijcai/HeGS0L20}.
Another approach involves \textbf{statistical priors}, leveraging the structural regularity inherent in visual scenes, as highlighted in~\cite{DBLP:conf/cvpr/ZellersYTC18}. These priors capitalize on typical object-relation statistical correlations, such as ``\textit{people wearing shoes}'' or ``\textit{mountains being near water}''. 

Several works adopt the KG representation learning techniques into SGG scenario. For example,
RLSV~\cite{DBLP:conf/ijcai/WanLPZ18} uses existing SGs and images to predict new relationships between entities, targeting SG completion and blending KG embedding methods with SG characteristics in a structural-visual embedding model.
Yu et al.~\cite{DBLP:conf/icmcs/YuCLSYJLW22} improve zero-shot performance in SGG by constructing a KG from training set SG triples, distinguishing existing (non-zero-shot) and missing (zero-shot) edges. They train a KG Embedding model to complete the graph and fills these missing edges, thereby integrating zero-shot triples similarly to their non-zero-shot counterparts. 
GLAT~\cite{DBLP:conf/eccv/ZareianWYC20} separates perception and commonsense into two models, training on annotated SGs with a BERT-like masking approach (akin to KG pre-training \cite{DBLP:journals/corr/abs-1909-03193-KGBERT}) for element prediction. This method, when added to any SGG model, can rectify errors in SGs by harnessing the synergy of perception and commonsense.

Some SGG studies \cite{DBLP:conf/cvpr/ChenYCL19,DBLP:conf/cvpr/GuZL0CL19,DBLP:conf/eccv/ZareianKC20,DBLP:conf/esws/KhanBC22,DBLP:conf/wacv/ChenRL23,DBLP:conf/mm/LuCSLWH23} also employ KGs for \textbf{triple prediction}, utilizing them to generate rich and expressive SGs.
Specifically,
KERN~\cite{DBLP:conf/cvpr/ChenYCL19} leverages structured KGs to capture statistical correlations between object pairs and relationships, boosting SGG by contextualizing and stabilizing relationship predictions to address distribution imbalances.
Gu et al.~\cite{DBLP:conf/cvpr/GuZL0CL19} utilize a knowledge-based module to identify relevant ConceptNet entities and retrieve commonsense facts, each assigned a weight indicating its real-world prevalence to filter candidate triples. Then a Dynamic Memory Network \cite{DBLP:conf/icml/KumarIOIBGZPS16} is applied for multi-hop reasoning on these facts, enabling inference of the most probable SG triples.
GB-Net~\cite{DBLP:conf/eccv/ZareianKC20} views SGs as image-conditioned versions of commonsense KGs, shifting the focus from traditional entity and predicate classification to linking these two graph types. Utilizing a graph-based neural network, GB-Net iteratively propagates and refines information between and within both graphs, effectively bridging scene and commonsense knowledge.
Khan et al.~\cite{DBLP:conf/esws/KhanBC22} enrich SGs using CSKG \cite{DBLP:conf/esws/IlievskiSZ21}, a substantial commonsense KG repository. By employing graph embeddings to assess the similarity of object nodes, their approach enables graph refinement and enrichment as shown in Fig.~\ref{fig:SG_example}. This upgrades SGG with additional information on objects' spatial proximity and potential interactions derived from external knowledge, improving higher-level reasoning and mitigating some missed or incorrect predictions made during SGG.
Explicit Ontological Adjustment framework~\cite{DBLP:conf/wacv/ChenRL23} mitigates predicate biases using knowledge priors from ConceptNet and Wikidata, refining relationship detection by integrating an edge matrix from the KG into a GNN model.
Tian et al.~\cite{DBLP:conf/icmcs/TianXWYZLL23} add a branch for independent  label confidence estimation in SGG network, which assesses the difficulty of visual recognition. This branch balances the need for commonsense knowledge in diverse scenes, especially for relations like ``\textit{throwing}'' that require supplementary knowledge compared to more straightforward spatial relations like ``\textit{sitting on}''.
    
\begin{discussion}
In the realm of SGG, KGs are instrumental in mitigating relationship bias and the long-tail phenomenon within training sets, serving as a form of refinement. However, existing SGG methods still face challenges, especially in complex scenarios where the spatial distance between objects may be significant enough to disregard potential interactions. Enhancing scene graph integrity could be achieved by incorporating larger-scale images to recognize relationships between distantly located objects~\cite{DBLP:journals/csur/HossainSSL19-ICSurvey}. Moreover, extending SGG to identify human interactions, both in terms of object relations and social dynamics, would enrich scene comprehension and widen its practical uses, thereby aiding in the development of MMKG.  Additionally, leveraging structured features from SGs in training LLMs represents a promising strategy to boost multi-modal learning, capitalizing on the combined strengths of SGs, KGs, and language priors.
\end{discussion}

%% file: 4.4-ret.tex
\subsection{Retrieval Tasks}\label{sec:kgret}
\begin{defn}{\textbf{KG-aware Retrieval}}\label{def:kgret}
aims to utilize textual descriptions ($x^\mathbbm{l}$) for ranking similar visual images ($x^\mathbbm{v}$), or vice versa, including  the sorting and retrieval of all relevant images or region proposals within an image. 
Utilizing a background KG $\mathcal{G}$, this approach transcends mere appearance-based retrieval by incorporating non-visual attributes, striving for a human-level semantic understanding, especially in scenarios lacking precise targets.
\end{defn}

\subsubsection{Cross-Modal Retrieval}\label{sec:kgcmr}
Cross-Modal Retrieval (CMR) focuses on fetching data across different modalities, such as images, text, audio, or video, in response to a query from another modality. Specifically, this section explores Image-Text Retrieval, aiming to identify semantically similar instances across visual and textual modalities.

\textbf{Image-Text Matching (ITM) vs. Image-Text Retrieval (ITR):}
ITM and ITR are closely related yet differ mainly in their application:  ITM evaluates relevance between an image and text, often used in image-caption correspondence~\cite{DBLP:journals/corr/abs-2305-06152,DBLP:conf/kcap/Gomez-PerezO19}, while ITR focuses on finding relevant matches in larger datasets based on textual or visual queries, crucial for visual search engines, digital asset management, and automated content generation~\cite{DBLP:conf/ijcai/CaoLLNZ22}. 
Both ITM and ITR leverage similar underlying technologies, metrics, and datasets such as Flickr30k~\cite{DBLP:journals/tacl/YoungLHH14} and MSCOCO~\cite{DBLP:conf/eccv/LinMBHPRDZ14}, which feature extensive labeled images with captions. In cross-modal pre-training, ITM serves as a foundational task, honing the model's ability to semantically correlate images and text, thereby improving its effectiveness in ITR~\cite{DBLP:conf/cvpr/ZhangLHY0WCG21,DBLP:conf/icml/0001LXH22,DBLP:conf/icml/0008LSH23}. This pre-training ranges from coarse-grained matching (assessing general semantic relatedness) to fine-grained matching (aligning specific image regions with text). Such granularity enhances the nuanced understanding and retrieval capabilities for pre-trained models, bridging the gap between general-purpose models and the specific demands of ITR tasks.

Early CMR research often overlook long-tail and occluded semantic concepts in images \cite{DBLP:conf/ccis/YangWX21,DBLP:conf/ijcai/CaoLLNZ22}.  
Recent advancements  \cite{DBLP:conf/ijcai/ShiJLND19,DBLP:conf/eccv/WangZJPM20,DBLP:conf/eccv/WangHWXYLYJD022,li2023commonsense,DBLP:conf/mm/YangLLLLL23} tend to fix this by leveraging knowledge from frequently co-occurring concept pairs in Visual Genome's scene graphs \cite{DBLP:journals/ijcv/KrishnaZGJHKCKL17} or image captioning corpus.
They create Scene Concept Graphs (SCGs) using heuristic or rule-based tools such as language parsers \cite{DBLP:conf/eccv/AndersonFJG16}, aiming to capture fine-grained details.
Shi et al. \cite{DBLP:conf/ijcai/ShiJLND19} initially identify broad concepts, further refined into detailed ones via SCG's co-occurrence relationships, followed by a concept prediction module for accurate labeling.
EKDM~\cite{DBLP:conf/mm/YangLLLLL23} employs an iterative concept filtering module that progressively incorporates candidate concepts into a static global representation in a dynamic manner, which uses the significance scores of these concepts to set the fusion order, integrating higher-scored concepts first.
CVSE~\cite{DBLP:conf/eccv/WangZJPM20,DBLP:conf/eccv/WangHWXYLYJD022} utilizes a GNN for semantic correlation propagation in SCG, enriching concept representations with commonsense knowledge via weighted embedding summation. A confidence scaling function is introduced to mitigate long-tail distribution challenges.
CSRC~\cite{li2023commonsense} further employs a multi-head self-attention mechanism to selectively focus on deeper conceptual emphasis, while
MACK~\cite{DBLP:conf/nips/HuangWZW22} eliminates the need for paired domain data during training. 

\begin{figure}[!htbp]
  \centering
   \vspace{-1pt}
\includegraphics[width=0.95\linewidth]{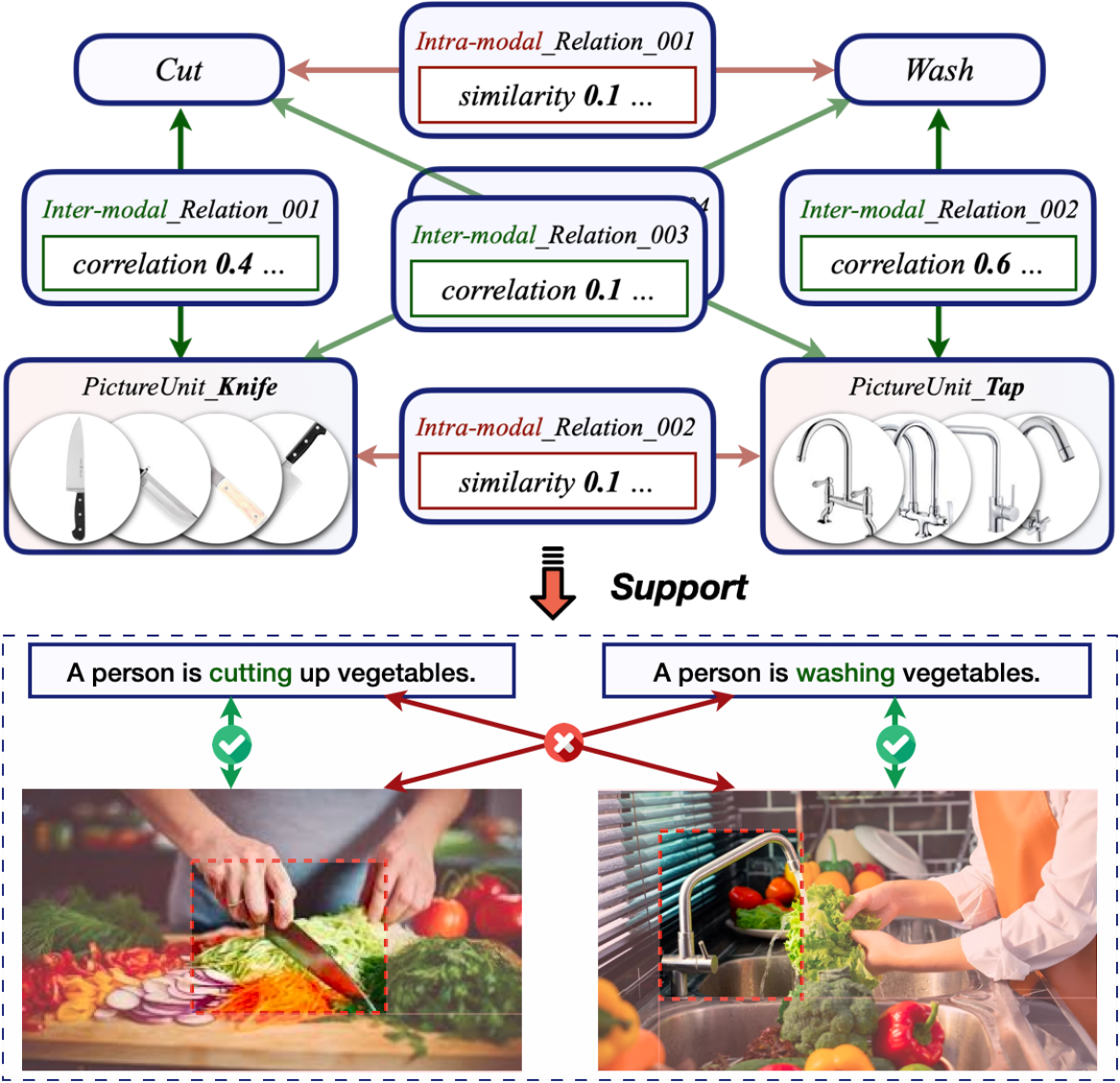}
  \caption{We illustrates the MMKG-supported Image-Text Retrieval process \cite{DBLP:journals/tomccap/FengHP23}, integrating the MMKG ontology outlined in \mbox{\S\,\ref{sec:mm-kg-cst}}. For simplicity, all URI prefixes and certain relations (\textit{sourceImg} and \textit{targetImg}) from the \textit{PictureRelation} (\textit{Inter-modal\_Relation} and \textit{Intra-modal\_Relation}) entity are omitted. This entity's values indicate intra-modal path similarities or inter-modal co-occurrence correlations, essential for training a model (e.g., multi-modal GCN) to produce knowledgeable image or text representations. Note: In cases of multiple images within a picture unit, mean pooling is used for a unified feature representation.}
  \label{fig:MMKG4CMR}
  \vspace{-9pt}
\end{figure}
Note that the background KGs in these works are typically derived from large-scale multi-modal datasets, rather than directly utilizing public KGs.
However, \textbf{a reliance on mere word co-occurrences for entity similarities can be misleading}, like wrongly linking ``\textit{man}'' and ``\textit{dog}'' due to frequent co-occurrences. Utilizing {WordNet}'s noun hierarchies helps distinguish such entities. Additionally, MMKGs could address this by capturing inter-modal co-occurrence relations, like temporal, causal, and logical connections. For instance, ``\textit{washing}'' with ``\textit{tap}'' or ``\textit{cutting}'' with ``\textit{knife}'' in image-text pairs enhances semantic understanding across modalities.
Building on this perspective, Fig.~\ref{fig:MMKG4CMR} illustrates MMKG-based approach MKVSE~\cite{DBLP:journals/tomccap/FengHP23}, which enhances image-text semantic connections, especially for images with indirect textual descriptions. 
It scores intra- and inter-modal relations in MMKGs using WordNet path similarity (calculated by NLTK~\cite{DBLP:books/daglib/0022921}) and co-occurrence correlations, improving ITR through GNN-based embeddings. 
Moreover, Yang et al. \cite{DBLP:conf/mm/YangLLLLL23} focus on a common limitation in visual concept modeling, where \textbf{varying spatial locations are often inaccurately linked by fixed relationships}, like ``\textit{man-on-bike}''  for any proximity of ``\textit{man}'' and ``\textit{bike}'' in an image.   They spatial information from a geometric graph~\cite{DBLP:conf/cvpr/MontiBMRSB17} to discern spatial relations between image regions and employ a location CNN model to refine visual-semantic representations.
EGE-CMP~\cite{DBLP:journals/pami/DongZWWWLCL23} is a entity-graph enhanced cross-modal pre-training framework that leverages entity knowledge extracted from captions instead of human labeling. It focuses on learning instance-level feature representations by infusing real semantic information into visual-text alignment, improving text-image cross-modal alignment.

\begin{discussion}
Current VLMs face challenges in fine-grained cross-modal semantic matching. Wang et al. \cite{DBLP:conf/wsdm/WangLLWZWHX23}  tackle this issue by using contrastive learning for aligning entities from Visual Genome in ITR, enhancing cross-modal sensitivity with entity masking. 
We note that a shift towards knowledge-guided strategies rather than relying solely on co-occurrence in VLM training could significantly improve retrieval and matching of fine-grained, long-tail objects, potentially leading to advanced semantic grounding \cite{DBLP:conf/aaai/ChenHCGZFPC23} and wider applications.
However, only limited studies \cite{DBLP:journals/tomccap/FengHP23} have considered the role of external knowledge like WordNet's semantic structures. 
Besides, as discussed in \mbox{\S\,\ref{sec:kbtype}}, various types of KGs, including trivia, commonsense, scientific, and situational knowledge, offer unique and complementary insights for reasoning processes. But the prevalent focus on co-occurrence information captures a fraction of commonsense knowledge. Looking ahead, exploiting long-tail knowledge from diverse large-scale KBs holds significant potential for enhancing models' generalization capabilities across various domains and real-world scenarios.
\end{discussion}

\subsubsection{Visual Referring Expressions \& Grounding}\label{sec:kgvre}
This section revisits KG-aware approaches in Visual Referring Expressions (also known as Phrase Grounding or Referring Expression Comprehension) and Visual Grounding. While CMR typically entails matching across diverse textual and visual contexts, VRE and VG focus on aligning fine-grained features within specific textual-visual pairs. From a certain point of view, these tasks are akin to adding an extra step of grounding answers in the conventional KG-based VQA, as illustrated in Fig.~\ref{fig:vqavre}.

\textbf{Visual Referring Expressions (VRE) vs. Visual Grounding (VG):}
VRE and VG~\cite{DBLP:journals/tmm/QiaoDW21} integrate linguistic and visual information, differing in focus~\cite{DBLP:journals/tmm/QiaoDW21}: 
VRE identifies and localizes a specific image region that corresponds to a given textual expression, typically involving a detailed description of one object. Conversely, VG is about localizing various object regions linked to multiple noun phrases in a sentence, aiming to establish fine-grained alignment between vision and language.
Despite these differences, both tasks require deep semantic language interpretation and manage ambiguities inherent in natural language and visual perception, relying on extensive annotated datasets. 
The line between VRE and VG often blurs in research, with some approaches~\cite{DBLP:conf/cvpr/YangLY19,DBLP:journals/pami/YangLY21,tang2023context} merging their key aspects: VRE's precise object localization and VG's broad contextual analysis.

KAC Net~\cite{DBLP:conf/cvpr/ChenGN18} utilizes the knowledge from pre-trained fixed category detectors, essential for selecting relevant proposals and ensuring visual consistency, to filter out unrelated proposals in VG progress.
Shi et al.~\cite{DBLP:conf/aaai/ShiSJZ22} tackle \textbf{zero-shot VRE}, where visual examples of queried object categories in the test set are not shown in the training set (i.e., open-vocabulary scene); they achieve this by dynamically building MMKGs using commonsense knowledge from WordNet and ConceptNet, combined with situational knowledge from Visual Genome. Query-derived entities, detected objects, and predefined relationships are integrated into these MMKGs, employing GCN for node representation and defining eight spatial relations to assist localization of noun phrases. 

The \textbf{KB-Ref} dataset~\cite{DBLP:conf/mm/WangL0020} emphasizes commonsense knowledge, with its construction process inspired by the F-VQA dataset~\cite{DBLP:journals/pami/WangWSDH18}, which involves creating a commonsense KG. Concretely, volunteers craft referring expressions for queried objects based on facts from this KG, deliberately avoiding the use of specific object names. 
Building upon the KB-Ref dataset,
ECIFA~\cite{DBLP:conf/mm/WangL0020} introduces a multi-hop facts attention module from the KG and a matching module  that utilizes expression-object scores for accurate grounding; 
CK-Transformer~\cite{DBLP:conf/eacl/ZhangYZS23}, leveraging the UNITER~\cite{DBLP:conf/eccv/ChenLYK0G0020} as its backbone, selects top-K retrieved facts from the KG for a given expression and visual region candidates, encoding these into multi-modal features to compute matching scores for each candidate.
Bu et al.~\cite{DBLP:conf/acl/BuWL00H23} observe that knowledge-based Referring Expressions often consist of two segments: visual segments (e.g., ``\textit{on the sofa}'' in Fig.~\ref{fig:vqavre}), interpretable directly from visual content like color and shape, and knowledge segments (e.g., ``\textit{used for sleeping}'' in Fig.~\ref{fig:vqavre}), requiring additional information beyond visuals like function and non-visual attributes. 
To mitigate similarity bias, they introduce the SLCO network, which uses knowledge segments for category retrieval and visual segments for object grounding.

The \textbf{SK-VG} dataset~\cite{DBLP:conf/cvpr/ChenZSWL23} targets at scene knowledge-guided VG, using movie scene images from the VCR dataset \cite{DBLP:conf/cvpr/ZellersBFC19}. Designed to promote reasoning beyond mere image content, SK-VG employs a detailed two-stage annotation process: firstly, generating story descriptions for each image, and secondly, crafting referring expressions tied to these stories and images, accompanied by object bounding box annotations. These annotations are crafted to ensure knowledge relevance to the scene context, uniqueness for accurate object identification, and diversity in both lexical use and objects. Chen et al.~\cite{DBLP:conf/cvpr/ChenZSWL23} further provide two benchmarking algorithms:  a one-stage approach that embeds knowledge into image features prior to query interaction, and a two-stage method that extracts features from images and text, subsequently employing structured linguistic data for computing region-entity similarity.

\begin{discussion}
A good VRE and VG system can beneﬁt various downstream tasks such as VQA, CMR, and IMGC.
Chen et al.\cite{DBLP:conf/aaai/ChenHCGZFPC23} develop a cross-modal semantic grounding network for ZS-IMGC, aimed at disentangling semantic attributes from images via a self-supervised method. This technique bridges knowledge from PLMs to visual models without needing region-attribute supervision. 
By leveraging AWA2-KG~\cite{DBLP:journals/ws/GengCZCPLYC23} for fine-grained labeling, it connects species to their attributes (e.g., ``\textit{zebra}'' to ``\textit{striped}'') and uses KG serialization to blend structured knowledge into cross-modal grounding. 
The network also incorporates attribute-level contrastive learning to tackle attribute imbalance and co-occurrence, thus refining the distinction of fine-grained visual features  across images from both seen and unseen classes. 
This highlights the value of KGs in Visual Grounding tasks, serving as a natural knowledge organizer and a conduit for transferring VG principles to related tasks without specialized annotations.
\end{discussion}

%% file: 4.5-plm.tex
\subsection{KG-aware Mutli-modal Pre-training}\label{sec:kgplm}
In this section, our primary focus is on pre-training definitions related to Transformer-based models, aligning with the current mainstream discourse in AI community. Other paradigms, such as Poincaré embedding pre-training~\cite{DBLP:conf/wsdm/XuRKKA20}, are not covered in this discussion.

\subsubsection{Structure Knowledge aware Pre-training}
The integration of structured knowledge into multi-modal content understanding has gradually gained momentum, drawing inspiration from advancements in the NLP field. 
KM-BART~\cite{DBLP:conf/acl/XingSMLMW20} adapts the BART~\cite{DBLP:conf/acl/LewisLGGMLSZ20} model to multi-modal tasks by incorporating a pre-trained visual feature extractor. It tackles knowledge-based commonsense generation by using COMET~\cite{bosselut2019comet} to augment image-caption datasets with commonsense context. The enriched datasets, combined with a next-token prediction target, empower KM-BART to deduce events and character intentions from image-text pairs.
ERNIE-ViL~\cite{DBLP:conf/aaai/0010TYSTW021} incorporates Scene Graph (SG) knowledge into a VLM, enhancing visual scene comprehension by adding SG completion and prediction tasks (covering objects, attributes, and relationships) during its multi-modal pre-training stage.
ROSITA~\cite{DBLP:conf/mm/CuiYWZZWY21} strengthens semantic alignments across visual and language modalities by employing a unified SG shared between the input image and text.
Existing VLMs often struggle with Image-Text Matching tasks that demand an understanding of reversed roles or actions, evident in scenarios like  ``\textit{An astronaut rides a horse}'' versus ``\textit{A horse rides an astronaut}''  (refer to \mbox{\S\,\ref{sec:kgcmr}}). To tackle this, Structure-CLIP~\cite{DBLP:journals/corr/abs-2305-06152} improves structured multi-modal representation learning by leveraging SGs to generate semantic negative examples.

\subsubsection{Knowledge Graph aware Pre-training}
Med-VLP~\cite{DBLP:conf/mm/ChenLW22} employs structured medical knowledge entities from UMLS KG~\cite{DBLP:journals/nar/Bodenreider04}  as mediators to align image and text features~\cite{DBLP:conf/nips/LiSGJXH21}, utilizing a whole-entity mask strategy~\cite{DBLP:journals/corr/abs-1904-09223} over sub-word masking.
It focuses the model's attention on crucial medical information across modalities, enabling medical VLMs to gain domain-specific knowledge for semantically aligned, knowledge-aware representations in downstream tasks.
DANCE~\cite{DBLP:conf/cvpr/YeX0XY0023} is a dataset for VLMs that converts commonsense KG triples into natural language riddles, each paired with a corresponding image. This dataset aims to bolster model learning by embedding knowledge relations between entities, linking KG entries $(h, r, t)$ with images that depict related entities,  where entities in the images are referred to as ``{\tt this item}''.
KGTransformer~\cite{DBLP:conf/www/ZhangZCGHXSC23} is pre-trained on KGs including WN18RR~\cite{DBLP:conf/iclr/SunDNT19}, FB15k-237~\cite{DBLP:conf/acl-cvsc/ToutanovaC15}, and CoDEx~\cite{DBLP:conf/emnlp/SafaviK20}, with pre-training objectives like masked relation/entity prediction and entity pair prediction.  This model can be applied to ZS-IMGC, framing the task as determining the match score between input images and target classes. 
It undergoes fine-tuning using AwA-KG~\cite{DBLP:journals/ws/GengCZCPLYC23}, with a pre-trained ResNet serving as the vision encoder and image representations further transformed through a trainable matrix.

\begin{discussion}
Current KG-equipped VLMs  primarily use triple contexts to enhance multi-modal data, with a few examples, like KGTransformer~\cite{DBLP:conf/www/ZhangZCGHXSC23}, incorporating KG's structural information into pre-training. However, its application is limited to Zero-shot Image Classification, using a uni-modal approach during pre-training.  Future research in this domain can focus on four key areas: Firstly, scaling up KG to exploit its rich knowledge and structural traits, rethinking the long-tail phenomenon in multi-modal pre-training data and expanding the knowledge scope to involve world knowledge. 
Secondly, the integration of MMKGs, which will be further discussed in \mbox{\S\,\ref{sec:mmapp}}.
Third, exploring unique pre-training paradigms suited for (MM)KGs to fully harness the value of structured knowledge in multi-modal pre-training.
Fourth, extending to more downstream tasksto align with the latest advancements in AGI, utilizing MLLMs like LLaVA~\cite{DBLP:journals/corr/abs-2304-08485}.
\end{discussion}


%% file: 5-mm-kg-task.tex
\section{Multi-modal Knowledge Graph Tasks}\label{sec:mm4kgtask}
\input{tab/mmkg-tree}
This section shifts focus to the latest trends and discussions within Multi-Modal Knowledge Graph (MMKG) research. As depicted in Fig.~\ref{fig:roadmap}, the MMKG construction process reflects human cognitive operations, including the acquisition, fusion, and inference of information. Throughout this development, a variety of tasks (i.e., {In-MMKG tasks}) have been identified, positioning the MMKG as a cornerstone for tackling a range of downstream multi-modal tasks.

\input{5.0-mmkgr}
\input{5.1-mmkge}

\input{5.2-mmkga}
\input{5.3-mmkgc}
\input{5.4-mmapp}

%% file: tab/mmkg-tree.tex
\tikzstyle{leaf}=[mybox,minimum height=1.5em,
fill=hidden-orange!60, text width=20em,  text=black,align=left,font=\scriptsize,
inner xsep=2pt,
inner ysep=4pt,
]

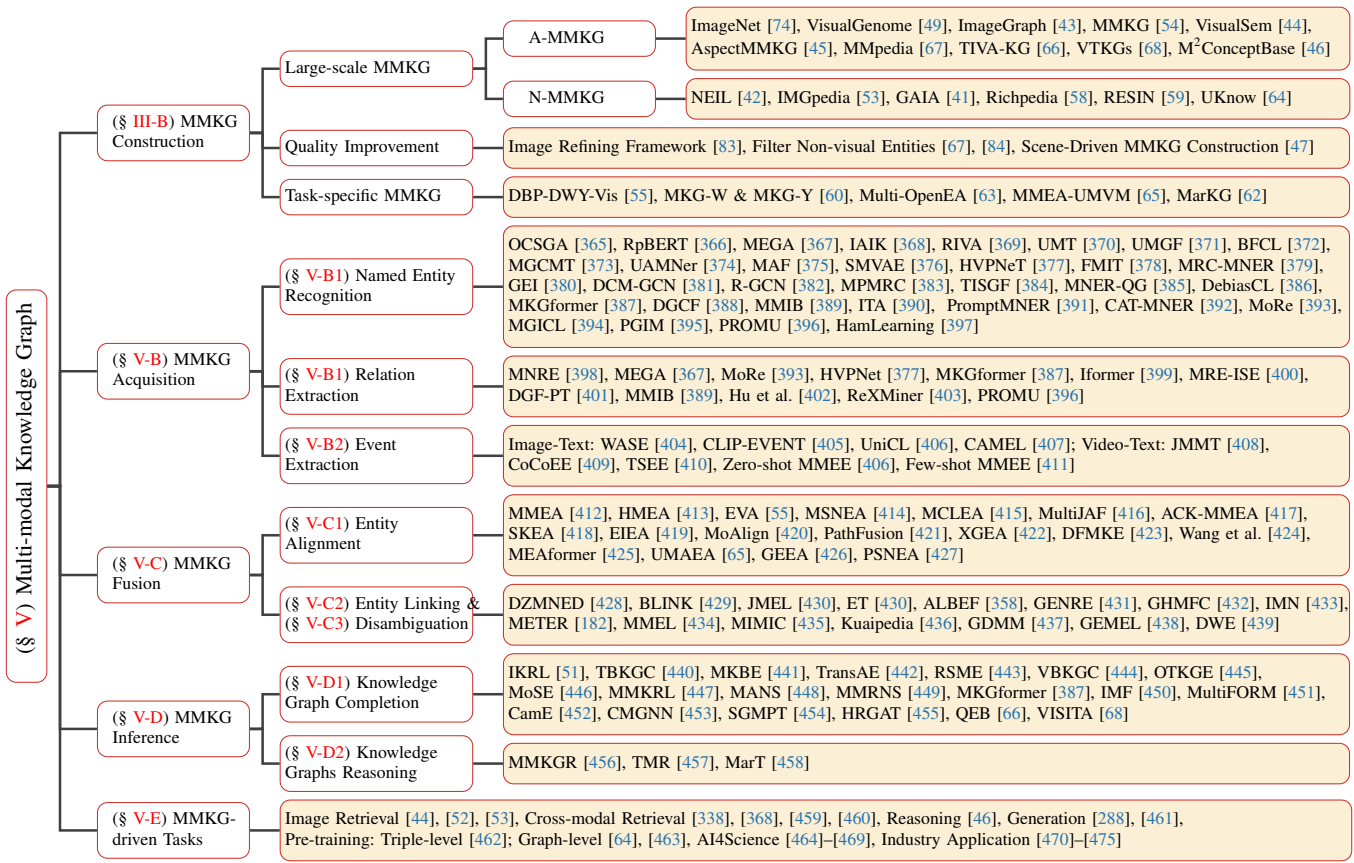
\begin{figure*}[!htbp]
  \centering
  \resizebox{\textwidth}{!}{
    \begin{forest}
      forked edges,
      for tree={
          grow=east,
          reversed=true,
          anchor=base west,
          parent anchor=east,
          child anchor=west,
          base=center,
          font=\small,
          rectangle,
          draw=hiddendraw,
          rounded corners,align=left,
          minimum width=6em,
            edge+={darkgray, line width=1pt},
            s sep=3pt,
            inner xsep=2pt,
            inner ysep=3pt,
            ver/.style={rotate=90, child anchor=north, parent anchor=south, anchor=center},
          },
        where level=1{text width=4.8em,font=\scriptsize,align=left}{},
        where level=2{text width=7.2em,font=\scriptsize,align=left}{},
        where level=3{text width=4.0em,font=\scriptsize,align=left}{},
        [
            \quad (\S~\ref{sec:mm4kgtask}) Multi-modal Knowledge Graph \quad , ver
            [
                (\S~\ref{sec:mm-kg-cst}) MMKG \\ Construction   
                [
                    Large-scale MMKG   
                    [ 
                        A-MMKG
                        [
                            {ImageNet} \cite{DBLP:conf/cvpr/DengDSLL009}{,} 
                            VisualGenome~\cite{DBLP:journals/ijcv/KrishnaZGJHKCKL17}{,} 
                            {ImageGraph} \cite{DBLP:conf/akbc/Onoro-RubioNGGL19}{,} 
                            MMKG~\cite{DBLP:conf/esws/LiuLGNOR19}{,} 
                            {VisualSem} \cite{DBLP:journals/corr/abs-2008-09150}{,} \\
                            {AspectMMKG} \cite{DBLP:conf/cikm/ZhangWWLX23}{,}
                            {MMpedia} \cite{wu2023mmpedia}{,}
                            TIVA-KG \cite{wang2023tiva}{,} 
                            VTKGs~\cite{DBLP:conf/emnlp/LeeCLJW23-VISITA}{,}
                            {M\textsuperscript{2}ConceptBase}~\cite{zha2023m2conceptbase}~
                            ,leaf,text width=25.8em
                        ]
                    ]
                    [
                        N-MMKG
                        [
                            NEIL \cite{DBLP:conf/iccv/ChenSG13}{,} 
                            {IMGpedia} \cite{DBLP:conf/semweb/FerradaBH17}{,} 
                            {GAIA} \cite{DBLP:conf/acl/LiZLPWCWJCVNF20}{,} 
                            {Richpedia} \cite{DBLP:journals/bdr/WangWQZ20}{,} 
                            RESIN~\cite{DBLP:conf/naacl/WenLLPLLZLWZYDW21}{,}
                            UKnow~\cite{DBLP:journals/corr/abs-2302-06891}~
                            ,leaf,text width=25.8em
                        ]
                    ]
                ]
                [
                     Quality Improvement
                    [ 
                        Image Refining Framework \cite{DBLP:conf/apweb/PengXTWH22}{,} 
                        Filter Non-visual Entities \cite{DBLP:conf/dasfaa/JiangLLLXWLX22,wu2023mmpedia}{,}
                         Scene-Driven MMKG Construction \cite{yaoxian2023scenedriven}~
                        ,leaf,text width=33em
                    ]
                ]
                [
                     Task-specific MMKG
                    [ 
                        DBP-DWY-Vis~\cite{DBLP:conf/aaai/0001CRC21}{,}
                        MKG-W \& MKG-Y \cite{DBLP:conf/mm/Xu0WZC22}{,} 
                        Multi-OpenEA \cite{DBLP:journals/corr/abs-2302-08774}{,} 
                        MMEA-UMVM \cite{chen2023rethinking}{,} 
                        MarKG \cite{DBLP:conf/iclr/000100LDC23}~ 
                        ,leaf,text width=33em
                    ]
                ]
            ]
            [
                (\S~\ref{sec:mmkge}) MMKG \\ Acquisition  
                [ 
                    (\S~\ref{sec:mmner})  Named Entity \\  Recognition
                [
                    OCSGA~{\cite{DBLP:conf/mm/WuZCCL020}}{,}
                    RpBERT~{\cite{DBLP:conf/aaai/0006W0SW21}}{,}
                    MEGA~{\cite{DBLP:conf/mm/ZhengFFCL021}}{,}
                    IAIK~\cite{attrDASFAA21}{,}
                    RIVA~\cite{DBLP:conf/coling/SunWSWSZC20}{,}
                    UMT~\cite{yu2020improving}{,} 
                    UMGF~{\cite{UMGF}}{,} 
                    BFCL~{\cite{DBLP:journals/ieicetd/WangCSK23}}{,} \\
                    MGCMT~{\cite{DBLP:journals/ipm/LiuWLLRZS24}}{,}
                    UAMNer~{\cite{DBLP:journals/apin/LiuWZQH22}}{,}
                    MAF~{\cite{MAF_wsdm22}}{,}
                    SMVAE~{\cite{DBLP:conf/emnlp/ZhouZSGZWY22 }}{,}
                    HVPNeT~{\cite{DBLP:conf/naacl/ChenZLYDTHSC22}}{,}
                    FMIT~{\cite{flatmner}}{,}  
                    {MRC-MNER \cite{DBLP:conf/mm/JiaSSPL00022}}{,} \\
                    GEI~{\cite{DBLP:conf/emnlp/ZhaoDSYX022}}{,}
                    DCM-GCN~{\cite{zhang2023and}}{,}
                    R-GCN~{\cite{DBLP:conf/mm/ZhaoLWXD22}}{,}
                    MPMRC~{\cite{DBLP:conf/cikm/BaoTZQ23}}{,}
                    TISGF~{\cite{10292546}}{,} 
                    MNER-QG~{\cite{DBLP:conf/aaai/JiaSSL00CL23}}{,} 
                    DebiasCL~{\cite{DBLP:conf/wsdm/ZhangYLL23}}{,} \\
                    {MKGformer~\cite{DBLP:conf/sigir/ChenZLDTXHSC22-MKGformer}}{,}
                    DGCF~{\cite{mai2023dynamic}}{,}
                    MMIB~{\cite{Bottleneck/23}}{,}
                    ITA~{\cite{DBLP:conf/naacl/WangGJJBWHT22}}{,}
                    { PromptMNER \cite{DBLP:conf/dasfaa/WangTGLYYX22}}{,} 
                    CAT-MNER~{\cite{CAT_ICME22}}{,}
                    MoRe~{\cite{DBLP:conf/emnlp/MoRe}}{,}\\
                    MGICL~{\cite{MGICL/cikm/Guo0TX23}}{,}
                    PGIM~{\cite{DBLP:conf/emnlp/PGIM}}{,}
                    PROMU~{\cite{DBLP:conf/mm/HuCLMWY23}}{,}
                    HamLearning~\cite{align/23}~
                ,leaf,text width=33em
                    ]
                ]
                [ 
                    (\S~\ref{sec:mmner}) Relation \\ Extraction
                    [ 
                        MNRE~\cite{DBLP:conf/icmcs/ZhengWFF021}{,}  
                        MEGA~\cite{DBLP:conf/mm/ZhengFFCL021}{,} 
                        MoRe~{\cite{DBLP:conf/emnlp/MoRe}}{,}
                        HVPNet~\cite{DBLP:conf/naacl/ChenZLYDTHSC22}{,} 
                        MKGformer~\cite{DBLP:conf/sigir/ChenZLDTXHSC22-MKGformer}{,}
                        Iformer~\cite{DBLP:conf/aaai/LiCQXC023}{,}
                        MRE-ISE~{\cite{Screening/acl23}}{,}\\
                        DGF-PT~\cite{DBLP:conf/acl/LiGJPC0W23}{,}  
                        MMIB~\cite{Bottleneck/23}{,} 
                        Hu et al.~\cite{MRE/acl/HuGTKY23}{,} 
                        ReXMiner~\cite{zeroshot/23}{,}
                        PROMU {\cite{DBLP:conf/mm/HuCLMWY23}}~
                        ,leaf,text width=33em
                    ]
                ]
                [ 
                    (\S~\ref{sec:mmee}) Event \\  Extraction
                    [ 
                       Image-Text: WASE \cite{DBLP:conf/acl/LiZZWLJC20}{,} 
                       CLIP-EVENT \cite{DBLP:conf/cvpr/LiXWZ0Z0JC22}{,} 
                       UniCL \cite{DBLP:conf/mm/Liu0X22}{,}
                       CAMEL  \cite{DBLP:conf/mm/DuLGSL23}{;}
                       Video-Text: 
                       JMMT \cite{DBLP:conf/emnlp/Chen0TLYCJC21}{,}\\
                       CoCoEE \cite{DBLP:conf/dasfaa/WangJZZWQ23}{,}
                       TSEE \cite{DBLP:conf/emnlp/LIEMNLP23}{,}
                       Zero-shot MMEE~\cite{DBLP:conf/mm/Liu0X22}{,}
                       Few-shot MMEE~\cite{DBLP:conf/fusion/MoghimifarSHLN23}~
                        ,leaf,text width=33em
                    ]
                ]
            ]
            [
                (\S~\ref{sec:mmkga}) MMKG \\ Fusion 
                [ 
                    (\S~\ref{sec:mmea}) Entity \\ Alignment
                    [ 
                        MMEA~\cite{DBLP:conf/ksem/ChenLWXWC20}{,} 
                        HMEA~\cite{DBLP:journals/ijon/GuoTZZL21}{,} 
                        EVA~\cite{DBLP:conf/aaai/0001CRC21}{,} 
                        MSNEA~\cite{DBLP:conf/kdd/ChenL00WYC22}{,} 
                        MCLEA~\cite{DBLP:conf/coling/LinZWSW022}{,}
                        MultiJAF~\cite{DBLP:journals/ijon/ChengZG22}{,}
                        ACK-MMEA~\cite{li2023attribute}{,}\\
                        SKEA~\cite{DBLP:journals/ipm/SuXYCJ23}{,}
                        EIEA~\cite{DBLP:conf/ksem/LiZCZ23}{,}
                        MoAlign~\cite{DBLP:journals/corr/abs-2310-06365}{,}
                        PathFusion~\cite{DBLP:journals/corr/abs-2310-05364}{,}
                        XGEA~\cite{DBLP:conf/mm/XuXS23}{,}
                        DFMKE~\cite{DBLP:journals/inffus/ZhuHM23}{,}
                        Wang et al.~\cite{DBLP:journals/dase/WangSYZLZ23}{,}\\
                        MEAformer~\cite{chen2023meaformer}{,}
                        UMAEA~\cite{chen2023rethinking}{,}
                        GEEA~\cite{DBLP:journals/corr/abs-2305-14651}{,}
                        PSNEA~\cite{DBLP:conf/mm/NiXJCCH23}~ 
                        ,leaf,text width=33em
                    ]
                ]
                [ 
                    (\S~\ref{sec:mmel}) Entity Linking \& \\(\S~\ref{sec:mmed}) Disambiguation
                    [ 
                        DZMNED~\cite{DBLP:conf/acl/CarvalhoMN18}{,} 
                        BLINK~\cite{DBLP:conf/emnlp/WuPJRZ20}{,}
                        JMEL~\cite{DBLP:conf/ecir/AdjaliBFBG20}{,} 
                        ET~\cite{DBLP:conf/ecir/AdjaliBFBG20}{,}
                        ALBEF~\cite{DBLP:conf/nips/LiSGJXH21}{,}
                        GENRE~\cite{DBLP:conf/iclr/CaoI0P21}{,} 
                        GHMFC~\cite{DBLP:conf/sigir/WangWC22}{,} 
                        IMN \cite{DBLP:conf/emnlp/ZhangH22}{,}\\
                        METER~\cite{DBLP:conf/cvpr/DouXGWWWZZYP0022}{,}
                        MMEL~\cite{DBLP:conf/uai/YangHWXH023}{,}
                        MIMIC~\cite{DBLP:conf/kdd/LuoXWZXC23}{,}
                        Kuaipedia~\cite{DBLP:journals/corr/abs-2211-00732}{,}
                        GDMM~\cite{DBLP:conf/acl/WangLZZPHMWWCXN23}{,}
                        GEMEL~\cite{DBLP:journals/corr/abs-2306-12725}{,}
                        DWE~\cite{DBLP:journals/corr/abs-2312-11816}~
                        ,leaf,text width=33em
                    ]
                ]
            ]
            [
                (\S~\ref{sec:mmkgc}) MMKG \\ Inference  
                [ 
                    (\S~\ref{sec:mmkgcc}) Knowledge \\ Graph Completion
                    [ 
                        IKRL~\cite{DBLP:conf/ijcai/XieLLS17-IKRL}{,} 
                        TBKGC~\cite{DBLP:conf/starsem/SergiehBGR18-TBKGC}{,} 
                        MKBE~\cite{DBLP:conf/emnlp/PezeshkpourC018-MKBE}{,} 
                        TransAE~\cite{DBLP:conf/ijcnn/WangLLZ19-TransAE}{,} 
                        RSME~\cite{DBLP:conf/mm/WangWYZCQ21-RSME}{,}
                        VBKGC~\cite{DBLP:journals/corr/abs-2209-07084-VBKGC}{,} 
                        OTKGE~\cite{DBLP:conf/nips/CaoXYHCH22-OTKGE}{,}\\
                        MoSE~\cite{DBLP:conf/emnlp/ZhaoCWZZZJ22-MOSE}{,}
                        MMKRL~\cite{DBLP:journals/apin/LuWJHL22-MMKRL}{,}
                        MANS~\cite{DBLP:journals/corr/abs-2304-11618-MANS}{,}
                        MMRNS~\cite{DBLP:conf/mm/Xu0WZC22-MMRNS}{,}
                        MKGformer~\cite{DBLP:conf/sigir/ChenZLDTXHSC22-MKGformer}{,}
                        IMF~\cite{DBLP:conf/www/LiZXZX23-IMF}{,}
                        MultiFORM~\cite{DBLP:conf/pkdd/ZhangLZWG22-MultiFORM}{,}\\
                        CamE~\cite{DBLP:conf/icde/XuZXXLCD23-CamE}{,}
                        CMGNN~\cite{DBLP:journals/tkde/FangZHWX23-CMGNN}{,}
                        SGMPT~\cite{DBLP:journals/corr/abs-2307-03591}{,}
                        HRGAT~\cite{DBLP:journals/tomccap/LiangZZ023-HRGAT}{,}
                        QEB~\cite{wang2023tiva}{,}
                        VISITA~\cite{DBLP:conf/emnlp/LeeCLJW23-VISITA}~
                        ,leaf,text width=33em
                    ]
                ]
                [ 
                    (\S~\ref{sec:mmkgrr}) Knowledge \\ Graphs Reasoning
                    [ 
                        MMKGR~\cite{DBLP:conf/icde/Zheng0QYCZ23-MMKGR}{,} 
                        TMR~\cite{DBLP:journals/corr/abs-2306-10345-TMR}{,} 
                        MarT~\cite{DBLP:conf/iclr/000100LDC23-MART}~
                        ,leaf,text width=33em
                    ]
                ]
            ]
            [
                (\S~\ref{sec:mmapp}) MMKG- \\driven Tasks  
                [ 
                    Image Retrieval~\cite{DBLP:journals/tip/LiuWZT17,DBLP:conf/semweb/FerradaBH17,DBLP:journals/corr/abs-2008-09150}{,} 
                    Cross-modal Retrieval~\cite{attrDASFAA21,DBLP:journals/corr/abs-2206-13163,DBLP:journals/tomccap/FengHP23,DBLP:conf/aaai/ZengJBL23}{,} 
                    Reasoning~\cite{zha2023m2conceptbase}{,} 
                    Generation~\cite{DBLP:journals/corr/abs-2107-11970-boosting,jin2023self}{,} 
                    \\
                    Pre-training: Triple-level~\cite{DBLP:conf/nips/PanYHSH22}; Graph-level~\cite{DBLP:journals/corr/abs-2302-06891,DBLP:journals/corr/abs-2309-13625}{,} AI4Science~\cite{DBLP:conf/ijcai/LinQWMZ20,DBLP:conf/iclr/ZhangBL0HDZLC22,DBLP:conf/aaai/FangZYZD0Q0FC22,fang2023knowledge,DBLP:conf/icde/XuZXXLCD23,DBLP:conf/aaai/0008LBC023}{,}
                    Industry Application~\cite{DBLP:conf/mm/ZhuZZYCZC21,DBLP:conf/cikm/XuCLSSZZZZ21,DBLP:conf/acl/Wang0LLCJHXG23,DBLP:conf/prcv/SunCLS23,DBLP:conf/cikm/SunCZWZZWZ20,DBLP:conf/mm/CaoSW0WY22}~
                    ,leaf,text width=41.9em
                ]
            ]
        ]
\end{forest}
}
\caption{Taxonomy of the Multi-modal Knowledge Graph Realm, with the "Multi-modal" prefix omitted for clarity.}
\label{fig:taxonomy_of_mmkg}
\vspace{-9pt}
\end{figure*}

%% file: 5.0-mmkgr.tex
\subsection{MMKG Representation Learning}\label{sec:mmkgr}
The current mainstream MMKG representation learning approaches  primarily concentrate on A-MMKGs, as their similarity to traditional KGs allows for more adaptable paradigm shifts. 
Those methods for integrating entity modalities within MMKGs generally fall into two categories, which are sometimes overlap within various frameworks.

\textbf{\textit{(\rmnum{1})}}  \textbf{Late Fusion} \cite{DBLP:conf/aaai/0001CRC21,DBLP:conf/coling/LinZWSW022,DBLP:journals/asc/LiZWZH22,DBLP:journals/inffus/WangYCSL22,DBLP:journals/inffus/WangYCSL22,DBLP:journals/apin/LuWJHL22-MMKRL,chen2023meaformer,chen2023rethinking} methods emphasize modality interactions and weight assignments, typically using summation, concatenation, MLPs, or gating mechanisms for feature aggregation just prior to output generation.
MKGRL-MS~\cite{DBLP:journals/inffus/WangYCSL22}  crafts unique single-modal embeddings, employing multi-head self-attention to determine each modality's contribution to semantic composition and \textbf{sum} the weighted multi-modal features for MMKG entity representation. 
MMKRL~\cite{DBLP:journals/apin/LuWJHL22-MMKRL} learns cross-modal embeddings in a unified translational semantic space, merging modality embeddings for each entity through \textbf{concatenation}.
Recent Transformer-based methods \cite{chen2023meaformer,chen2023rethinking} introduce fine-grained entity-level modality \textbf{preference} for entity representation in Multi-modal Entity Alignment.
DuMF~\cite{DBLP:journals/asc/LiZWZH22} , a dual-track approach, utilizes a bilinear layer for feature projection and an attention block for modality preference learning in each track, with a \textbf{gate network} integrating these features into a unified representation.

\textbf{\textit{(\rmnum{2})}} \textbf{Early Fusion} \cite{DBLP:journals/tkde/FangZHWX23,DBLP:journals/corr/abs-2307-03591,DBLP:journals/kais/WeiCWLZ23,DBLP:conf/sigir/ChenZLDTXHSC22-MKGformer,DBLP:journals/corr/abs-2307-03591,DBLP:conf/iclr/000100LDC23} methods integrate multi-modal feature at an initial stage, enabling deeper modality interactions suitable for complex reasoning tasks. This approach fosters a unified and potent entity representation, enhancing their compatibility in the process of integrating with other models.
CMGNN~\cite{DBLP:journals/tkde/FangZHWX23} first normalizes entity modalities into a unified embedding using a MLP, then refines them by contrasting with perturbed negative samples.
MMRotatH~\cite{DBLP:journals/kais/WeiCWLZ23} utilizes a gated encoder to merge textual and structural data, filtering irrelevant information within a rotational dynamics-based KGE framework.
Recent studies \cite{DBLP:conf/sigir/ChenZLDTXHSC22-MKGformer,DBLP:journals/corr/abs-2307-03591,DBLP:conf/iclr/000100LDC23,DBLP:conf/emnlp/LeeCLJW23-VISITA}  utilize (V)PLMs like BERT and ViT for multi-modal data integration.  They format graph structures, text, and images into sequences or dense embeddings compatible with LMs, thereby utilizing the LMs' reasoning capabilities and the knowledge embedded in their parameters to support tasks such as Multi-modal Link Prediction.

%% file: 5.1-mmkge.tex
\subsection{MMKG Acquisition}\label{sec:mmkge}
MMKG Acquisition (or Extraction) involves creating an MMKG by integrating multi-modal data such as text, images, audio, and video. This process utilizes multi-modal information from other sources, such as Internet search engines or public databases,  either to enhance an existing KG or to develop a new MMKG, thereby enabling a comprehensive understanding of complex, interconnected concepts. The resulting MMKG leverages the unique strengths of each modality to provide a more cohesive and detailed knowledge representation.

\subsubsection{Multi-modal Named Entity Recognition \& Relation Extraction}\label{sec:mmner}
\begin{figure}[!htbp]
  \centering
   \vspace{-1pt}
\includegraphics[width=0.95\linewidth]{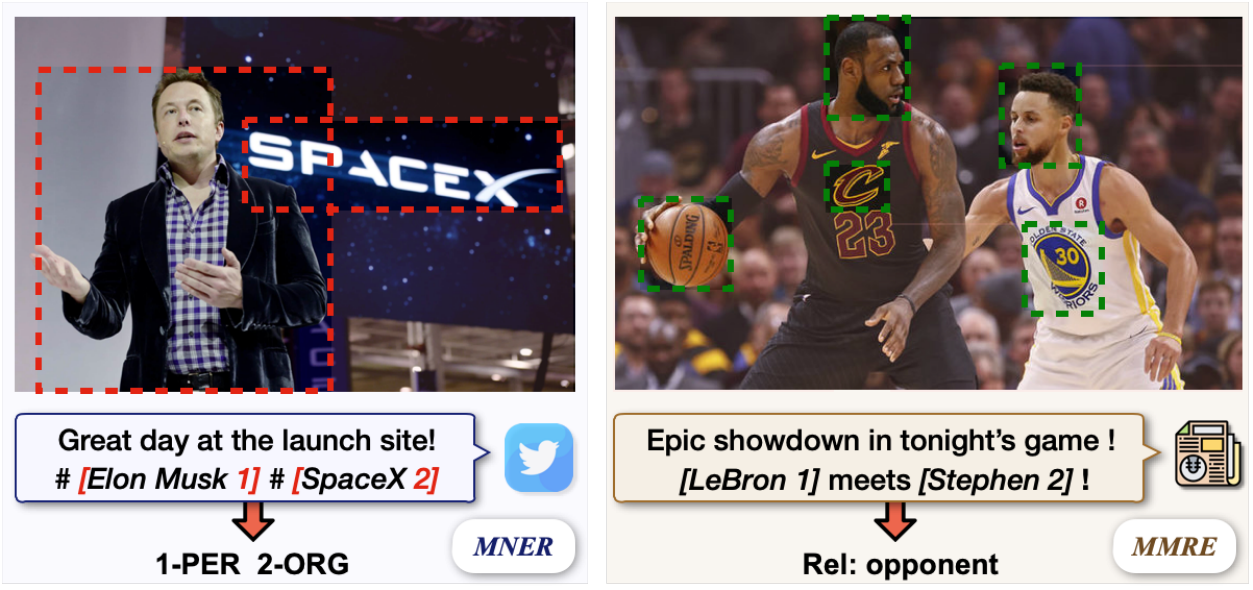}
  \caption{Illustrative examples demonstrating the application scenarios for MNER (left) and MMRE (right).}
  \label{fig:MNERandMMRE}
  \vspace{-8pt}
\end{figure}

Named Entity Recognition (NER) identifies and classifies named entities in text into categories like persons, organizations, and locations. For example, in the sentence ``\textit{Apple Inc. is founded by Steve Jobs in California}'', NER models would identify ``\textit{Apple Inc.}'' as an organization, ``\textit{Steve Jobs}'' as a person, and ``\textit{California}'' as a location. \textbf{Multi-modal Named Entity Recognition (MNER)} extends this by incorporating visual information, significantly enhancing NER in multi-modal contexts \cite{yang2022survey, chen2023survey}. 
As shown in Fig.~\ref{fig:MNERandMMRE} (left), suppose there is a social media post with a photo of Elon Musk standing in front of a SpaceX signboard, accompanied by a caption: ``\textit{Great day at the launch site!}''.
An MNER model would not only use the textual cues (``\textit{Elon Musk}'', ``\textit{SpaceX}'') but also recognize the entities in the image. This visual information reinforces the identification of ``\textit{Elon Musk}'' as a person and ``\textit{SpaceX}'' as an organization.

Relation Extraction (RE) involves detecting and classifying semantic relationships between entities within text. For example, using the same sentence, RE would discern a ``\textit{founded by}'' relationship between ``\textit{Apple Inc.}'' and ``\textit{Steve Jobs}''. \textbf{Multi-modal Relation Extraction (MMRE)} integrates visual information to enrich textual relationship analysis, proving effective in applications like news article analysis, where text is combined with related images or videos.
As shown in Fig.~\ref{fig:MNERandMMRE} (right), consider a sports article with a photo of \textit{LeBron James} and \textit{Stephen Curry} during an NBA game, with the caption: ``\textit{Epic showdown in tonight’s game!}''
In this scenario, an MMRE model analyzes the text and visual content, interpreting visual cues like their competitive stances and team logos, to infer a opponent and competitive relationship between them as opponents in the game.

\textbf{Overlap Between MNER \& MRE:}
Typically, both MNER and MMRE enhance text analysis by incorporating visual information, yet they focus on different aspects: MNER on identifying entities, and MMRE on classifying relationships between these entities. In MMKG construction frameworks, MMRE can be considered as a subsequent task to MNER. 
Despite these differences, the development methods for these tasks are increasingly converging, with many studies employing similar model designs for both MNER and MMRE~\cite{DBLP:conf/emnlp/MoRe,DBLP:conf/naacl/ChenZLYDTHSC22,DBLP:conf/mm/HuCLMWY23}. Therefore, we discuss them jointly in this section.

\begin{defn}{\textbf{Multi-modal Named Entity Recognition.}}\label{def:mmner}
MNER is typically considered as a sequence labeling problem, where a model takes a sentence  $x^\mathbbm{l}=\{w_1, w_2, \ldots, w_L\}$ along with an associated image $x^\mathbbm{v}$ as input to determine the presence and types of named entities in the text.
The goal of MNER is to predict a label sequence  $\mathcal{Y}= \{y_1,\ldots,y_n\}$, where each label $y_i$ corresponds to a named entity category for each token $w_i$ in the sentence. This process, including the probability calculation for the label sequence, follows foundational sequence labeling techniques in NER~\cite{lample2016neural}.
\end{defn}

\begin{defn}{\textbf{Multi-modal Relation Extraction.}}\label{def:mme} 
MMRE analyzes a sentence $x^\mathbbm{l}=\{w_1, w_2, \ldots, w_L\}$ alongside a corresponding image $x^\mathbbm{v}$, focusing on an entity pair  $(e_1, e_2)$ within the sentence. 
The task involves classifying the relationship between these entities, leveraging both textual and visual cues such as object interactions depicted in the image.
For each potential relation $r_i\in R$, a confidence score  $p(r_i|e_1, e_2, x^\mathbbm{l}, x^\mathbbm{v})$ is assigned. The relation set $\mathcal{R}=\{r_1, \ldots, r_C, \text{{\tt None}}\}$  includes pre-defined relation types, with ``{\tt None}'' indicating  the absence of a specific relation.
\end{defn}

\textbf{Evolution of MNER Methods:}
Advancements in MNER can be marked by diverse approaches to integrating visual and textual information. 
\textbf{\textit{(\rmnum{1})}  BiLSTM-based Methods:}
Early works~\cite{moon2018multimodal,twitter2017,DBLP:conf/mm/WuZCCL020,DBLP:conf/coling/SunWSWSZC20,attrDASFAA21} primarily employ a modality attention network to fuse text and image features, 
introducing a visual attention gate in LSTM to enhance the understanding of named entities in social media posts.

With the rising popularity of Transformer models, methods based on PLMs emerged as mainstream.
\textbf{\textit{(\rmnum{2})} PLM-based Methods:} 
Among these, encoder-based PLMs like BERT led the way in applying to MNER, with a focus on designing modality fusion methods to enhance text-only NER performance while minimizing visual noise~\cite{yu2020improving,DBLP:conf/naacl/WangGJJBWHT22,DBLP:journals/ieicetd/WangCSK23,UMGF,flatmner,MAF_wsdm22,CAT_ICME22,retrieval_mner_emnlp22}.
UMT~\cite{yu2020improving} adapts the standard BERT architecture for  MNER by adding a Transformer layer for extra contextualized text representation and a cross-modal Transformer for visual integration. 
It suggests that visual representations aid in identifying entity types, but not in detecting entity spans.
Consequently, UMT includes an auxiliary text-based module specifically for more accurate entity span detection.
FMIT~\cite{flatmner} leverages flat lattice structure and relative position encoding to enable direct interaction between fine-grained semantic units across different modalities. 
MAF~\cite{MAF_wsdm22} includes a cross-modal matching module that calculates the similarity score between text and image, using this score to adjust the amount of visual information integrated. Additionally, a cross-modal alignment module aligns the representations of both modalities, creating a unified representation that bridges the semantic gap and facilitates better text-image connections.
ITA~\cite{DBLP:conf/naacl/WangGJJBWHT22} transforms images into textual object tags and captions for cross-modal input, enabling a text-only PLM to effectively model interactions between modalities and improve robustness against image-related noise.
Wang et al.~\cite{DBLP:journals/ieicetd/WangCSK23} further propose a Transformer-based bottleneck fusion mechanism that limits noise spread by allowing modalities to interact only through trainable bottleneck tokens.
CAT-MNER~\cite{CAT_ICME22} utilizes entity label-derived saliency scores to refine attention mechanisms, addressing complexities in cross-modal exchanges.
MoRe~\cite{retrieval_mner_emnlp22} utilizes a multi-modal retrieval framework with distinct textual and image retrievers to gather relevant paragraphs and related images, respectively. This data trains separate models for NER and RE tasks, followed by a Mixture of Experts (MoE) module that synergizes their predictions.
TISGF~\cite{10292546} creates visual and textual scene graphs, encoding them to extract object-level and relationship-level features across modalities. It then employs a text-image similarity module to determine the fusion extent of visual information. Finally, multi-modal features are integrated using a fusion module, with a Conditional Random Fields (CRF) determining entity types.
PromptMNER~\cite{DBLP:conf/dasfaa/WangTGLYYX22} utilizes entity-related prompts to extract visual clues by assessing their match with an image using the CLIP~\cite{DBLP:conf/icml/RadfordKHRGASAM21} VLM.
MGICL~\cite{MGICL/cikm/Guo0TX23} analyzes data at varying granularities, including sentence and word token levels for text, and image and object levels for visuals.  Its cross-modal contrast approach enhances text analysis with visual features, supplemented by a visual gate mechanism to filter out noise.

\textbf{\textit{(\rmnum{3})} Special Cases:}
Some works highlight unique scenarios within MNER.
For example,
Liu et al.~\cite{integrating/23} propose integrating uncertainty estimation in MNER to improve prediction reliability.
DebiasCL~\cite{DBLP:conf/wsdm/ZhangYLL23} focuses on bias mitigation in MNER through a visual object density-guided hard sample mining strategy and a debiased contrastive loss.
Encoder-Decoder-based PLMs like T5~\cite{DBLP:journals/jmlr/RaffelSRLNMZLL20} and BART~\cite{DBLP:conf/acl/LewisLGGMLSZ20}, known for their strengths in NLU and NLG, have gained popularity in recent MNER studies.
Wang et al.~\cite{genarative_mm23} introduces a Fine-grained NER and Grounding (FMNERG) task, which involves extracting named entities in text, their detailed types, and corresponding visual objects in images. Here, \textit{(entity, type, object)} triples are converted into a target sequence, and T5 is used to generate this sequence, incorporating a linear transformation layer to adapt the visual object representations into T5's semantic space.

\textbf{Evolution of MMRE Methods:}
MMRE evaluates the potential relationships between entity pairs in textual content, leveraging additional multi-modal information such as images to capture complementary information for more accurate relation classification. 
Zheng et al.~\cite{DBLP:conf/icmcs/ZhengWFF021} first demonstrate the benefits of multi-modal data in filling semantic gaps and enhancing social media text analysis.
Building on this concept, some works~\cite{DBLP:conf/mm/ZhengFFCL021,Screening/acl23} introduces a textual-visual relation alignment method that aligns the sentence parsing tree and the visual scene graph, as a result improving textual relation identification.
For those PLM-based methods, {HVPNet}~\cite{DBLP:conf/naacl/ChenZLYDTHSC22} introduces object-level visual information, employing hierarchical visual features and visual prefix-guided fusion for enhanced integration;
{DGF-PT}~\cite{DBLP:conf/acl/LiGJPC0W23} implements a dual-gated fusion module, using local and global visual gates to filter unhelpful visual data, followed by a generative decoder which leverages entity types to refine candidate relations, thus capturing meaningful visual cues for MMRE.

\textbf{Resources \& Benchmarks:}
\textbf{\textit{(\rmnum{1})}} \textbf{Twitter2015}~\cite{twitter2015} and \textbf{Twitter2017}\cite{twitter2017}: Key MNER datasets featuring diverse multi-modal content from Twitter, covering 2015-2017. They include image-text pairs categorized into Location, Person, Organization, and Miscellaneous. Each record is annotated by experts for named entities.
\textbf{\textit{(\rmnum{2})}} \textbf{Twitter-FMNERG}~\cite{genarative_mm23}: Accompanying the Fine-grained NER and Grounding (FMNERG) task, this dataset provides annotations for named entities in text and their corresponding visual objects, including bounding box coordinates.
\textbf{\textit{(\rmnum{3})}}  \textbf{MNRE}~\cite{DBLP:conf/mm/ZhengFFCL021}: The main dataset for MMRE sourced from Twitter. The brevity of tweets and the varied nature of social media content make MNRE a challenging benchmark for assessing the representation, fusion, and reasoning in multi-modal techniques.
\textbf{\textit{(\rmnum{4})}}  \textbf{JMERE}~\cite{DBLP:conf/aaai/Yuan0WL23}: A joint Multi-modal Entity-Relation Extraction dataset that combines MNER and MMRE.

\input{tab/mmner-bm}

\subsubsection{Multi-modal Event Extraction}\label{sec:mmee}
Event Extraction (EE) differs from NER and RE by focusing on the dynamic and temporal aspects of events within data:
\textbf{\textit{(\rmnum{1})}} \textbf{Dynamic Nature}:  While NER and RE focus on static aspects of text (i.e., identifying entities and their relationships), EE captures the unfolding and context of events. It involves understanding not just who or what is involved, but also what is happening, when, where, and other event-related details.
\textbf{\textit{(\rmnum{2})}} \textbf{Integration of Components}:  EE integrates aspects of NER and RE, linking identified entities and their relationships to specific events, thus providing a more complete narrative.
\textbf{\textit{(\rmnum{3})}} \textbf{Contextual Richness}: EE delves into the subtleties surrounding event triggers and arguments, offering insights into how events develop and affect the involved entities.

Typically, EE focuses on identifying event \textbf{triggers} and \textbf{arguments}, capturing the dynamic aspects of events. For example, in the sentence ``\textit{The company launched a new product}'', ``\textit{launched}'' is the event trigger, with ``\textit{company}'' and ``\textit{product}'' as arguments, indicating the key participants and elements of the event. This concept contrasts with relation and entity in KGs, which primarily represent static entities and their relationships without delving into the evolving nature of events. EE's emphasis on the temporal and contextual aspects of events distinguishes it from the static, entity-focused nature of KGs, highlighting its unique role in dynamic data analysis and knowledge representation.

Early text-based EE methods leverage techniques like CNNs~\cite{DBLP:conf/acl/NguyenG15} and RNNs~\cite{DBLP:conf/naacl/NguyenCG16,DBLP:conf/emnlp/LiuCLZ19,DBLP:conf/emnlp/LiuCLBL20}, with subsequent models adopting GNNs~\cite{DBLP:conf/iccv/LiTLJUF17} to better understand event-context dependencies. The advent of PLMs further improve EE capabilities~\cite{DBLP:conf/emnlp/WaddenWLH19, DBLP:conf/acl/WangLCH0S22,DBLP:conf/acl/0001LDXLHSW22}. In Computer Vision field, EE aligns with situation recognition~\cite{DBLP:conf/eccv/PrattYWFK20,DBLP:conf/nips/KhanJT22}, focusing on identifying visual events in images or videos. 
This progression reflects a broader shift towards a more holistic understanding of events in diverse contexts, paving the way for the development of Multi-modal Event Extraction (MMEE). 

\begin{defn}{\textbf{Multi-modal Event Extraction.}}\label{def:mmee} MMEE simultaneously analyze textual data  (e.g., sentences or paragraphs)  $x^\mathbbm{l}=\{ w_{1},w_{2},...,w_{n} \}$ and visual data  (e.g., images or videos) $x^\mathbbm{v}$,  both potentially annotated with predefined event types $\mathcal{Y}_e$ and argument types $\mathcal{Y}_a$. 
In a multi-modal document  $\mathcal{D}=\{\mathcal{X}^\mathbbm{l}, \mathcal{X}^\mathbbm{v}\}$, 
an event mention $m$ is classified under an event type $y_e$ and is identified by a trigger, which can be a word $w$, an image $x^\mathbbm{v}$, or both. The task extends to extracting and classifying all event participants (i.e., arguments) within $\mathcal{D}$, assigning each to a specific argument type $y_a$. Arguments are based on textual spans or object bounding boxes in the image, with their positions explicitly identified. 
\end{defn}

\textbf{METHODS:}
Some works~\cite{DBLP:conf/acl/LiZZWLJC20,DBLP:conf/emnlp/Chen0TLYCJC21,DBLP:conf/mm/DuLGSL23} focus on region feature refinement for MMEE. Specifically,
WASE~\cite{DBLP:conf/acl/LiZZWLJC20} utilizes graphical representations of multi-modal documents for cross-modal event co-reference and image-sentence matching, targeting the challenge of limited multi-modal event annotations with a weakly supervised approach which leverages annotated uni-modal corpora and an image-caption alignment dataset.
JMMT~\cite{DBLP:conf/emnlp/Chen0TLYCJC21} employs multi-instance learning to assess region and sentence combinations, identifying key areas for multi-modal event co-reference and linking events across visual and textual modalities.
CAMEL~\cite{DBLP:conf/mm/DuLGSL23} enhances object representation in images by focusing on three specific areas within each object's bounding box and averages the encoded embeddings to aid argument extraction.
 
Recent advances emphasize refining representations via Contrastive Learning (CL)~\cite{DBLP:conf/cvpr/LiXWZ0Z0JC22,DBLP:conf/dasfaa/WangJZZWQ23,DBLP:conf/emnlp/LIEMNLP23}. Concretely, 
CLIP-EVENT~\cite{DBLP:conf/cvpr/LiXWZ0Z0JC22} contrasts images with event-aware text descriptions to training the VLMs;
CoCoEE~\cite{DBLP:conf/dasfaa/WangJZZWQ23} employs CL with weighted samples  according to event frequency;
TSEE~\cite{DBLP:conf/emnlp/LIEMNLP23} aligns optical flow with event triggers and types, observing a strong correlation between similar motion patterns and identical triggers with multi-level CL. 

Moreover, emerging research explores zero-shot~\cite{DBLP:conf/mm/Liu0X22} and few-shot~\cite{DBLP:conf/fusion/MoghimifarSHLN23} approaches to MMEE, potentially enhancing model adaptability to new or sparse data scenarios.

\input{tab/mmee-bm}

\textbf{Resources \& Benchmarks:} 
\textbf{\textit{(\rmnum{1})} M2E2}~\cite{DBLP:conf/acl/LiZZWLJC20}: Comprising multi-media news articles from the Voice of America website (2016-2017), M2E2 covers a wide range of topics like military affairs, economy, and health. 
\textbf{\textit{(\rmnum{2})} VOANews}~\cite{DBLP:conf/cvpr/LiXWZ0Z0JC22}: Constructed with image captions from various news websites, selected for their event-rich content, VOANews aims to provide a challenging benchmark for image retrieval tasks.
\textbf{\textit{(\rmnum{3})} VM2E2}~\cite{DBLP:conf/emnlp/Chen0TLYCJC21}: This first text-video dataset for MMEE is curated using YouTube searches with event types and news source names, focusing on sources like VOA, BBC, and Reuters.
\textbf{\textit{(\rmnum{4})} TVEE}~\cite{DBLP:conf/dasfaa/WangJZZWQ23}: TVEE features international news videos with captions from the On Demand News channel, aligning with the ACE2005 benchmark's partial event types.   

\textbf{METRICS:} 
Precision (P), recall (R), and F1 score are pivotal in evaluating these tasks. Precision is the ratio of correctly identified entities (or relations) to the total identified. E.g., in MNER, it reflects the proportion of accurately identified named entities from text and associated multi-modal data. Recall is the ratio of correctly identified entities (or relations) to the total relevant entities (or relations) in the dataset. E.g., in MMEE, it gauges the accuracy of extracting entities from text and multi-modal content. The F1 score, harmonizing precision, and recall, offers a comprehensive measure of both metrics. E.g., in MMRE, it provides an equilibrium, assessing the system's performance in discerning text-based entity relationships, integrating precision and recall considerations.

\begin{discussion}
Recent advancements for these tasks show a trend towards unified model designs, as evidenced by a range of studies~\cite{DBLP:conf/emnlp/MoRe,DBLP:conf/naacl/ChenZLYDTHSC22,DBLP:conf/mm/HuCLMWY23,visualRE/23,sun2024umie}. 
In certain MMEE datasets such as VM2E2~\cite{DBLP:conf/emnlp/Chen0TLYCJC21}, 
the visual modality lacks direct event and argument annotations, positioning visual features as supportive elements in benchmarking. However, 
the prevalent multi-modal F1 score, focusing mainly on text-based event type classification, overlooks the contribution evaluation of visual elements. This scenario highlights the need for future research to devise more balanced multi-modal evaluation metrics that thoroughly integrate visual and textual components.
Looking forward, the emergence of MLLMs and their zero-shot extraction capabilities~\cite{DBLP:conf/nips/Wei0SBIXCLZ22,li2023prompting} heralds a pivot towards generative-based approaches. This shift implies a broader horizon for MNER, MMRE, and MMEE, urging the expansion into more intricate, specialized, and inherently comprehensive multi-modal extraction tasks.
\end{discussion}

%% file: tab/mmner-bm.tex
\begin{table}[ht]
 \centering
 \renewcommand\arraystretch{1.0}
 \caption{Comparison of MNER performance on the Twitter-2015~\cite{twitter2015} and Twitter-2017~\cite{twitter2017} datasets, evaluated using precision (P), recall (R), and F1 score as metrics. Results for CLIP~\cite{DBLP:conf/icml/RadfordKHRGASAM21} and BLIP~\cite{DBLP:conf/icml/0001LXH22} are sourced from Hu et al.~\cite{DBLP:conf/mm/HuCLMWY23}.}
 \label{tab:Me}
 \resizebox{0.92\linewidth}{!}{
 \begin{NiceTabular}{lcccccc}
 \CodeBefore
 \rowcolors{2}{gray!10}{white}
 \columncolor{gray!1}{1}
 \rowcolor{gray!30}{1}
 \rowcolor{gray!30}{2}
 \Body
 \toprule[0.8pt]
 \multirow{3}*{\makebox[2cm][c]{\textbf{Models}}} & \multicolumn{3}{c}{\footnotesize \textbf{Twitter-2015}} & \multicolumn{3}{c}{\footnotesize \textbf{Twitter-2017}} \\
 \cmidrule(r){2-4} \cmidrule(r){5-7} 
 & {\scriptsize \textbf{P}} & {\scriptsize \textbf{R}} & {\scriptsize \textbf{F1}} & {\scriptsize \textbf{P}} & {\scriptsize \textbf{R}} & {\scriptsize \textbf{F1}} \\
 \midrule[0.8pt]
Zhang et al~\cite{twitter2015} & 72.75 & 68.74 & 70.69 & - & - & - \\
OCSGA {\footnotesize \cite{DBLP:conf/mm/WuZCCL020}} & 74.71 & 71.21 & 72.92 & - & - & - \\
Lu et al.~\cite{twitter2017} & - & - & - & 81.62 & 79.90 & 80.75 \\
RpBERT {\footnotesize \cite{DBLP:conf/aaai/0006W0SW21}} & 71.15 & 74.30 & 72.69 & 82.85 & 84.38 & 83.61 \\
MEGA {\footnotesize \cite{DBLP:conf/mm/ZhengFFCL021}} & 70.35 & 74.58 & 72.35 & 84.03 & 84.75 & 84.39 \\
{\scriptsize VisualBERT \cite{DBLP:journals/corr/abs-1908-03557}} & 68.84 & 71.39 & 70.09 & 84.06 & 85.39 & 84.72 \\
IAIK \cite{attrDASFAA21} & 74.78 & 71.82 & 73.27 & - & - & - \\ 
RIVA \cite{DBLP:conf/coling/SunWSWSZC20} & 75.02 & 71.94 & 73.45 & - & - & - \\ 
UMT \cite{yu2020improving} & 71.67 & 75.23 & 73.41 & 85.28 & 85.34 & 85.31 \\ 
CLIP \cite{DBLP:conf/icml/RadfordKHRGASAM21} & 74.25 & 74.64 & 74.44 & 85.34 & 85.29 & 85.31 \\ 
UMGF {\footnotesize \cite{UMGF}} & 74.49 & 75.21 & 74.85 & 86.54 & 84.50 & 85.51 \\
BFCL {\footnotesize \cite{DBLP:journals/ieicetd/WangCSK23}} & 74.02 & 75.07 & 74.54 & 85.99 & 85.42 & 85.70 \\
MGCMT {\footnotesize \cite{DBLP:journals/ipm/LiuWLLRZS24}} & 73.57 & 75.59 & 74.57 & 86.03 & 86.16 & 86.09 \\
UAMNer {\footnotesize \cite{DBLP:journals/apin/LiuWZQH22}} & 73.02 & 74.75 & 73.87 & 86.17 & 86.23 & 86.20 \\
MAF {\footnotesize \cite{MAF_wsdm22}} & 71.86 & 75.10 & 73.42 & 86.13 & 86.38 & 86.25 \\
SMVAE {\footnotesize \cite{DBLP:conf/emnlp/ZhouZSGZWY22 }} & 74.40 & 75.76 & 75.07 & 85.77 & 86.97 & 86.37 \\
GEI {\footnotesize \cite{DBLP:conf/emnlp/ZhaoDSYX022}} & 73.39 & 75.51 & 74.43 & 87.50 & 86.01 & 86.75 \\
FMIT {\footnotesize \cite{flatmner}} & 75.11 & 77.43 & 76.25 & 87.57 & 86.26 & 86.79 \\
DebiasCL {\footnotesize \cite{DBLP:conf/wsdm/ZhangYLL23}} & 74.45 & 76.13 & 75.28 & 87.59 & 86.11 & 86.84 \\
{\scriptsize MRC-MNER \cite{DBLP:conf/mm/JiaSSPL00022}} & 78.10 & 71.45 & 74.63 & 88.78 & 85.00 & 86.85 \\
HVPNeT {\footnotesize \cite{DBLP:conf/naacl/ChenZLYDTHSC22}} & 73.87 & 76.82 & 75.32 & 85.84 & 87.93 & 86.87 \\
DCM-GCN {\footnotesize \cite{zhang2023and}} & 73.41 & 75.88 & 74.63 & 86.09 & 87.93 & 87.00 \\
R-GCN {\footnotesize \cite{DBLP:conf/mm/ZhaoLWXD22}} & 73.95 & 76.18 & 75.00 & 86.72 & 87.53 & 87.11 \\
MPMRC {\footnotesize \cite{DBLP:conf/cikm/BaoTZQ23}} & 77.15 & 75.39 & 76.26 & 87.10 & 87.16 & 87.13 \\
TISGF {\footnotesize \cite{10292546}} & 71.15 & 75.35 & 73.19 & 86.48 & 87.78 & 87.18 \\
MNER-QG {\footnotesize \cite{DBLP:conf/aaai/JiaSSL00CL23}} & 77.76 & 72.31 & 74.94 & 88.57 & 85.96 & 87.25 \\
{\scriptsize MKGformer \cite{DBLP:conf/sigir/ChenZLDTXHSC22-MKGformer}} & - & - & - & 86.98 & 88.01 & 87.49 \\
DGCF {\footnotesize \cite{mai2023dynamic}} & 74.76 & 75.50 & 75.13 & 88.50 & 87.65 & 88.07 \\
MMIB {\footnotesize \cite{Bottleneck/23}} & 74.44 & 77.68 & 76.02 & 87.34 & 87.86 & 87.60 \\
ITA {\footnotesize \cite{DBLP:conf/naacl/WangGJJBWHT22}} & 78.93 & 78.14 & 78.53 & 88.52 & 90.16 & 89.33 \\
BLIP {\footnotesize \cite{DBLP:conf/icml/0001LXH22}} & 77.73 & 76.58 & 77.15 & 88.92 & 88.67 & 88.79 \\
{\scriptsize PromptMNER \cite{DBLP:conf/dasfaa/WangTGLYYX22}} & 78.03 & 79.17 & 78.60 & 89.93 & 90.60 & 90.26 \\
CAT-MNER {\footnotesize \cite{CAT_ICME22}} & 78.75 & 78.69 & 78.72 & 90.27 & 90.67 & 90.47 \\
MoRe {\footnotesize \cite{DBLP:conf/emnlp/MoRe}} & 79.33 & 79.11 & 79.22 & 90.74 & 90.53 & 90.63 \\
MGICL {\footnotesize \cite{MGICL/cikm/Guo0TX23}} & 80.31 & 80.06 & 80.18 & 91.07 & 90.61 & 90.94 \\
PGIM {\footnotesize \cite{DBLP:conf/emnlp/PGIM}} & 79.21 & 79.45 & 79.33 & 90.86 & 92.01 & 91.43 \\
PROMU {\footnotesize \cite{DBLP:conf/mm/HuCLMWY23}} & 80.03 & 80.97 & 80.50 & 91.97 & 91.33 & 91.65\\
\bottomrule[0.8pt]
\end{NiceTabular}
\vspace{-4pt}
}
\end{table}

\begin{table}[ht]
\centering
\renewcommand\arraystretch{1.0}
\caption{Comparison of MMRE performance on MNRE~\cite{DBLP:conf/mm/ZhengFFCL021}.}
\label{tab:MMRE}
\resizebox{0.6\linewidth}{!}{
\begin{NiceTabular}{lccc}
\CodeBefore
\rowcolors{2}{gray!10}{white}
\columncolor{gray!1}{1}
\rowcolor{gray!30}{1}
\Body
\toprule[0.8pt]
\makebox[2cm][c]{\textbf{Models}} & { \textbf{P}} & { \textbf{R}} & {\textbf{F1}} \\
\midrule[0.8pt]
MEGA {\footnotesize \cite{DBLP:conf/mm/ZhengFFCL021}} & 64.51 & 68.44 & 66.41 \\ 
MoRe {\footnotesize \cite{DBLP:conf/emnlp/MoRe}} & 66.66 & 70.58 & 68.56 \\
HVPNet {\footnotesize \cite{DBLP:conf/naacl/ChenZLYDTHSC22}} & 83.64 & 80.78 & 81.85 \\
MKGformer {\footnotesize \cite{DBLP:conf/sigir/ChenZLDTXHSC22-MKGformer}} & 82.67 & 81.25 & 81.95 \\
Wu et al. {\footnotesize \cite{Screening/acl23}} & 84.69 & 83.38 & 84.03 \\
DGF-PT {\footnotesize \cite{DBLP:conf/acl/LiGJPC0W23}} & 84.35 & 83.83 & 84.47 \\
Hu et al. {\footnotesize \cite{MRE/acl/HuGTKY23}} & 85.03 & 84.25 & 84.64 \\
PROMU {\footnotesize \cite{DBLP:conf/mm/HuCLMWY23}} & 84.95 & 85.76 & 84.86 \\
\bottomrule[0.8pt]
\end{NiceTabular}
\vspace{-8pt}
}
\end{table}

%% file: tab/mmee-bm.tex

\begin{table}[!htbp]
\renewcommand{\arraystretch}{0.95}
\renewcommand{\ttdefault}{pcr}
\caption{Comparative analysis of MMEE results across diverse datasets. M2E2~\cite{DBLP:conf/acl/LiZZWLJC20} utilizes image and text inputs. Both  TVEE~\cite{DBLP:conf/emnlp/Chen0TLYCJC21} and VM2E2~\cite{DBLP:conf/dasfaa/WangJZZWQ23} employ video and text inputs. 
}
\centering
\resizebox{0.98\linewidth}{!}{
\begin{NiceTabular}{llcccccc }
\CodeBefore
\rowcolors{2}{gray!10}{white}
\columncolor{gray!1}{1}
\rowcolor{gray!30}{1}
\rowcolor{gray!30}{2}
\Body
\toprule[0.8pt]
\multirow{3}{*}{\makebox[0.9cm][c]{\textbf{Dataset}}} & \multirow{3}{*}{\makebox[2.3cm][c]{\textbf{Models}}} & \multicolumn{3}{c}{\textbf{Trigger}} & \multicolumn{3}{c}{\textbf{Argument}} \\
\cmidrule(r){3-5} \cmidrule(r){6-8}
 & &\textbf{P} &\textbf{R}&\textbf{F1}&\textbf{P} &\textbf{R}&\textbf{F1}\\
\midrule[0.8pt]
\multirow{5}{*}{{M2E2}}  & Flat \cite{DBLP:conf/acl/LiZZWLJC20}          & 33.9 & 59.8 &   42.2   &   12.9   &17.6 & 14.9 \\ 
& WASE {\footnotesize \cite{DBLP:conf/acl/LiZZWLJC20}} & 38.2 & 67.1 & 49.1 & 18.6 & 21.6 & 19.9 \\
& {\footnotesize CLIP-EVENT \cite{DBLP:conf/cvpr/LiXWZ0Z0JC22}}  & 41.3 & 72.8 & 52.7 & 21.1 & 13.1 & 17.1 \\
& UniCL {\footnotesize \cite{DBLP:conf/mm/Liu0X22}}  & 44.1 & 67.7 & 53.4 & 24.3 & 22.6 & 23.4 \\
& CAMEL {\footnotesize \cite{DBLP:conf/mm/DuLGSL23}}  & 55.6 & 59.5 & 57.5 & 31.4 & 35.1 & 33.2 \\
\midrule
\multirow{3}{*}{{TVEE}}   & JMMT {\footnotesize \cite{DBLP:conf/emnlp/Chen0TLYCJC21}}          & 74.3 & 80.2 &   77.1   &   50.1   & 54.9 & 52.3 \\ 
& CoCoEE {\footnotesize \cite{DBLP:conf/dasfaa/WangJZZWQ23}} & 80.7 & 76.4 & 78.5 & 65.6 & 45.4 & 53.6 \\
& TSEE {\footnotesize \cite{DBLP:conf/emnlp/LIEMNLP23}}  & 82.6 & 80.5 & 81.5 & 67.0 & 49.3 & 56.8 \\

\midrule
\multirow{3}{*}{{VM2E2}} & JMMT \cite{DBLP:conf/emnlp/Chen0TLYCJC21}          & 39.7 & 56.3 &   46.6   &   17.9   & 24.3 & 20.6 \\ 
& CoCoEE {\footnotesize \cite{DBLP:conf/dasfaa/WangJZZWQ23}} & 47.3 & 47.7 & 47.5 & 26.7 & 18.5 & 21.8 \\
& TSEE {\footnotesize \cite{DBLP:conf/emnlp/LIEMNLP23}}  & 49.2 & 53.5 & 51.6 & 24.5 & 27.4 & 25.9 \\

\bottomrule[0.8pt]
\end{NiceTabular}}
\label{tab:re}
\end{table}

%% file: 5.2-mmkga.tex
\subsection{MMKG Fusion}\label{sec:mmkga}

The proliferation of heterogeneous data across the Internet has led to the creation of numerous independent MMKGs. Integrating these MMKGs from diverse data sources is essential, making MMKG fusion a critical stage in MMKG construction~\cite{DBLP:conf/ijcnn/WangLLZ19}. This process involves various tasks, including Multi-Modal Entity Alignment (MMEA), Entity Linking (MMEL), and Entity Disambiguation (MMED).

\subsubsection{Multi-modal Entity Alignment}\label{sec:mmea}
Entity Alignment (EA) is pivotal for KG integration, which aims to match identical entities across different KGs using the relational, attributive, and literal (surface) features of entities. Specifically,  symbolic logic approaches~\cite{DBLP:conf/acl/XiangZCCLZ21,DBLP:conf/ijcai/QiZCCXZZ21} apply manually defined rules, such as logical inference and lexical matching, to guide the alignment.  Embedding-based methods~\cite{DBLP:conf/icml/GuoZSCHC22,DBLP:conf/acl/GuoHZC22,DBLP:journals/tkde/SunHWWQ23, DBLP:conf/kdd/GaoLW0W022,DBLP:conf/www/LiuW00G23,DBLP:conf/wsdm/XinSH0022,DBLP:conf/ijcai/CaiMZJ22,DBLP:conf/icml/SunHXCRH23} utilize learned entity embeddings to expedite the alignment, avoiding the need for pre-defined heuristics. Multi-Modal Entity Alignment (MMEA) incorporates visual data from the MMKGs, associating each entity with images to enhance the EA~\cite{DBLP:conf/esws/LiuLGNOR19}.

\begin{defn}{\textbf{Multi-modal Entity Alignment.}}\label{def:mmea}
A MMKG is denoted as $\mathcal{G}=\{\mathcal{E}, \mathcal{R}, \mathcal{A}, \mathcal{T}, \mathcal{V}\}$ with $\mathcal{T} = \{\mathcal{T_A}, \mathcal{T_R}\}$. 
Given two aligned \textbf{A-MMKGs} $\mathcal{G}_1$ $=$ $\{\mathcal{E}_1, \mathcal{R}_1, \mathcal{A}_1, \mathcal{V}_1, \mathcal{T}_1\}$ and $\mathcal{G}_2$ $=$ $\{\mathcal{E}_2, \mathcal{R}_2, \mathcal{A}_2, \mathcal{V}_2, \mathcal{T}_2\}$, the goal of MMEA is to identify pairs of entities ($e^1_i$, $e^2_i$) from $\mathcal{E}_1$ and $\mathcal{E}_2$ respectively, that represent the same real-world entity $e_i$. A set of pre-aligned entity pairs serves as a reference, divided into a training set (seed alignments $\mathcal{S}$) and a test set $\mathcal{S}_{te}$, proportioned by a pre-defined seed alignment ratio $R_{sa}$. 
The available modalities associated with an entity are denoted by $\mathcal{M}=\{\mathbbm{g}, \mathbbm{r}, \mathbbm{a}, \mathbbm{v}, \mathbbm{s}\}$, which represent the graph structure, relation,  attribute, vision, and surface (i.e., entity name) modalities, respectively.
\end{defn}

While both relation, attribute, and surface modalities can be categorized under language modalities, they are frequently distinguished as separate modalities in MMEA communities~\cite{DBLP:conf/aaai/0001CRC21,DBLP:conf/coling/LinZWSW022,DBLP:journals/ijon/ChengZG22,chen2023meaformer,chen2023rethinking,DBLP:journals/corr/abs-2305-14651,DBLP:journals/ipm/SuXYCJ23,DBLP:journals/inffus/ZhuHM23}.  Besides, research shows a variety of modal usage patterns: some studies focus solely on the \textbf{types} of attributes and relations during the alignment process \cite{chen2023meaformer,chen2023rethinking}, while others incorporate their \textbf{textual content} into entity representations via using PLM (e.g., BERT~\cite{DBLP:conf/naacl/DevlinCLT19}) \cite{wu2022leveraging,DBLP:journals/inffus/ZhuWHCXLBD23,DBLP:journals/corr/abs-2310-05364,DBLP:journals/corr/abs-2310-06365, EASY, LargeEA} or word embeddings (e.g., Glove~\cite{DBLP:conf/emnlp/PenningtonSM14}) \cite{DBLP:conf/aaai/0001CRC21,DBLP:conf/coling/LinZWSW022,chen2023meaformer,chen2023rethinking,DBLP:conf/kdd/ChenL00WYC22}.
Additionally, some methods are proposed for entities that have only one image \cite{DBLP:conf/aaai/0001CRC21,DBLP:conf/coling/LinZWSW022}, while others are prepared to handle cases where the number of images per entity can be multiple~\cite{DBLP:journals/corr/abs-2302-08774} or even missing~\cite{chen2023rethinking}.

\textbf{Progress:}
Current MMEA research can be broadly divided into two streams according to their underlying motivation:

\textbf{\textit{(\rmnum{1})} Exploring better cross-KG modality feature fusion.}
{MMEA} \cite{DBLP:conf/ksem/ChenLWXWC20} is first introduced in 2020 
as a method that merges knowledge representations from multiple modalities and aligns entities by minimizing the distance between their holistic embeddings; 
{HMEA} \cite{DBLP:journals/ijon/GuoTZZL21} expands MMKG representation from the Euclidean space to the hyperbolic manifold, offering a more refined geometric interpretation. 
{EVA} \cite{DBLP:conf/aaai/0001CRC21} assigns different importance to each modality via an attention mechanism. It further introduces an unsupervised MMEA approach that leverages visual similarities between entities to create a pseudo seed dictionary, thus reducing dependence on gold-standard labels.
{MSNEA} \cite{DBLP:conf/kdd/ChenL00WYC22} leverages visual cues to guide relational feature learning and weights valuable attributes for alignment. 
{MCLEA} \cite{DBLP:conf/coling/LinZWSW022} applies KL divergence to bridge the modality distribution gap between joint and uni-modal embedding. 
{ACK-MMEA} \cite{li2023attribute} presents an attribute-consistent KG representation learning method to solve the contextual gap caused by different attributes. 
{PathFusion}~\cite{DBLP:journals/corr/abs-2310-05364} combines information from different modalities using the modality similarity path as an information carrier.
{DFMKE}~\cite{DBLP:journals/inffus/ZhuHM23} employs a late fusion approach with modality-specific low-rank factors that enhance feature integration across various knowledge spaces, complementing early fusion output vectors.
Considering that the surrounding modality of each entity is inconsistent,
{MEAformer}~\cite{chen2023meaformer} dynamically adjusts the mutual modality preference for entity-level modality fusion. Recent works like {MoAlign}~\cite{DBLP:journals/corr/abs-2310-06365}, UMAEA~\cite{chen2023rethinking} and DESAlign~\cite{wang2024towards} follow similar settings.
{XGEA}~\cite{DBLP:conf/mm/XuXS23} leverages the information from one modality as complementary relation information to enrich entity embeddings by computing inter-modal attention within the GAT layers.

\textbf{\textit{(\rmnum{2})} Analyzing the practical limitations and challenges in MMKG alignment.}
Wang et al. \cite{DBLP:journals/dase/WangSYZLZ23} tackled the issue of image-type mismatches in aligned multi-modal entities by filtering out incongruent images using pre-defined ontologies and an image type classifier. 
The inherent incompleteness of visual data in MMKGs poses another challenge, where many entities lack images (e.g., $67.58\%$ in DBP15K$_{JA\text{-}EN}$ \cite{DBLP:conf/aaai/0001CRC21}). Furthermore, the intrinsic ambiguity of visual images also impacts the alignment quality (i.e., each entity has multiple visual aspects as elaborated in \mbox{\S~\ref{sec:mm-kg-cst}}). 
Chen et al. \cite{chen2023rethinking} introduces the MMEA-UMVM dataset to study the impact of training noise and performance degradation at high rates of missing modalities. They further propose {UMAEA}, which employs a multi-scale modality hybrid approach with a circularly missing modality imagination module equipped.
Considering that many entities in the source KG may not have aligned entities in
the target KG (i.e., the dangling entities \cite{DBLP:conf/acl/SunCH20,DBLP:conf/acl/Luo022}), Guo et al. \cite{DBLP:journals/corr/abs-2305-14651} introduce the entity synthesis task to generate new entities either conditionally or unconditionally, and propose the {GEEA} framework, which employs a mutual variational autoencoder (M-VAE) for entity synthesis.
To overcome the costly and time-intensive process of acquiring initial seeds, Ni et al. developed the Pseudo-Siamese Network ({PSNEA}) \cite{DBLP:conf/mm/NiXJCCH23}, complemented by an Incremental Alignment Pool that labels probable alignments, reducing reliance on data swapping and sample re-weighting.

\input{tab/mmea-bm}
\textbf{Resources \& Benchmarks:}
\textbf{\textit{(\rmnum{1})}} The first MMEA dataset includes FB15K-DB15K (\textbf{FBDB15K}) and FB15K-YAGO15K (\textbf{FBYG15K}) ~\cite{DBLP:conf/esws/LiuLGNOR19} with three data splits: $R_{sa} \in \{0.2, 0.5, 0.8\}$.
\textbf{\textit{(\rmnum{2})}} \textbf{Multi-modal DBP15K}~\cite{DBLP:conf/aaai/0001CRC21}: An extension of the DBP15K~\cite{DBLP:conf/semweb/SunHL17} which attaches entity-matched images from DBpedia~\cite{DBLP:conf/semweb/AuerBKLCI07} and Wikipedia~\cite{denoyer2006wikipedia} to the original cross-lingual EA benchmark. It includes four language-specific KGs from DBpedia, with three bilingual settings ($R_{sa}=0.3$), namely DBP15K$_{ZH\text{-}EN}$, DBP15K$_{JA\text{-}EN}$, and DBP15K$_{FR\text{-}EN}$. 
Each setting contains approximately $400$K triples and $15$K pre-aligned entity pairs. 
We benchmark those recent MMEA methods using this series of datasets as outlined in Table \ref{tab:mmea-bm}.
\textbf{\textit{(\rmnum{3})}} \textbf{Multi-OpenEA}~\cite{DBLP:journals/corr/abs-2302-08774}: A multi-modal expansion of the OpenEA benchmarks~\cite{DBLP:journals/pvldb/SunZHWCAL20} which links entities with their top-3 related images sourced through Google search.
\textbf{\textit{(\rmnum{4})}} \textbf{MMEA-UMVM}\cite{chen2023rethinking}: It contains two bilingual datasets (EN-FR-15K, EN-DE-15K) and two monolingual datasets (D-W-15K-V1, D-W-15K-V2) derived from Multi-OpenEA datasets ($R_{sa}=0.2$)~\cite{DBLP:journals/corr/abs-2302-08774} and all three bilingual datasets from DBP15K~\cite{DBLP:conf/aaai/0001CRC21}. It introduces variability in visual information by randomly removing images, resulting in 97 distinct dataset splits.

\begin{discussion}
Adopting strategies beyond model architecture is recognized for boosting performance. 
Iterative training~\cite{DBLP:conf/coling/LinZWSW022,DBLP:conf/aaai/0001CRC21}, for example, 
incrementally refines model performance by identifying and adding cross-KG entity pairs as mutual nearest neighbors in the embedding space every $K_e$ epochs (e.g., 5), with pairs confirmed for inclusion in the training set after remaining mutual nearest neighbors across $K_s$ successive iterations (e.g., 10).
Similarly, the STEA framework \cite{DBLP:conf/wsdm/0025LHZ23} can be utilized to generate additional pseudo-aligned pairs, thereby expanding the training data. 
Additionally, the CMMI module  \cite{chen2023rethinking}  can be integrated into models to create synthetic visual embeddings, mitigating the impact of  missing images.
For fair evaluation, models employing these strategies should be assessed separately from those that do not.
Moreover, considerations like the use of entity names (surface forms), computational complexity, textual encoding methods, and the integration of additional data warrant careful attention in comparing methodologies in future research.
\end{discussion}


\subsubsection{Multi-modal Entity Linking}\label{sec:mmel}
Entity Linking (EL) serves as a crucial component in various applications~\cite{shen2014entity,shen2021entity,sevgili2022neural}, including Question Answering, Relation Extraction, and Semantic Search.
The main target of EL is to associate textual mentions within documents with their respective entities in a KG (e.g., Freebase~\cite{DBLP:conf/sigmod/BollackerEPST08}). 
Notably, mentions extend beyond textual forms, including images, audio, and video content, all of which can be linked to KG entities.
Recent studies in Multi-Modal Entity Linking (MMEL) find that leveraging the multi-modal information can  significantly enhance the efficacy of conventional EL methods. 
\begin{defn}{\textbf{Multi-modal Entity Linking.}}\label{def:mmel}
A MMKG is denoted as $\mathcal{G}=\{\mathcal{E}, \mathcal{R}, \mathcal{A}, \mathcal{T}, \mathcal{V}\}$, where $\mathcal{E} = \{e_1, e_2, ..., e_i \}$ are the entity set. $\mathcal{M}=\{\mathbbm{g}, \mathbbm{r}, \mathbbm{a}, \mathbbm{v}, \mathbbm{s}\}$ are the graph structure, relation,  attribute, vision, and surface information, respectively. For example, $x^\mathbbm{s}_{e_1}$, $x^\mathbbm{v}_{e_1}$ denotes the name and visual information of $e_1$, respectively. The mention set is defined as $\mathcal{N} = \{m_1, ..., m_i\}$ with $\{x^\mathbbm{s}_{m_1}, ..., x^\mathbbm{s}_{m_i}\}$, $\{x^\mathbbm{v}_{m_1}, ..., x^\mathbbm{v}_{m_i}\}$ being the corresponding name and visual information. 
The objective of MMEL is to determine the linkage between entities and mentions, denoted by $(e_i, m_i)$, based on the multi-modal information $({x^\mathbbm{s}_{e_1}, ..., x^\mathbbm{v}_{e_1}}, {x^\mathbbm{s}_{m_1}, ..., x^\mathbbm{v}_{m_1}})$.
\end{defn}

\textbf{Progress:}  
Early MMEL research~\cite{DBLP:conf/acl/CarvalhoMN18,DBLP:conf/ecir/AdjaliBFBG20,DBLP:conf/dasfaa/ZhangLY21} focuses on fusing and expanding multi-modal data, such as merging visual and textual elements from media posts, to enhance textual mentions and predict corresponding KB entities. For example, DZMNED~\cite{DBLP:conf/acl/CarvalhoMN18} utilizes KG embeddings along with a blend of word-level and char-level lexical embeddings, a strategy crafted to adeptly manage the challenge of identifying previously unseen entities during testing. 
Zhang et al.~\cite{DBLP:conf/dasfaa/ZhangLY21} focus on the removal of noisy images to enhance performance. 
Subsequent research extends these methods, exploring strategies for integrating diverse multi-modal contexts and developing more reasonable multi-modal datasets~\cite{DBLP:conf/mm/GanLWWHH21,DBLP:journals/dint/ZhengWWQ22,DBLP:journals/dint/ZhengWWQB22,DBLP:conf/sigir/WangWC22,DBLP:conf/uai/YangHWXH023,DBLP:conf/kdd/LuoXWZXC23,DBLP:journals/corr/abs-2305-14725}. 
GHMFC~\cite{DBLP:conf/sigir/WangWC22}, for example, employs gated fusion and contrastive training for improved mention representations, while MIMIC~\cite{DBLP:conf/kdd/LuoXWZXC23} introduces a multi-grained interaction network for universal feature extraction. AMELI~\cite{DBLP:journals/corr/abs-2305-14725} implements an entity candidate retrieval pipeline, enhancing MMEL models using attribute information.

Recent explorations in MMEL mainly employ (V)PLMs for feature representation.
BERT~\cite{DBLP:conf/naacl/DevlinCLT19} is frequently used for textual processing~\cite{DBLP:conf/uai/YangHWXH023,DBLP:conf/acl/WangLZZPHMWWCXN23}, while CLIP~\cite{DBLP:conf/icml/RadfordKHRGASAM21} is preferred for visual encoding~\cite{DBLP:journals/corr/abs-2312-11816,DBLP:journals/corr/abs-2306-12725}. Typically, most parameters of these (V)PLMs remain frozen, complemented by focused fine-tuning strategies.
Among them, GEMEL~\cite{DBLP:journals/corr/abs-2306-12725} effectively combines LLaMA~\cite{DBLP:journals/corr/abs-2302-13971} for language processing and CLIP for visual encoding, showing the potential of GPT 3.5 in MMEL. Yang et al.~\cite{DBLP:conf/uai/YangHWXH023} introduce a multi-mention MMEL task that considers different mentions within the same context as a single sample, employing a multi-mention collaborative ranking method for testing to uncover potential connections between mentions.
Pan et al.~\cite{DBLP:journals/corr/abs-2211-00732} present Multi-modal Item-aspect Linking, focusing on linking short videos to related items in a short-video encyclopedia. GDMM~\cite{DBLP:conf/acl/WangLZZPHMWWCXN23} approaches MMEL by incorporating all three modalities: text, image, and table, adhering to a multi-modal encoder-decoder paradigm. DWE~\cite{DBLP:journals/corr/abs-2312-11816} augments visual features with detailed image attributes, like facial characteristics and scene features, enhancing textual representations using Wikipedia descriptions which bridges the gap between text and KG entities.

\textbf{Resources \& Benchmarks:} 
\textbf{\textit{(\rmnum{1})}} \textbf{SnapCaptionsKB}~\cite{DBLP:conf/acl/CarvalhoMN18}: A MMEL dataset featuring 12,000 manually labeled image-caption pairs, designed to capture diverse multi-modal interactions. Currently unavailable due to the General Data Protection Regulation (GDPR). In response, Adjali et al.~\cite{DBLP:conf/ecir/AdjaliBFBG20} develop an automated MMEL dataset construction tool from Twitter.
\textbf{\textit{(\rmnum{2})}} \textbf{M3EL}~\cite{DBLP:conf/mm/GanLWWHH21}: A dataset comprising 181,240 textual mentions and 45,297 images related to movies,  offering fine-grained annotations.
\textbf{\textit{(\rmnum{3})}} \textbf{NYTimes-MEL}~\cite{DBLP:conf/uai/YangHWXH023}:  Originates from the New York Times'~\cite{DBLP:conf/cvpr/TranMX20,DBLP:journals/corr/abs-2107-11970-boosting} images and captions, focusing on {\tt PERSON} entities. StanfordNLP tool~\cite{DBLP:conf/conll/0003DZM18} is used for NER in captions, where some entities were replaced with nicknames for mention construction. Similar to \cite{DBLP:conf/sigir/WangWC22}, it is enriched with images and 14 properties for each entity from Wikidata \cite{DBLP:conf/emnlp/XuLCPWSL23}, excluding samples with invalid entities or those without corresponding images.
\textbf{\textit{(\rmnum{4})}} \textbf{WikiData-Based Datasets}: Including \textbf{WikiDiverse}~\cite{DBLP:conf/acl/WangTGLWYCX22} and \textbf{WikiMEL}~\cite{DBLP:conf/sigir/WangWC22}, these datasets offer human-annotated mentions spanning diverse topics and entity types. WikiDiverse includes data from WikiNews categories like sports and technology, while WikiMEL collates mentions from Wikipedia and WikiData. 
\input{tab/mmel-bm}


\subsubsection{Multi-modal Entity Disambiguation}\label{sec:mmed}
In many studies, EL and Entity Disambiguation (ED) are often treated synonymously due to their methodological and task-setting similarities~\cite{DBLP:conf/acl/CarvalhoMN18,DBLP:conf/kdd/LuoXWZXC23}. However, it is crucial to distinguish between the two. While EL includes  the broader process of identifying and linking named entities in text to their corresponding entities in a KG, ED specifically focuses on resolving cases where a named entity might correspond to multiple potential candidates. In ED, each dataset sample typically includes a named entity alongside a set of candidates that bear close resemblance, highlighting the task's emphasis on disambiguating among these options \cite{DBLP:conf/acl/CarvalhoMN18}. 

Although EL and Entity Disambiguation (ED) are often treated synonymously in many studies due to their methodological and task-setting parallels~\cite{DBLP:conf/acl/CarvalhoMN18,DBLP:conf/kdd/LuoXWZXC23}, distinguishing between them is still vital. EL includes the broader process of identifying and linking named entities in text to their corresponding entries in a KG. In contrast, ED specifically targets resolving ambiguities when a named entity could match multiple candidates. ED emphasizes disambiguating among these potential candidates, often presented with a named entity and a closely related set of options in each dataset sample.

In Multi-modal Entity Disambiguation (MMED), methods leverage not just textual but also visual information to refine disambiguation. For example, DZMNED~\cite{DBLP:conf/acl/CarvalhoMN18} utilizes a convolutional LSTM for integrating multi-modal data. ET~\cite{DBLP:conf/ecir/AdjaliBFBG20} applies an Extra-Tree Classifier to effectively distinguish among ambiguous candidates. IMN~\cite{DBLP:conf/emnlp/ZhangH22} adopts meta-learning for multi-modal knowledge acquisition and a knowledge-guided transfer learning strategy, facilitating the extraction of cohesive representations across modalities.

%% file: tab/mmea-bm.tex
\begin{table}[ht]
    \centering
    \renewcommand\arraystretch{1.0}
        \caption{Comparison of MMEA results with (w/o) and without (w/o) surface forms (SF) on the DBP15K dataset \cite{DBLP:conf/aaai/0001CRC21}, where ``iter.'' signifies iterative learning applied. The symbol $\dagger$ indicates that the PLMs were applied for generating surface or attribute embeddings.  $\ast$ marks the results reproduced in \cite{chen2023rethinking,chen2023meaformer,DBLP:conf/mm/XuXS23}.
    }
    \label{tab:mmea-bm}
    \resizebox{0.97\linewidth}{!}{
    \begin{NiceTabular}{@{}llcccccc}
        \CodeBefore
        \rowcolors{2}{gray!10}{white}
        \columncolor{gray!1}{1}
        \rowcolor{gray!30}{1}
        \rowcolor{gray!30}{2}
        \Body
        \toprule[0.8pt]
        & \multirow{3}*{\makebox[2cm][c]{\textbf{Models}}} & \multicolumn{2}{c}{\footnotesize \textbf{DBP15K$_{\bm{\text{ZH-EN}}}$}} & \multicolumn{2}{c}{\footnotesize \textbf{DBP15K$_{\bm{\text{JA-EN}}}$}} & \multicolumn{2}{c}{\footnotesize \textbf{DBP15K$_{\bm{\text{FR-EN}}}$}} \\
        \cmidrule(r){3-4} \cmidrule(r){5-6} \cmidrule(r){7-8}
        & & {\scriptsize \textbf{H@1}}  & {\scriptsize \textbf{MRR}} & {\scriptsize \textbf{H@1}}  & {\scriptsize \textbf{MRR}} & {\scriptsize \textbf{H@1}}  & {\scriptsize \textbf{MRR}} \\
        \midrule[0.8pt]
        \parbox[t]{2mm}{\multirow{7}{*}{\rotatebox[origin=c]{90}{\footnotesize \textbf{w/o SF}}}} 
        & HMEA {\footnotesize {\cite{DBLP:journals/ijon/GuoTZZL21}}} & .540  & - & .531 & - & .484 & - \\
        & EVA {\footnotesize \cite{DBLP:conf/aaai/0001CRC21}}  &
        {.720}  & {.793} & {.716}  & {.792} & {.715}  & {.795} \\
        & MCLEA* {\footnotesize {\cite{DBLP:conf/coling/LinZWSW022}}} &
        {.726} & {.796} & {.719} & {.789} & {.719} & {.792} \\
        & {GEEA} {\footnotesize \cite{DBLP:journals/corr/abs-2305-14651}} &
        {.761} & {.827} & {.755} & {.827} & {.776} & {.844} \\
        & {MEAformer} {\footnotesize \cite{chen2023meaformer}} &
        {.772} & {.835} & {.769} & {.840} & {.771} & {.841} \\
        & {UMAEA} {\footnotesize \cite{chen2023rethinking}} &
        {.800} & {.860} & {.801} & {.862} & {.818} & {.877} \\
        & {DESAlign} {\footnotesize \cite{wang2024towards}} &
        {.810} & {.865} & {.811} & {.869} & {.826} & {.885} \\
        \midrule
        \parbox[t]{2mm}{\multirow{9}{*}{\rotatebox[origin=c]{90}{\footnotesize \textbf{w/o SF} (iter.)}}} 
        & EVA {\footnotesize \cite{DBLP:conf/aaai/0001CRC21}} &
        {.761} & {.814} & {.762} & {.817} & {.793} & {.847} \\
        & MSNEA* {\footnotesize {\cite{DBLP:conf/kdd/ChenL00WYC22}}} & .821 & .877 & .805 & .849 & .822 & .859 \\
        & PSNEA {\footnotesize {\cite{DBLP:conf/mm/NiXJCCH23}}} &
        {.811} & {.858} & {.807} & {.846} & {.843} & {.871} \\
        & MCLEA {\footnotesize {\cite{DBLP:conf/coling/LinZWSW022}}} &
        {.816} & {.865} & {.812} & {.865} & {.834} & {.885} \\
        & {MEAformer} {\footnotesize \cite{chen2023meaformer}} &
        {.847} & {.892} & {.842} & {.892} & {.845} & {.894} \\
        & {SKEA} {\footnotesize \cite{DBLP:journals/ipm/SuXYCJ23}} &
        {.849} & {.897} & {.844} & {.895} & {.878} & {.921} \\
        & {UMAEA} {\footnotesize \cite{chen2023rethinking}} &
        {.856} & {.900} & {.857} & {.904} & {.873} & {.917} \\
        & {DESAlign} {\footnotesize \cite{wang2024towards}} &
        {.868} & {.909} & {.871} & {.913} & {.882} & {.924} \\
        & {XGEA} {\footnotesize \cite{DBLP:conf/mm/XuXS23}} &
        {.876} & {.910} & {.878} & {.914} & {.889} & {.924} \\
        \midrule
        \parbox[t]{2mm}{\multirow{5}{*}{\rotatebox[origin=c]{90}{\footnotesize \textbf{w/ SF}}}} 
	& CLEM$\dagger$ {\footnotesize \cite{wu2022leveraging}}  & .854 & .879 & .885 & .904 & .936 & .952 \\
         & MSNEA* {\footnotesize {\cite{DBLP:conf/kdd/ChenL00WYC22}}} & .887 & .913 & .938 & .955 & .969 & .980 \\
        & EVA* {\footnotesize \cite{DBLP:conf/aaai/0001CRC21}} & {.929} & {.951} & {.964} & {.976} & {.990} & {.994} \\
        & MCLEA* {\footnotesize {
        \cite{DBLP:conf/coling/LinZWSW022}}} &
        {.926} & {.946} & {.961} & {.973} & {.987} & {.992} \\   
        & {MEAformer} {\footnotesize \cite{chen2023meaformer}} &
        {.949} & {.965} & {.978} & {.986} & {.991} & {.995} \\
        \midrule
        \parbox[t]{2mm}{\multirow{6}{*}{\rotatebox[origin=c]{90}{\footnotesize \textbf{w/ SF} (iter.)}}} 
        & MSNEA* {\footnotesize {\cite{DBLP:conf/kdd/ChenL00WYC22}}} & .896 & .922 & .942 & .958 & .971 & .982 \\
        & EVA {\footnotesize \cite{DBLP:conf/aaai/0001CRC21}} & .956 & .969 & .979 & .987 & {.995} & {.997} \\
        & {SKEA} {\footnotesize \cite{DBLP:journals/ipm/SuXYCJ23}} &
        {.913} & {.938} & {.923} & {.948} & {.978} & {.985} \\
        & MCLEA {\footnotesize {\cite{DBLP:conf/coling/LinZWSW022}}} &
        {.972} & {.981} & {.986} & {.991} & {.997} & {.998} \\
        & {XGEA} {\footnotesize \cite{DBLP:conf/mm/XuXS23}} &
        {.968} & {.978} & {.985} & {.991} & {.994} & {.996} \\
        & {MEAformer} {\footnotesize \cite{chen2023meaformer}} &
        {.973} & {.983} & {.991} & {.995} & {.996} & {.998} \\
        \bottomrule[0.8pt]
    \end{NiceTabular}
    \vspace{-4pt}
    }
\end{table}

%% file: tab/mmel-bm.tex

\begin{table}[!htbp]
    \centering
    \renewcommand\arraystretch{1.0}
        \caption{Comparison of MMEL results on the WikiMEL \cite{DBLP:conf/sigir/WangWC22} and Wikidiverse \cite{DBLP:conf/acl/WangTGLWYCX22} dataset.
    }
    \label{tab:mmel-bm}
    \resizebox{0.97\linewidth}{!}{
    \begin{NiceTabular}{@{}llcccccc}
        \CodeBefore
        \rowcolors{2}{gray!10}{white}
        \columncolor{gray!1}{1}
        \rowcolor{gray!30}{1}
        \rowcolor{gray!30}{2}
        \Body
        \toprule[0.8pt]
        & \multirow{3}*{\makebox[2cm][c]{\textbf{Models}}} & \multicolumn{3}{c}{\footnotesize \textbf{WikiMEL}} & \multicolumn{3}{c}{\footnotesize \textbf{Wikidiverse}} \\
        \cmidrule(r){3-5} \cmidrule(r){6-8} 
        & & {\scriptsize \textbf{H@1}}  & {\scriptsize \textbf{H@5}} & {\scriptsize \textbf{MRR}}  & {\scriptsize \textbf{H@1}}  & {\scriptsize \textbf{H@5}} & {\scriptsize \textbf{MRR}}  \\
        \midrule[0.8pt]
        \parbox[t]{2mm}{\multirow{5}{*}{\rotatebox[origin=c]{90}{\footnotesize \textbf{Text}}}} 
        & BLINK {\footnotesize {\cite{DBLP:conf/emnlp/WuPJRZ20}}} & .747  & .906 & .817 & .571 & .853 & .692 \\
        & BERT {\footnotesize \cite{DBLP:conf/naacl/DevlinCLT19}} & .748 & .905 & .818 & .558 & .831 & .674 \\
        & RoBERTa {\footnotesize {\cite{DBLP:journals/corr/abs-1907-11692}}} & .738 & .898 & .809 & .595 & .851 & .705 \\
        & GENRE {\footnotesize {\cite{DBLP:conf/iclr/CaoI0P21}}} & .601 & - & - & .601 & - & -\\
        & GPT 3.5 Turbo & .727 & - & - & .738 & - & -\\
        \midrule
        \parbox[t]{2mm}{\multirow{10}{*}{\rotatebox[origin=c]{90}{\footnotesize \textbf{Text + Vision}}}} 
        & JMEL {\footnotesize {\cite{DBLP:conf/ecir/AdjaliBFBG20}}} & .647 & .834 & .734 & .374 & .610 & .482\\
        & DZMNED {\footnotesize {\cite{DBLP:conf/acl/CarvalhoMN18}}} & .788 & .926 & .850 & .569 & .814 & .676 \\
        & GHMFC {\footnotesize {\cite{DBLP:conf/sigir/WangWC22}}} & .765 & .920 & .834 & .603 & .847 & .710 \\
        & CLIP {\footnotesize {\cite{DBLP:conf/icml/RadfordKHRGASAM21}}} & .832 & .945 & .882 & .612 & .852 & .717\\
        & ViLT {\footnotesize {\cite{DBLP:conf/icml/KimSK21}}} & .726 & .879 & .795 & .344 & .578 & .452\\
        & MMEL {\footnotesize {\cite{DBLP:conf/uai/YangHWXH023}}} & .715 & .917 & - & - & - & -\\
        & GEMEL {\footnotesize {\cite{DBLP:journals/corr/abs-2306-12725}}} & .826 & - & - & .863 & - & -\\
        & ALBEF {\footnotesize {\cite{DBLP:conf/nips/LiSGJXH21}}} & .786 & .918 & .846 & .606 & .813 & .699\\
        & METER {\footnotesize {\cite{DBLP:conf/cvpr/DouXGWWWZZYP0022}}} & .725 & .882 & .795 & .531 & .776 & .637\\
        & MIMIC {\footnotesize {\cite{DBLP:conf/kdd/LuoXWZXC23}}} & .880 & .964 & .918 & .635 & .864 & .734\\
        \bottomrule[0.8pt]
    \end{NiceTabular}
    \vspace{-8pt}
    }
\end{table}

%% file: 5.3-mmkgc.tex
\subsection{MMKG Inference}\label{sec:mmkgc}
MMKG data inherently contain missing elements, errors, and contradictions, making inference a critical task for KG completion. 
This stage, following extraction and fusion within the MMKG construction cycle, aims to bolster the model's reasoning abilities and deepen its understanding of the KG's overall knowledge.

\subsubsection{Multi-modal Knowledge Graph Completion}\label{sec:mmkgcc}
Multi-modal Knowledge Graph Completion (MKGC) plays a vital role in mining missing triples from existing KGs. This process involves three sub-tasks: Entity Prediction, Relation Prediction, and Triple Classification, defined as follows:

\begin{defn}{\textbf{MMKG Completion.}}\label{def:mmkgc}
A MMKG is denoted as $\mathcal{G}={\mathcal{E}, \mathcal{R}, \mathcal{A}, \mathcal{T}, \mathcal{V}}$, where $\mathcal{T} = {\mathcal{T_A}, \mathcal{T_R}}$. The goal of MKGC is to enrich the relational triple set $\mathcal{T}_{R}$ in \textbf{A-MMKGs} by identifying missing relational triples among existing entities and relations, leveraging  attribute triples $\mathcal{T}_A$. Specifically, Entity Prediction determines missing head/tail entities in queries $(h, r, ?)$ or $(?, r, t)$; Relation Prediction identifies missing relations in $(h, ?, t)$; and Triple Classification assesses the validity of given triples $(h, r, t)$ as true or false.
\end{defn}

\textbf{METHODS:}
It's noteworthy that most current MKGC tasks concentrate on Entity Prediction, commonly known as Link Prediction. Mainstream MKGC approaches primarily follow two paths: embedding-based and fine-tuning based (FT-based) methods.
Considering the intersection between MKGC and KGC methods, this section also discusses several typical KGC techniques to offer deeper insights into MKGC.

\textbf{\ul{Embedding-based Approaches}}
evolve from traditional KGE techniques \cite{DBLP:conf/nips/BordesUGWY13,DBLP:conf/iclr/SunDNT19}, adapting them to include multi-modal data, thus forming multi-modal entity embeddings. They're divided into modal fusion, modal ensemble, and negative sampling approaches:

\textbf{\textit{(\rmnum{1})}} \textbf{Modality Fusion} methods \cite{ DBLP:journals/corr/abs-2309-01169-EEMMKG, DBLP:journals/inffus/WangYCSL22, DBLP:journals/corr/abs-2206-13163} integrate multi-modal embeddings of entities with their structural embeddings for triple plausibility estimation. Early efforts, like IKRL \cite{DBLP:conf/ijcai/XieLLS17-IKRL}, use multiple TransE-based scoring functions \cite{DBLP:conf/nips/BordesUGWY13} for modal interaction. Subsequent developments, like TBKGC \cite{DBLP:conf/starsem/SergiehBGR18-TBKGC}, TransAE \cite{DBLP:conf/ijcnn/WangLLZ19-TransAE}, and MKBE \cite{DBLP:conf/emnlp/PezeshkpourC018-MKBE} further incorporate modalities such as textual numerical attributes. 
RSME \cite{DBLP:conf/mm/WangWYZCQ21-RSME} introduces gates for adaptive modal information selection. 
OTKGE \cite{DBLP:conf/nips/CaoXYHCH22-OTKGE} applies optimal transport for multi-modal fusion, while CMGNN \cite{DBLP:journals/tkde/FangZHWX23-CMGNN} implements a multi-modal GNN with cross-modal contrastive learning.
HRGAT \cite{DBLP:journals/tomccap/LiangZZ023-HRGAT} builds a hyper-node relational graph for multi-modal entity representation. CamE \cite{DBLP:conf/icde/XuZXXLCD23-CamE} introduces a triple co-attention module for biological KGs, and VISITA \cite{DBLP:conf/emnlp/LeeCLJW23-VISITA} develops a transformer-based framework which utilizes relation and triple-level multi-modal information for MKGC.

\textbf{\textit{(\rmnum{2})}}  \textbf{Modality Ensemble} methods train separate models using distinct modalities, combining their outputs for final predictions. For example, MoSE \cite{DBLP:conf/emnlp/ZhaoCWZZZJ22-MOSE} utilizes structural, textual, and visual data to train three KGC models and employs, using ensemble strategies for joint predictions. Similarly, IMF \cite{DBLP:conf/www/LiZXZX23-IMF} proposes an interactive model to achieve modal disentanglement and entanglement to make robust predictions.

\textbf{\textit{(\rmnum{3})}} \textbf{Modality-aware Negative Sampling} involves generating false triples to enhance a model's ability to differentiate between accurate and potentially erroneous KG triples.  During KG Embedding training, models map entities and relations to vectors, guided by both positive and negative samples, with their efficacy relying on the strategic selection and quality of negative samples to balance scoring between positive and negative instances.
Multi-modal data in KGs enhance traditional negative triple sampling \cite{DBLP:conf/nips/BordesUGWY13} by providing additional context for selecting higher-quality negative samples, thereby addressing a key performance bottleneck in KGC model training. Concretely,  MMKRL~\cite{DBLP:journals/apin/LuWJHL22-MMKRL} introduces adversarial training to MKGC, adding perturbations to modal embeddings. This pioneers the use of adversarial methods for augmenting MKGC models. Following this, VBKGC ~\cite{DBLP:journals/corr/abs-2209-07084-VBKGC} and MANS \cite{DBLP:journals/corr/abs-2304-11618-MANS} develop fine-grained visual negative sampling to better align visual with structural embeddings for more nuanced comparison training. MMRNS~\cite{DBLP:conf/mm/Xu0WZC22-MMRNS} introduces a relation-enhanced negative sampling method, utilizing a differentiable strategy to adaptively select high-quality negative samples.

\textbf{\ul{FT-based Approaches}} 
leverage pre-trained Transformer models such as BERT \cite{DBLP:conf/naacl/DevlinCLT19} and VisualBERT \cite{DBLP:journals/corr/abs-1908-03557}, capitalizing on their profound multi-modal comprehension for MKGC. 
These methods transform MMKG triples into token sequences, feeding them into PLMs~\cite{DBLP:journals/corr/abs-2212-05767}.

\textbf{\textit{(\rmnum{1})}} \textbf{Discriminative} 
strategies model KGC tasks as classification problems, with PLMs encoding textual information. 
KG-BERT~\cite{DBLP:journals/corr/abs-1909-03193-KGBERT}, a forerunner in this field, fine-tunes BERT for triple classification, assessing triple plausibility based on the model's positive probability. 
Subsequent methods introduce additional tasks like relation classification and triple ranking~ \cite{DBLP:conf/coling/KimHKS20-MTLKGC,DBLP:conf/www/WangSLZW021-STAR,DBLP:journals/corr/abs-2205-08012-CASCADER}, or  explore prompt tuning in KGC~\cite{DBLP:conf/acl/LvL00LLLZ22-PKGC,DBLP:conf/acl/ChenWSLL23-CSPROM,geng2023prompting}.  
FT-based MKGC methods emphasizes modal fusion over traditional KGC. 
Among them, 
MKGformer~\cite{DBLP:conf/sigir/ChenZLDTXHSC22-MKGformer} employs a hybrid Transformer for multi-level multi-modal fusion, treating MKGC as an MLM task and predicting masked entities by combining entity descriptions, relations, and images
SGMPT~\cite{DBLP:journals/corr/abs-2307-03591} extends MKGformer's capabilities by adding structural data integration through a graph structure encoder and a dual-strategy fusion module.

\textbf{\textit{(\rmnum{2})}} \textbf{Generative} models frame KGC as a sequence-to-sequence task~\cite{DBLP:conf/acl/SaxenaKG22-KGT5,DBLP:conf/www/XieZLDCXCC22-GenKGC,DBLP:conf/coling/ChenWLL22-KGS2S}, employing PLMs for text generation. 
KGLLaMA \cite{DBLP:journals/corr/abs-2308-13916-KGllama} and KoPA \cite{DBLP:journals/corr/abs-2310-06671-Kopa} explore the application  of LLMs with instruction tuning for generative KGC,  a relatively unexplored approach in MKGC, presenting a vast area for further exploration.

\input{tab/mmkgc-bm}
\textbf{Resources \& Benchmarks:}
\textbf{\textit{(\rmnum{1})}} \textbf{Initial MKGC Datasets}: Early MKGC research primarily utilize established KG benchmarks such as WordNet (WN9-IMG \cite{DBLP:conf/ijcai/XieLLS17-IKRL}, WN18-IMG \cite{DBLP:conf/mm/WangWYZCQ21-RSME}), MovieLens100K \cite{DBLP:conf/emnlp/PezeshkpourC018-MKBE}, YAGO-10 \cite{DBLP:conf/emnlp/PezeshkpourC018-MKBE}, and FreeBase (FB) \cite{DBLP:conf/starsem/SergiehBGR18-TBKGC}, extended with multi-modal information. For example, WN9-IMG incorporates images from ImageNet.
\textbf{\textit{(\rmnum{2})}} \textbf{Systematic MKGC Datasets}: 
Liu et al. \cite{DBLP:conf/esws/LiuLGNOR19} transforms FB15K, DB15K, and YAGO15K into MMKGs by adding web-crawled images and numeric modal data.  We benchmark those (M)KGC methods using this series of datasets as outlined in Table \ref{tab:mmkgc-bm}. Xu et al.~\cite{DBLP:conf/mm/Xu0WZC22-MMRNS} construct MKG-W and MKG-Y based on WikiData and YAGO, where the images are obtained through web search engines.
\textbf{\textit{(\rmnum{3})}} \textbf{Multi-faceted MKGC Datasets}: Recent MMKGs include a broader range of modal information, represent the evolution towards more sophisticated datasets.
For example, {MMpedia} \cite{wu2023mmpedia} is a scalable, high-quality MMKG developed using a novel pipeline based on DBpedia \cite{DBLP:conf/semweb/AuerBKLCI07}, designed to filter out non-visual entities and refine entity-related images through textual and type information.
{TIVA-KG} \cite{wang2023tiva} spans text, image, video, and audio modalities, built upon ConceptNet \cite{speer2017conceptnet}. It introduces triplet grounding, aligning symbolic knowledge with tangible representations. In a similar vein, {VTKG} \cite{DBLP:conf/emnlp/LeeCLJW23-VISITA} attaches entities and triplets with images, supplemented by textual descriptions for each entity and relation.

\begin{discussion}
In MKGC, extracting modal information using pre-trained encoders like VGG or BERT is essential. Embedding-based approaches generally freeze these encoders during training and use the extracted data to initialize modal embeddings, while FT-based methods optimize them, aligning more closely with the model’s inherent knowledge and memory. This leads to the underutilization of modal information in embedding-based methods, while FT-based methods struggle with complex KG structural information.
Furthermore, 
the challenge of missing modal information in real-world KGs is significant. Initial solutions involved random initialization of missing modal embeddings, as seen in early works \cite{DBLP:conf/ijcai/XieLLS17-IKRL, DBLP:conf/starsem/SergiehBGR18-TBKGC}. Recently, MACO~\cite{DBLP:conf/nlpcc/ZhangCZ23-MACO} introduce adversarial training to address this issue, but these methods remain basic, with a need for more innovative approaches.
\end{discussion}

\subsubsection{Multi-modal Knowledge Graphs Reasoning}\label{sec:mmkgrr}
MKGC methods typically focus on single-hop reasoning in MMKGs, which may limit the exploitation of KGs for multi-hop knowledge inference~\cite{DBLP:conf/iclr/DasDZVDKSM18}. 
Multi-modal knowledge graph reasoning (MKGR) aims to enable complex multi-hop reasoning on MMKGs, an area still in the early stages of research.

\begin{defn}{\textbf{MMKG Reasoning.}}\label{def:mmkgr}
MKGR predicts a missing query element in one of three forms: $(h, r, ?)$, $(h, ?, t)$, or $(?, r, t)$, where ``$?$'' denotes the missing element. 
The objective is to infer this element through a multi-hop reasoning path in $\mathcal{T}_{R}$ of an \textbf{A-MMKG}, where the path length is shorter or equal to $k$ hops, and $k$ is an integer greater than or equal to 1.
\end{defn}

MMKGR \cite{DBLP:conf/icde/Zheng0QYCZ23-MMKGR} combines a gate-attention network with feature-aware reinforcement learning for multi-hop reasoning in MMKGs, guided by analogical examples.  
TMR \cite{DBLP:journals/corr/abs-2306-10345-TMR} aggregates query-related topology features through an attentive mechanism to generate entity-independent features for effective MMKG reasoning under both inductive and transductive settings.
MarT \cite{DBLP:conf/iclr/000100LDC23-MART} introduces the concept of multi-modal analogical reasoning, akin to cross-modal link prediction but without explicitly defined relations. This task, framed as $(e_h, e_t):(e_q, ?)$, leverages a background MMKG for missing element ($?$) prediction. Its categorization under MKGR stems from its reliance on another triplet for tail (or head) entity prediction, differing from traditional MKGR in not requiring an explicit reasoning path. To facilitate this task, MarT presents a dedicated dataset (MARS) and an accompanying MMKG, MarKG. Additionally, they develop a model-agnostic baseline method inspired by structure mapping theory to address this unique reasoning challenge.

As this domain continues to evolve, it promises to become a pivotal direction in MMKG Inference, offering rich opportunities for groundbreaking discoveries and advancements.

%% file: tab/mmkgc-bm.tex
\begin{table}[!htbp]
    \centering
    \renewcommand\arraystretch{1.0}
        \caption{Comparison of MKGC results on FB15K-237 and DB15K datasets~\cite{DBLP:conf/esws/LiuLGNOR19}, with methods marked by $\dag$ utilizing only text information for KGC with PLMs.
    }
    \label{tab:mmkgc-bm}
    \resizebox{0.97\linewidth}{!}{
    \begin{NiceTabular}{@{}llcccccc}
        \CodeBefore
        \rowcolors{2}{gray!10}{white}
        \columncolor{gray!1}{1}
        \rowcolor{gray!30}{1}
        \rowcolor{gray!30}{2}
        \Body
        \toprule[0.8pt]
        & \multirow{3}*{\makebox[2cm][c]{\textbf{Models}}} & \multicolumn{3}{c}{\footnotesize \textbf{FB15K-237}} & \multicolumn{3}{c}{\footnotesize \textbf{DB15K}} \\
        \cmidrule(r){3-5} \cmidrule(r){6-8}
        & & {\scriptsize \textbf{H@1}}  & {\scriptsize \textbf{H@10}} & {\scriptsize \textbf{MRR}} & {\scriptsize \textbf{H@1}}  & {\scriptsize \textbf{H@10}} & {\scriptsize \textbf{MRR}} \\
        \midrule[0.8pt]
        \parbox[t]{2mm}{\multirow{10}{*}{\rotatebox[origin=c]{90}{\footnotesize \textbf{Embedding-based}}}} 
        & IKRL {\footnotesize {\cite{DBLP:conf/ijcai/XieLLS17-IKRL}}} & .232  & .493 & .309 & .111 & .426  & .222 \\
        & TBKGC {\footnotesize \cite{DBLP:conf/starsem/SergiehBGR18-TBKGC}}  &
        {.229}  & {.494}  & {.297} & {.108}& {.419}   & {.208} \\
        & MKBE {\footnotesize \cite{DBLP:conf/emnlp/PezeshkpourC018-MKBE}}  &
        {.258} & {.532} & {.347} & {.235} & {.513} & {.332} \\
        & VBKGC {\footnotesize {\cite{DBLP:journals/corr/abs-2209-07084-VBKGC}}} &
        {.239} & {.478} & {.332} & {-} & {-} & {-} \\
        & MANS {\footnotesize {\cite{DBLP:journals/corr/abs-2304-11618-MANS}}} &
        {-} & {-} & {-} & {.204} & {.550} & {.332} \\
        & MoSE {\footnotesize {\cite{DBLP:conf/emnlp/ZhaoCWZZZJ22-MOSE}}} &
        {-} & {.565} & {.281} & {-} & {-} & {-} \\
        & MMRNS {\footnotesize {\cite{DBLP:conf/mm/Xu0WZC22-MMRNS}}} &
        {-} & {-} & {-} & {.231} & {.510} & {.327} \\
        & HRGAT {\footnotesize {\cite{DBLP:journals/tomccap/LiangZZ023-HRGAT}}} &
        {.271} & {.542} & {.366} & {.597} & {.694} & {.630} \\
        & {IMF} {\footnotesize \cite{DBLP:conf/www/LiZXZX23-IMF}} &
        {.287} & {.593} & {.389} & {.427} & {.604} & {.485} \\
        & {VISITA} {\footnotesize \cite{DBLP:conf/emnlp/LeeCLJW23-VISITA}} &
        {.287} & {.572} & {.381} & {-} & {-} & {-} \\
        \midrule
        \parbox[t]{2mm}{\multirow{10}{*}{\rotatebox[origin=c]{90}{\footnotesize \textbf{FT-based}}}} 
        & MTL-KGC$\dag$ {\footnotesize {\cite{DBLP:conf/coling/KimHKS20-MTLKGC}}}  & .172 & {.458} & .267 &  & {-} & - \\
        & StAR$\dag$ {\footnotesize {\cite{DBLP:conf/www/WangSLZW021-STAR}}}  &
        {.205} & {.482} & {.269} & {-} & {-} & {-} \\
        & SimKGC$\dag$ {\footnotesize {\cite{DBLP:conf/acl/0046ZWL22-SIMKGC}}}  &
        {.249} & {.511} & {.336} & {.} & {-} & {.} \\
        & {KGT5}$\dag$ {\footnotesize \cite{DBLP:conf/acl/SaxenaKG22-KGT5}}  &
        {.210} & {.414} & {.276} & {-} & {-} & {-} \\
        & {GenKGC}$\dag$ {\footnotesize \cite{DBLP:conf/www/XieZLDCXCC22-GenKGC}}  &
        {.192} & {.439} & {-} & {-} & {-} & {-} \\
        & {KG-S2S}$\dag$ {\footnotesize \cite{DBLP:conf/coling/ChenWLL22-KGS2S}}  &
        {.257} & {.498} & {.336} & {-} & {-} & {-} \\
        & {CSProm-KG}$\dag$ {\footnotesize \cite{DBLP:conf/acl/ChenWSLL23-CSPROM}}  &
        {.269} & {.538} & {.358} & {-} & {-} & {-} \\
        & {MKGformer} {\footnotesize \cite{DBLP:conf/sigir/ChenZLDTXHSC22-MKGformer}} &
        {.256} & {.504} & {-} & {-} & {-} & {-} \\
        & {SGMPT} {\footnotesize \cite{DBLP:journals/corr/abs-2307-03591}} &
        {.252} & {.510} & {-} & {-} & {-} & {-} \\
        \bottomrule[0.8pt]
    \end{NiceTabular}
    \vspace{-4pt}
    }
\end{table}

%% file: 5.4-mmapp.tex
\subsection{MMKG-driven Tasks}\label{sec:mmapp}
In this section, we explore several key directions where MMKGs have shown notable impact in downstream task applications, specifically in retrieval, reasoning, pre-training, and industrial applications. 

\subsubsection{Retrieval}
As discussed in \mbox{\S\,\ref{sec:mm-kg-cst}}, several MMKGs could naturally support retrieval related tasks:
ImageGraph~\cite{DBLP:journals/tip/LiuWZT17} connects a query to its top-K nearest neighbors, expanding via Bayes similarity-weighted edges up to a certain graph depth;
IMGpedia~\cite{DBLP:conf/semweb/FerradaBH17}, formatted in RDF, links visual descriptors and similarity relations with image metadata from DBpedia Commons, supporting SPARQL-based retrieval based on visual similarity, metadata, or DBpedia resources; 
VisualSem~\cite{DBLP:journals/corr/abs-2008-09150} use a neural multi-modal retrieval model that processes both images and sentences to retrieve entities in the KG with pre-trained CLIP~\cite{DBLP:conf/icml/RadfordKHRGASAM21} as the encoder. 
Chen et al.~\cite{attrDASFAA21} enhance MNER by searching the entire MMKG to acquire knowledge about poster images, using (mention, candidate entity) pairs from post text and MMKG for efficient image knowledge retrieval through iterative breadth-first traversal.

\mbox{\S\,\ref{sec:kgret}} introduces MMKG-driven Cross-modal Retrieval methods like MKVSE \cite{DBLP:journals/tomccap/FengHP23}, which scores intra- and inter-modal relations in MMKGs using WordNet path similarity and co-occurrence correlations (Fig. \ref{fig:MMKG4CMR}), improving Image-Text Retrieval through GNN-based embeddings. 
Zeng et al.~\cite{DBLP:conf/aaai/ZengJBL23} provide a multi-modal knowledge hypergraph (MKHG) for linking diverse data in MMKGs and retrieval databases. a hyper-graph construction module with varied hyper-edges, multi-modal instance bagging for instance selection, and a diverse concept aggregator for sub-semantic adaptation, thus advancing representation learning in image retrieval.
Huang et al.~\cite{DBLP:journals/corr/abs-2206-13163} propose a unified continuous learning framework, iteratively updating the MMKG with MKGC as the target task and subsequently pre-training an MMKG-based VLM, using image-text matching as the core pre-training task without the need for paired image-text training data.

\subsubsection{Reasoning \& Generation}
\begin{figure}[!htbp]
  \centering
   \vspace{-1pt}
\includegraphics[width=0.99\linewidth]{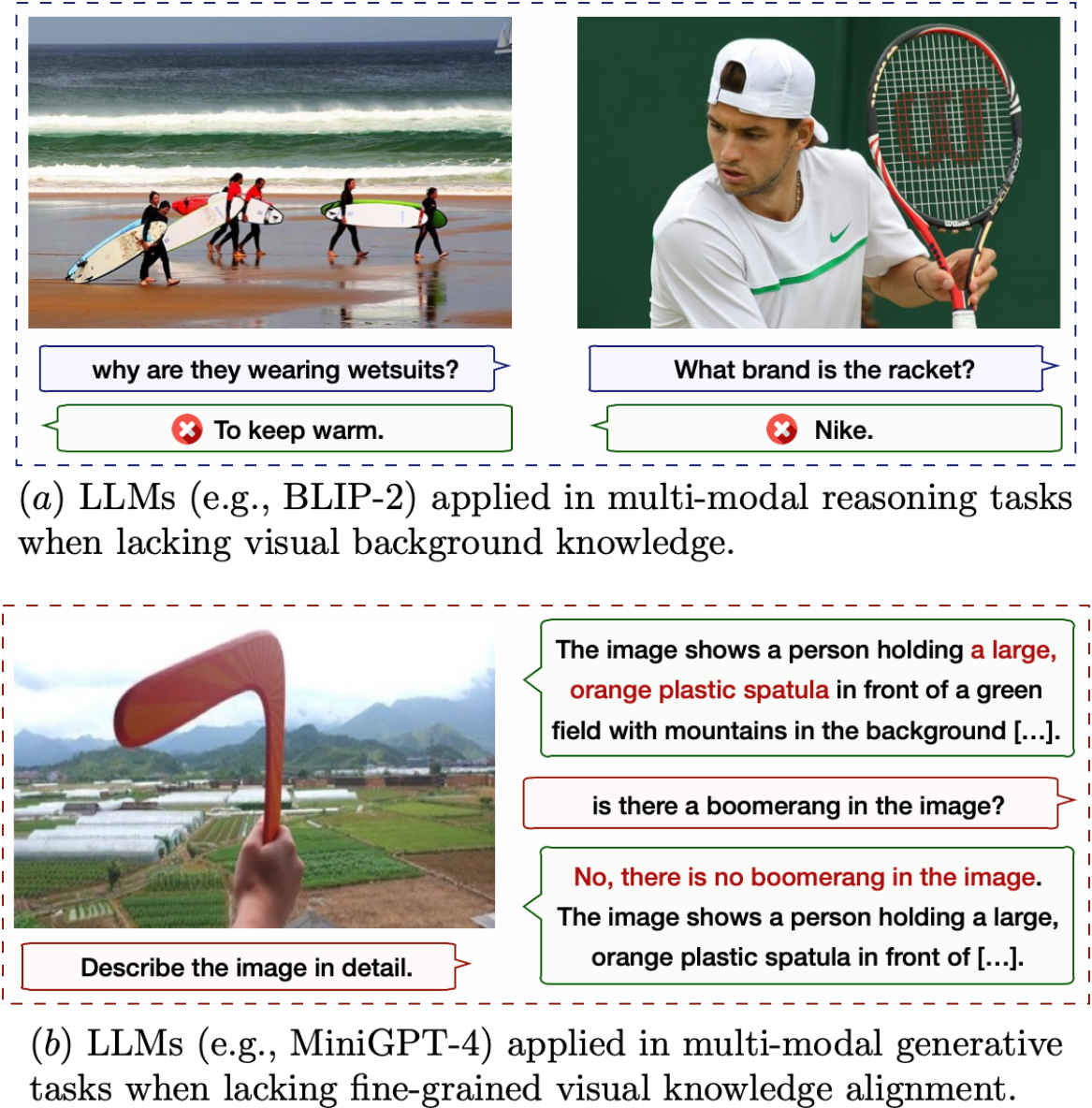}
  \caption{Examples of limited cross-modal knowledge alignment ability in current multi-modal LMMs~\cite{zha2023m2conceptbase}, specifically (a) BLIP-2\cite{DBLP:conf/icml/0008LSH23} and (b) MiniGPT-4~\cite{DBLP:journals/corr/abs-2304-10592}, leading to multi-modal hallucinations.}
  \label{fig:kgmm_case}
  \vspace{-9pt}
\end{figure}
In \mbox{\S\,\ref{sec:kgr}} and \mbox{\S\,\ref{sec:kggen}}, we highlight that multi-modal reasoning and generation tasks often require an extensive range of specialized knowledge, typically involving long-tail information that goes beyond everyday experiences. KGs are crucial in these scenarios, serving as structured repositories for such diverse knowledge. However, there exists a notable gap between KGs and multi-modal tasks, as current methods frequently depend on indirect approaches like modal transformation for knowledge representation, retrieval, and interaction in multi-modal contexts. A significant challenge arises in tasks requiring visual common sense, where models may falter due to limited cross-modal alignment capabilities, leading to multi-modal hallucinations as evidenced in Fig.~\ref{fig:kgmm_case}. Recent works~\cite{zha2023m2conceptbase} demonstrate that MMKGs can effectively bridge this gap, enhancing the potential of multi-modal methods and offering a robust solution to multi-modal hallucinations in the era of LLMs.
Specifically,
Zha et al.~\cite{zha2023m2conceptbase} introduce {M\textsuperscript{2}ConceptBase} (detailed in \mbox{\S\,\ref{sec:mm-kg-cst}}), a multi-modal conceptual MMKG. They develop a pipeline using {M\textsuperscript{2}ConceptBase} to improve knowledge-based VQA performance  by retrieving multi-modal concept descriptions and crafting instructions to refine answers with MLLMs.
Zhao et al.~\cite{DBLP:journals/corr/abs-2107-11970-boosting} introduce an Image Captioning method utilizing an MMKG that associates visual objects with named entities, leveraging external multi-modal knowledge from Wikipedia and Google Images for supplementary. The MMKG, after processing through a GAT~\cite{DBLP:conf/iclr/VelickovicCCRLB18}, feeds its final layer output into a Transformer decoder, enhancing the precision of entity-aware caption generation.
Jin et al.~\cite{jin2023self} involve the MMKG into multi-modal summarization in a similar manner.

\subsubsection{Pre-training}
Building on \mbox{\S\,\ref{sec:kgplm}}, we now shift our focus to MMKG-based VLM pre-training, moving away from traditional KG-based methods. Current VLMs mainly pre-train on basic image-text pairs, often overlooking the rich knowledge connections among concepts across modalities. To bridge this gap, two representative MMKG-based methods have emerged:
\textbf{\textit{(\rmnum{1})} Triple-level}
methods views triples as independent knowledge units,  implicitly embedding the (head entity, relationship, tail entity) structure into the VLM’s embedding space. For example,
Pan et al.~\cite{DBLP:conf/nips/PanYHSH22} present Knowledge-CLIP to integrate knowledge-based objectives into the CLIP framework using MMKGs such as Visual Genome~\cite{DBLP:journals/ijcv/KrishnaZGJHKCKL17} and VisualSem~\cite{DBLP:journals/corr/abs-2008-09150}.  By encoding both textual and visual entities along with their relations using the CLIP encoder and fusing these via a multi-modal Transformer, Knowledge-CLIP effectively optimizes pre-training with a triple-based loss function and boosts CLIP's performance in multiple multi-modal tasks.
\textbf{\textit{(\rmnum{2})} Graph-level}
methods capitalize on the structural connections among entities in a global MMKG. By selectively gathering multi-modal neighbor nodes around each entity featured in the training corpus, they apply techniques such as GNNs or concatenation to effectively incorporate knowledge during the pre-training process.
Gong et al.~\cite{DBLP:journals/corr/abs-2302-06891} aggregate various knowledge-view of the entities in MMKG (i.e., embeddings of neighbors connected by specific relations) to obtain their knowledge representation. These, combine with the entities' textual and visual embeddings, are integrated into CLIP's similarity computation process for multi-modal knowledge pre-training.
Li et al.~\cite{DBLP:journals/corr/abs-2309-13625} introduce GraphAdapter for CLIP, a method that leverages dual-modality structure knowledge through a unique dual knowledge graph, comprising textual and visual knowledge sub-graphs which represent semantics and their interrelations in both modalities. GraphAdapter enables textual features of prompts to utilize task-specific structural knowledge from both textual and visual domains, enhancing CLIP's classifier performance in downstream tasks. 

\subsubsection{AI for Science}
AI for science refers to the application of AI techniques into scientific disciplines to drive discovery, innovation, and understanding. It employs AI to analyze, interpret, and predict complex scientific data, effectively supplementing traditional scientific methods with advanced computational tools.
Within this domain, the concept of MMKGs is broadened beyond the conventional text and image modality to incorporate a diverse array of scientific data, including molecules, proteins, genes, drugs, and disease information~\cite{maclean2021knowledge}. This broader definition of ``multi-modality'' not only enriches the scope and depth of scientific research with varied data sources but also introduces new vitality and potential application value into the MMKG field.

In biology, MMKGs effectively integrate domain-specific data sources~\cite{DBLP:journals/bib/BonnerBYSEBHH22} like Uniprot for proteins~\cite{uniprot2019uniprot}, ChEMBL for small molecule-protein interactions~\cite{gaulton2012chembl}, SIDER for side effects~\cite{kuhn2016sider}, and Signor for protein-protein interactions~\cite{lo2023signor}. These well-curated sources provide robust information to MMKGs. Additionally, data mined from extensive literature using NLP methods~\cite{kilicoglu2012semmeddb,percha2018global} further enrich MMKGs with diverse scientific insights.
In those MMKGs, entities represent specific biological elements such as drugs or proteins, with relations depicting their experimentally verified interactions. These links, often augmented with additional attributes like molecular structures or external identifiers, can be directional to indicate causality, such as a drug causing a side effect~\cite{drkg2020}.

However, in the process of modeling complex biological systems, these MMKGs face challenges in MKGC due to data incompleteness, which hinders downstream applications. To address this, Xu et al.~\cite{DBLP:conf/icde/XuZXXLCD23} 
create a co-attention-based multi-modal embedding framework, merging molecular structures and textual data. It features a Triple Co-Attention (TCA) fusion module for unified modality representation and a relation-aware TCA for detailed entity-relation interactions, enhancing missing link inference.
Moreover, biological MMKGs have also broadened their applications in \textbf{drug discovery}, extending beyond KGC to facilitate advanced tasks by leveraging rich graph knowledge.
Lin et al.~\cite{DBLP:conf/ijcai/LinQWMZ20} convert DrugBank data into an RDF graph using Bio2RDF, linking various biological entities and extracting triples for their KGNN framework. This framework predicts drug-drug interactions, adapting spatial-based GNN approaches to MMKGs by aggregating neighborhood information, which efficiently maps drugs and their potential interactions within the MMKG.
Fang et al.~\cite{DBLP:conf/aaai/FangZYZD0Q0FC22,fang2023knowledge} develop a chemical-oriented MMKG to summarize elemental knowledge and functional groups. They introduce an element-guided graph augmentation strategy for contrastive pre-training, exploring atomic associations at a microscopic level. Their approach, integrating functional prompts during fine-tuning, significantly improves molecular property prediction and yields interpretable results.
Zhang et al.~\cite{DBLP:conf/iclr/ZhangBL0HDZLC22} construct a large-scale MMKG containing the Gene Ontology and related proteins. They implement a contrastive learning approach with knowledge-aware negative sampling to optimize MMKG and protein embeddings, enhancing protein interaction and function prediction.
Cheng et al.~\cite{DBLP:conf/aaai/0008LBC023} create an MMKG for protein science, integrating the Gene Ontology and Uniprot knowledge base. They develop a system for protein analysis, aiding predictions related to protein structure, function, and drug molecule binding, and supporting biological question answering.
MMKGs thus serve not only as tools for direct query and pattern discovery but also as invaluable resources for augmenting and refining the performance of diverse computational tasks in drug discovery.


\input{6.2-mmkg-indus}

%% file: 6.2-mmkg-indus.tex
\begin{figure}[!htbp]
    \centering
    \includegraphics[width=0.95\linewidth]{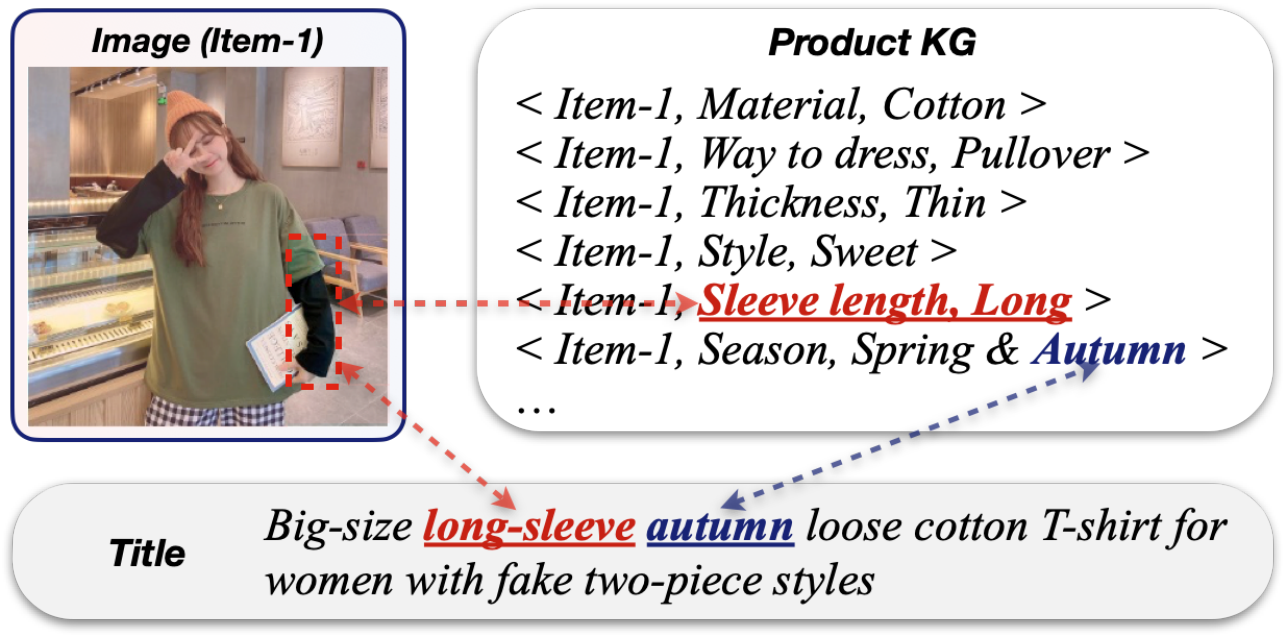}
    \caption{Illustration of multi-modal product data in MMPKGs~\cite{DBLP:conf/mm/ZhuZZYCZC21}. Each product is represented by a title, an image, and part of the Product Knowledge Graph (PKG) detailing objective properties through triples as \textit{({item}, {property}, {value})}. The pre-training process of MMPKG enhances the VLM's natural visual grounding abilities and its comprehension of multi-modal knowledge specific to the E-commerce domain.}
    \label{fig:mmkg_appli_ec}
    \vspace{-8pt}
\end{figure}

\subsubsection{Industry Application} \label{sec:indus-mm4kg}
\textbf{E-commerce} greatly benefits from Multi-modal Product KGs (MMPKGs), which amalgamate images, text, and structured knowledge for an enriched product representation, as depicted in Fig.~\ref{fig:mmkg_appli_ec}. This integration supports key applications such as product management, comparison, and recommendation. K3M~\cite{DBLP:conf/mm/ZhuZZYCZC21} framework advances e-commerce platforms 
by leveraging MMPKGs to refine product representations, employing techniques such as masked object prediction, masked language reconstruction, and link prediction for pre-training and multi-modal knowledge integration.
Wang et al.~\cite{DBLP:conf/acl/Wang0LLCJHXG23} introduce FashionKLIP, an MMKG-enhanced VLM for e-commerce, by developing and integrating the FashionMMKG into a CLIP-style model for image-text retrieval. This approach uses contrastive learning for modal alignment and conceptual matching through visual prototypes from FashionMMKG for training.

MKGAT~\cite{DBLP:conf/cikm/SunCZWZZWZ20} applies MMKGs to \textbf{movie and restaurant recommendation} systems, 
using a Collaborative MMKG (CMMKG) that merges user behavior with multi-modal item data. This model adopts entity-specific encoders and a GAT for entity representation, leveraging TransE for knowledge space learning. 
CKGC~\cite{DBLP:conf/mm/CaoSW0WY22} further categorizes traditional relations in MMKG into two types: descriptive attributes and structural connections, employing cross-modal contrastive learning for more effective node representation in recommendation.

%% file: 7-kg-mm-fut.tex
\section{Challenges and Opportunities}\label{sec:fut}

\input{7.0-com}
\input{7.1-llm}

%% file: 7.0-com.tex
MMKGs, along with KGs, aim to mitigate the scarcity of long-tail knowledge in various tasks, reflecting the patterns of frequent co-occurrences and human experience in the real world. Current research operates under the optimistic assumption that an infinitely expansive MMKG could contain  a near-complete spectrum of relevant world knowledge, providing the necessary information to address all multi-modal challenges effectively. However, critical questions persist: How can we acquire \textbf{ideal multi-modal knowledge}? What characteristics should the ideal MMKG possess, and can it \textbf{accurately reflect the human brain's advanced understanding of world knowledge}? Moreover, when compared to the \textbf{knowledge capabilities of LLMs}, does MMKG offer unique, irreplaceable benefits? Exploring these inquiries is vital in our ongoing exploration of this domain.

\subsection{MMKG Construction \& Acquisition}\label{sec:fut-mmkg}
\textbf{\textit{(\rmnum{1})}} 
As outlined in \mbox{\S\,\ref{sec:mm-kg-cst}},  MMKG construction primarily involves two paradigms: annotating images with KG symbols or grounding KG symbols to images.
Recent developments, as highlighted in \cite{DBLP:conf/emnlp/LeeCLJW23-VISITA}, start to explore a new path, \textbf{aligning locally extracted triples from multiple images with large-scale KGs}, which can be regarded as a mixture of MMKG and hyper-MMKG. The advantage of this hybrid approach is twofold: it not only extends the coverage of image quantity, as seen in the first paradigm, but also incorporates the extensive knowledge scale characteristic of the second. It promotes the generation of large-scale, triple-level multimodal information, posing both opportunities and challenges for future work in Multi-modal Entity Alignment and MMKG-driven applications like MLLM Pre-training and VQA.

\textbf{\textit{(\rmnum{2})}} 
Refining and aligning fine-grained knowledge within MMKGs is crucial.  An ideal MMKG should be hierarchical, possessing deep levels with detailed and abstract multi-modal knowledge. Such a structure  allows for the automatic decomposition of large-scale cross-modal data, enabling a single image to ground multiple concepts~\cite{huang2023open}. Moreover, segmentation represents an advanced requirement for grounding. With technologies like \textit{Segment Anything}~\cite{DBLP:journals/corr/abs-2304-02643} already in place, such approaches can significantly reduce background noise impact in visual modalities. Thus, evolving towards \textbf{segmentation-level, hierarchical, and multi-grained} MMKGs marks a significant future direction.

\textbf{\textit{(\rmnum{3})}} 
In visual modalities, we hold that abstract concepts should correspond to abstract visual representations, while concrete concepts align with specific visuals. For example, general concepts like cats and dogs manifest in the brain as generic, averaged visual animal images, whereas specific qualifiers, such as "\textit{Alaskan sled dogs}", provide clarity, similar to route-based image retrieval in MMKGs. Additionally, we also posit that every concept, visualizable or not, can be associated with certain modal representations. The abstract concept of ``\textit{mind}'', for example, may evoke images of ``\textit{brains}'' or ``\textit{people thinking}'', still showing MMKGs' ability to represent non-visualizable concepts. This perspective contrasts with previous views~\cite{DBLP:conf/dasfaa/JiangLLLXWLX22,DBLP:journals/bdr/PengHHY23}.
Interestingly, in human cognition, rarer concepts, such as ``\textit{unicorns}'', are often more vividly depicted. If we know a \textit{unicorns} only as a horned horse, this specific image is what we remember, rather than a horned seal or lion. This mirrors MMKG data structuring: concepts with fewer images are represented more distinctly, while those with more images are generalized and blurrier.

\textbf{\textit{(\rmnum{4})}} 
\textbf{Efficiency in MMKG storage and utilization} remains a concern.  Despite traditional KGs' lightweight nature and vast knowledge storage with minimal parameters, MMKGs demand more space, challenging efficient data storage and application across tasks. Enhancing efficiency might involve embedding multi-modal information into dense spaces as a temporary solution. Future research should strive to improve usage and storage efficiency without sacrificing MMKG's interpretability and structural integrity, a delicate balance that presents a continuing challenge.

\textbf{\textit{(\rmnum{5})}} 
\textbf{Quality control} in MMKGs 
introduces unique challenges with multi-modal (e.g., visual) content such as incorrect, missing, or outdated images. 
Limited fine-grained alignment between images and text in existing MMKGs and the noise from automated MMKG construction methods necessitate developing quality control techniques, possibly by assigning scores based on modal information quality.
Given the dynamic nature of world knowledge, regularly updating MMKGs is essential. An important research direction lies in efficiently implementing \textbf{multi-modal knowledge conflict detection and updates}.
The development of dynamic, temporal, and even spatiotemporal MMKGs~\cite{DualMatch} is also crucial, enhancing their adaptability to diverse environments and user needs.
Moreover, cross-lingual MMKGs can facilitate intercultural communication by enabling understanding and collaboration across languages and cultures, overcoming understanding barriers and supporting global cultural sharing.

\subsection{KG4MM Tasks}
In evaluating KG-aware multi-modal tasks, it's crucial to discern the unique advantages of multi-modal knowledge, especially compared to large-scale textual or multi-modal corpora. 
A pivotal question is whether structured (MM)KGs offer irreplaceable benefits that maximize their potential.
Additionally, we should consider whether non-LLM models augmented by (MM)KG can rival or outperform MLLMs in specific tasks, providing compelling reasons to support the future development. 


\textbf{Multi-modal Content Generation:}
Current applications of MMKGs in multi-modal content generation are quite limited. Most existing efforts only integrate KGs to supplement additional context beyond datasets or to connect different visual scenes. Future developments should aim for larger, more detailed MMKGs to employ multi-modal structural data in training, fostering more controlled and logically coherent generation and mitigating hallucinations.

\textbf{Multi-modal Task  Integration}:
Different domains currently evolve independently with limited cross-interaction. In Cross-Modal Retrieval (CMR), (MM)KGs are widely employed for information enhancement, whereas in knowledge-based VQA, the focus is mainly on dense vector retrieval and modality conversion techniques. This highlights the potential for future advancements like integrating KG-based CMR methods into KG-based VQA. In a similar vein, generation tasks can enhance retrieval, reasoning, and discrimination, with knowledge-enhanced discrimination tasks playing a key role in refining answers for other tasks. As knowledge-intensive multi-modal tasks gain prominence, merging these distinct domains with (MM)KG at the core will becomes crucial.

\textbf{Challenges in Scaling MMKG for Multi-modal Tasks}:
As discussed in \mbox{\S\,\ref{sec:mmapp}}, MMKG-driven tasks often emphasize retrieval-related activities, leveraging the natural database-like capabilities of MMKGs. However, the utilization of large-scale MMKGs in varied tasks, especially reasoning, is still nascent with limited exploratory studies. For example, Zha et al.~\cite{zha2023m2conceptbase} enhance knowledge-based VQA by employing multi-modal concept descriptions and integrating MLLMs for refined answers. Nevertheless, these methods only use MMKGs as ``\textit{key:value}'' based retrieval databases, not fully leveraging their multi-modal structured capabilities.

The constrained utilization of MMKGs in diverse tasks can be attributed to several factors.
\textbf{\textit{(\rmnum{1})} Non-Uniform Organization and Ontology of MMKGs:} 
Current MMKGs, lacking a standardized format, vary significantly in their focal points and the knowledge domains they cover for each downstream task. Predominantly, MMKGs cater to encyclopedic or trivia knowledge~\cite{DBLP:journals/corr/abs-2302-06891,DBLP:conf/cikm/ZhangWWLX23,wu2023mmpedia,zha2023m2conceptbase}, with commonsense and scientific  related MMKGs~\cite{wang2023tiva,DBLP:conf/emnlp/LeeCLJW23-VISITA} being notably scarce. Moreover, the ``non-visualizable'' nature of some abstract knowledge components restricts their practical application~\cite{DBLP:conf/dasfaa/JiangLLLXWLX22,wu2023mmpedia}.
\textbf{\textit{(\rmnum{2})} Storage and Processing Overheads:}
The substantial storage space requirements and extended processing times for large-scale MMKGs hinder their extensive adoption. Conversely, small-scale MMKGs frequently offer limited value for cross-task generalization.
\textbf{\textit{(\rmnum{3})}  Data Timeliness and Completeness Issues in MMKGs} heightens the risk of multi-modal hallucinations.
\textbf{\textit{(\rmnum{4})}  Comparative Advantages of LLMs and MLLMs:} LLMs and MLLMs excel in generalizability and AGI potential across various domains~\cite{zhang2024mmllms}, complementing the interpretability and editing flexibility of MMKGs. While MMKGs bring unique value, their development, maintenance, and application also involve certain costs. The evolving feedback from downstream tasks will continue to shape the industry's perspective on their respective roles and potentials.

\textbf{Unlocking the Potential of Large-Scale MMKGs for Multi-Modal Tasks.}
\textbf{\textit{(\rmnum{1})} Integration with Non-text Modalities:}
 Future downstream tasks driven by large-scale MMKGs can integrate methods from current KG-driven VQA methods, placing equal emphasis on non-textual modalities.  This may further involve using modality projection or adapters for cross-modal alignment~\cite{DBLP:journals/corr/abs-2309-13625,DBLP:journals/corr/abs-2309-01516}, along with multi-modal GNN methods~\cite{DBLP:journals/corr/abs-2310-07478} and modal feature decoupling techniques to enrich the granularity and hierarchy of multi-modal information~\cite{DBLP:conf/aaai/ChenHCGZFPC23}.
\textbf{\textit{(\rmnum{2})} Rich Semantic MMKG Construction:}
MMKG data can transcend traditional specialized or general formats. By developing task-specific pipelines, multi-modal datasets can be converted into MMKGs with enhanced semantics, using existing KGs as foundational references or bridges. This process can not only augments MLLM training with structured multi-modal input but also enriches the MMKG community with valuable, semantically rich datasets.
\textbf{\textit{(\rmnum{3})} Reconstruction of Multi-Modal Tasks with LLM:}
Combining LLM's text understanding and generation capabilities, multi-modal tasks can be restructured. Transforming KG-driven multi-modal tasks into in-MMKG-tasks, such as MKGC, MMEA, can enhance domain integration. There are already some attempts in this direction~\cite{pahuja2023bringing}, which will be discussed in-depth later.

\subsection{MM4KG Tasks}
\textbf{MMKG Fusion:}
Currently, Multi-modal Entity Alignment (MMEA) primarily targets Entity Alignment within A-MMKGs, treating images and other modalities as only attributes. 
Future research should rethink the role of images, incorporating multi-image Entity Alignment, exploring N-MMKG level alignment, and even aligning image entities with text entities. Additionally, the model's time and space efficiency, its generalizability, and usability across various scenarios (e.g., uncertainly missing modality~\cite{chen2023rethinking}) are critical factors that require careful consideration and evaluation. 
Considering MMEA methods often exhibit quadratic or higher complexity, addressing scalability is crucial, especially when aligning entities in large-scale~\cite{LargeEA, ClusterEA}, multi-modal KGs involving images, attributes, and timestamps~\cite{DBLP:journals/corr/abs-2310-05364}. 
Another essential aspect of future work is to identify practical application scenarios for MMEA and to determine if MMEA can intersect with other multi-modal tasks to uncover new research directions, such as AI-for-Science.

\textbf{MMKG Inference:} Similar to MMEA, current MKGC methods focus mainly on A-MMKGs, with modality missing being a common issue~\cite{DBLP:conf/nlpcc/ZhangCZ23-MACO}.  Future work should explore more principle-based analyses, such as investigating why and how image modality aids KGC and devising strategies to enhance this effect. A key consideration is how to explain these benefits clearly, especially in N-MMKG contexts~\cite{DBLP:conf/iclr/000100LDC23-MART}. Additionally, expanding MKGC to include a variety of modalities like numeric, audio, and video is also essential for advancing MMKG inference capabilities, accommodating any extra modality. Another future direction for MKGC is integrating tasks into MLLMs. Initiatives like KGLlaMA \cite{DBLP:journals/corr/abs-2308-13916-KGllama} and KoPA \cite{DBLP:journals/corr/abs-2310-06671-Kopa} have  explored LLM-based KGC. However, integrating more modalities, enabling complex reasoning, and expanding datasets are challenges that lie ahead.

\textbf{Transfer Multi-modal Task into MMKG Paradigm:}
Pahuja et al. \cite{pahuja2023bringing} reframe species classification by incorporating species images and contextual data into an MMKG, transforming classification into a link prediction task. This method links images to species labels, utilizing multi-modal contexts, such as visual cues, shooting times, and GPS coordinates, to enhance classification. Notably, it excels in identifying species outside their typical distribution ranges; by leveraging biological taxonomy, it improves generalization and the recognition of rare species. For example, \textit{a feline image captured in Africa is more likely to be classified as a tiger}. 
This establishes a new paradigm for task organization and execution, extending the scope of MKGC to cover tasks beyond conventional in-(MM)KG predictions and across various domains, thus signaling a significant and forward-looking development in the MMKG community.

\textbf{Apply Multi-modal Task for In-MMKG Task Augmentation:}
Various multi-modal tasks can enhance in-MMKG tasks. For example, Conditional Text-to-Image Generation could fill in missing modalities in MMKGs, thereby boosting the performance of MKGC and MMEA; Cross-modal Retrieval may serve to refine and expand image quality in MMKGs; Multi-modal Reasoning and Classification techniques can offer reasoning-based re-ranking of candidate entities in MKGC tasks. Additionally, there exists a plethora of unexplored potential feedback mechanisms that could further augment MMKG capabilities.
While these endeavors hold promise, we suggest a balanced approach, emphasizing the importance of future work to broaden the MMKG community's reach across various tasks and to more deeply explore MMKGs' unique potential and value.

%% file: 7.1-llm.tex
\subsection{Large Language Models}
The academic definition of LLMs, often associated with models possessing extensive parameters such as LLaMA-7B~\cite{DBLP:journals/corr/abs-2302-13971}, remains broad. 
These models' emergent abilities and Zero-shot Learning capabilities edge them closer to achieving AGI, underscoring their importance in NLP and multi-modal domains.
The integration of multi-modal knowledge within LLMs, as seen in recent studies, prompts the semantic web community to delineate their distinct value amidst evolving (MM)KG-driven multi-modal methodologies.

\textbf{\textit{(\rmnum{1})} Fine-Tuning:}
MMKGs provide a rich source of structured multi-modal data for Supervised Fine-Tuning (SFT) of MLLMs, especially in domain-specific applications~\cite{zheng2024exploring,DBLP:journals/corr/abs-2311-06503}. Training techniques effective for MMKGs in VLMs can also be applied to MLLMs, as discussed in \mbox{\S\,\ref{sec:mmapp}}. 
The challenge of extracting sufficient visual knowledge, as identified by Chen et al.~\cite{DBLP:journals/corr/abs-2311-11860}, alongside Zhou et al.'s~\cite{DBLP:journals/corr/abs-2308-15851} finding that 43\% of BLIP2~\cite{DBLP:conf/icml/0008LSH23} errors on the A-OKVQA dataset \cite{DBLP:conf/eccv/SchwenkKCMM22} could be addressed with proper knowledge integration, emphasizes the need for embedding explicit and especially long-tail knowledge into MLLMs~\cite{DBLP:journals/corr/abs-2312-06185}.
This process within MMKGs  can be realized along two distinct pathways: one involves active KG routing exploration for constructing specific instructions~\cite{DBLP:conf/emnlp/WanHYQB023}, and the other leverages self-instructing techniques to autonomously evolve and generate multi-grained, multi-modal instructional data~\cite{DBLP:conf/acl/WangKMLSKH23,DBLP:journals/corr/abs-2304-12244,DBLP:journals/corr/abs-2311-01487,DBLP:journals/corr/abs-2401-04695}.
Besides, the structured multi-modal relational data inherent in MMKGs provides an essential foundation for investigating the visual extrapolation abilities of purely visual LLMs, or Large Vision Models (LVMs)~\cite{DBLP:journals/corr/abs-2312-00785}, as well as MLLMs~\cite{DBLP:journals/corr/abs-2312-13286,DBLP:journals/corr/abs-2312-06109}.
Furthermore, MMKG data can be utilized to further explore the concept of multi-modal reversal curse~\cite{DBLP:journals/corr/abs-2311-07468}, where the ordering of knowledge entities in training data influences model comprehension, potentially limiting the model's understanding.

\textbf{\textit{(\rmnum{2})} Hallucination:} 
As LLMs rapidly advance, the risk of generating seemingly authentic but factually inaccurate web content is increasing. This phenomenon, known as \textit{hallucination}~\cite{suvery-Siren,found_hallucination,DBLP:journals/corr/abs-2311-07914}, often arises from outdated or incorrect training  encountered during the model training process, or from the frequent co-occurrence bindings of objects, affecting both LLMs and MLLMs~\cite{DBLP:journals/corr/abs-2311-17911,DBLP:journals/corr/abs-2401-06209,liu2024survey}.
To combat this, LAMM~\cite{DBLP:journals/corr/abs-2306-06687} incorporates 42K KG facts from Wikipedia and leveraged the Bamboo dataset~\cite{DBLP:journals/corr/abs-2203-07845} to refine commonsense knowledge in  Q\&A, underscoring the role of quality (MM)KGs in mitigating LLM hallucinations~\cite{DBLP:journals/corr/abs-2311-07914,DBLP:conf/emnlp/XuLCPWSL23}.
Developing robust hallucination detectors~\cite{DBLP:journals/corr/fact-hallu,mishra2024finegrained}  is crucial for identifying and curbing errors in LLM outputs. 
Future efforts could focus on pairing MMKGs with detection methods to improve multi-modal task precision and leveraging (MM)KGs for knowledge-aware statement rewriting to diminish factual hallucinations in LLM reasoning~\cite{DBLP:journals/corr/abs-2311-13314,DBLP:journals/corr/abs-2312-01701}.

\textbf{\textit{(\rmnum{3})} Agent:} Multi-agent Collaboration~\cite{xu2023urban,DBLP:journals/corr/abs-2312-17115,lu2024large}, simulating human cognitive processes, can dissect VQA reasoning paths and engage multiple (M)LLMs in collective problem-solving~\cite{wang2023towards,qiao2024autoact}. In this framework, KGs can initialize agent personalities~\cite{DBLP:journals/corr/abs-2310-02168,DBLP:journals/corr/abs-2308-10278}, providing a structured basis for intuitively designing character brains, enriching the interaction between agents and enhancing their collective reasoning capabilities.

Chain-of-thought (CoT) reasoning~\cite{DBLP:conf/nips/Wei0SBIXCLZ22} significantly improves LLMs' complex reasoning abilities by incorporating intermediate reasoning steps. This progress has catalyzed the emergence of various KG-focused applications~\cite{DBLP:journals/corr/abs-2311-09762,DBLP:journals/corr/abs-2307-07697}. For example, Sun et al.~\cite{DBLP:journals/corr/abs-2307-07697} 
demonstrate how LLMs can be used to interactively navigate KGs to extract knowledge for reasoning. Their Think-on-Graph (ToG) approach utilizes beam search to identify effective reasoning paths within KGs. Merging these innovations with MMKGs promises to expand the scope of tasks, especially in improving the ability of models to interpret and interact with diverse data types, such as images and text~\cite{mondal2024kamcot}. This integration moves us closer to achieving human-like multi-modal proficiency and paves the way for advanced machine intelligence.

\textbf{\textit{(\rmnum{4})} RAG:}
Retrieval Augmented Generation (RAG)~\cite{DBLP:journals/corr/abs-2312-05934} systems enhance (M)LLMs by incorporating long-tail knowledge beyond their parameter limits. However, excessive document retrieval can lead to contextually inappropriate answers~\cite{DBLP:journals/corr/abs-2401-05856}, increasing hallucination risks unless carefully designed prompts are used~\cite{DBLP:journals/corr/abs-2311-08377}.  The high information density and structured organization in KGs can mitigate this issue. Moreover, MMKGs can further aid multi-modal RAG by using various modalities as anchors~\cite{DBLP:journals/corr/abs-2312-04763}, offering more relevant and explanatorily powerful results than vector-based searches~\cite{DBLP:journals/corr/abs-2312-14135,DBLP:conf/iclr/0002IWXJ000023}.

\textbf{\textit{(\rmnum{5})} Editing:}
Editing LLMs typically involves fine-tuning a  limited number of parameters, similar to Lora-based methods, with the primary goal of correcting factual inaccuracies. For MLLMs, editing extends to updating information to preserve both factual accuracy and cross-modal consistency, a challenging endeavor due to the difficulty of synchronizing knowledge across different modalities~\cite{DBLP:conf/emnlp/0008TL0WC023}. Utilizing MMKG-driven strategies becomes crucial for accurately adjusting pertinent facts during the editing process~\cite{DBLP:journals/corr/abs-2301-10405}, thus ensuring contextually accurate updates.

\textbf{\textit{(\rmnum{6})} Alignment:} 
Aligning LLMs with human preferences is crucial for their success as language assistants.  Teaching models to acknowledge their limits, such as admitting "\textit{I don’t know}" when appropriate, is essential. However, previous SFT-based models often generate responses without indicating uncertainty. Utilizing Reinforcement Learning from Human Feedback (RLHF)~\cite{DBLP:journals/corr/abs-2310-06452,DBLP:journals/corr/abs-2310-11971} has proven effective in aligning LLMs more closely with human preferences,  improving user satisfaction by ensuring models not only respond accurately but also recognize the limits of their knowledge~\cite{DBLP:journals/corr/abs-2312-09390,DBLP:journals/corr/abs-2312-07000}. Recent research trends focus on structuring LLM preferences with structured knowledge, a notable advancement. KnowPAT~\cite{DBLP:journals/corr/abs-2311-06503} aligns model preferences with knowledge, guiding LLMs to select pertinent knowledge for specific inquiries. This strategy, which involves tailoring data to human preferences and ensuring factual accuracy in domain-specific Q\&A, is a promising direction for leveraging MMKG in multi-modal knowledge alignment within LLMs.

\textbf{\textit{(\rmnum{7})} MMKG Refinement:} 
LLMs offer the capability to augment MMKGs through their advanced text comprehension and generation skills. Recent work, such as \cite{DBLP:journals/corr/abs-2308-13916-KGllama, DBLP:journals/corr/abs-2310-06671-Kopa}, explores LLM-based KGC. Specifically,  KoPA \cite{DBLP:journals/corr/abs-2310-06671-Kopa} integrates KG structural knowledge into LLMs to enable structure-aware reasoning.
Moreover, with the continuous growth and evolution of online data, LLMs can support the continuous learning and self-updating of MMKGs, serving as active annotators~\cite{DBLP:conf/emnlp/ZhangLMZ023}.

\textbf{\textit{(\rmnum{8})} MMKG MoE:}
The Mixed of Expert (MoE) architecture shows outstanding performance in LLM applications. Initially, it engages input samples through a GateNet or router for multi-class categorization, determining the allocation of tokens to appropriate experts. This critical process, known as \textit{experts selection}, is central to MoE's concept, often characterized as sparse activation in academia~\cite{ismail2023interpretable,DBLP:journals/corr/abs-2312-09979,llama-moe-2023,DBLP:journals/corr/abs-2401-06066,DBLP:journals/corr/abs-2401-15947}. These experts then process the inputs to formulate final predictions.
Regarding domain-specific MMKGs in fields like biology, e-commerce, and world geography, an innovative direction involves creating an extensive MMKG library (or repository). 
This library would house varied MMKGs, each tailored to specific domains, allowing downstream tasks to adaptively select relevant MMKG information in a manner akin to MoE's.  Exploring this conceptual approach could not only catalyze developments in MMKG-level retrieval and re-ranking but also foster the seamless integration of MMKGs into model parameters, merging their utility with the dynamic allocation efficiency of MoE architecture.

%% file: 8-conclusion.tex
\section{Conclusion}\label{sec:conclusion}
This paper provides a comprehensive review of  Multi-Modal Knowledge Graphs (MM4KG) development and the historical progression of (MM)KG in multi-modal tasks (KG4MM). We have carefully reviewed and analyzed all relevant works up to \textbf{January 2024},  tracing the field's evolution. Our focus includes detailed benchmarking and methodological analysis of critical tasks such as KG-driven Visual Question Answering in KG4MM and Multi-Modal Entity Alignment in MM4KG. Our goal is to construct a systematic blueprint of the domain, offering a valuable reference for researchers currently involved in or planning to explore this area. 
In summary, this review aspires to be a foundational guide, highlighting the path and prospects of MMKG research, thereby supporting future academic efforts.